\documentclass[10pt,twocolumn,letterpaper]{article}
\usepackage[pagenumbers]{cvpr}

\usepackage{cuted} %

\usepackage{currfile} %

\usepackage{caption} %

 \usepackage{microtype}

\usepackage{microtype}
\usepackage{booktabs}
\usepackage{pifont}
\newcommand{\cmark}{\ding{51}}
\newcommand{\xmark}{\ding{55}}
\usepackage{amsmath,amssymb}
\usepackage{enumitem}
\usepackage{algorithm}
\usepackage{algpseudocode}
\usepackage{xcolor}
\DeclareUnicodeCharacter{2212}{\ensuremath{-}}
\setlist{nosep,leftmargin=*}
\definecolor{cvprblue}{rgb}{0.21,0.49,0.74}
\usepackage[pagebackref,breaklinks,colorlinks,allcolors=cvprblue]{hyperref}
\def\confName{}
\def\confYear{2027}

\newif\ifshowcomments
\showcommentsfalse

\title{NRG-Twins: Pixel-Level Ground Truth for Rigidity Violations in Generated Videos}
\author{Chenrui Fan\qquad Qiyang Hu\qquad Luca Scharr\qquad Paolo Favaro\\University of Bern}

\newif\ifwithappendix\withappendixfalse %
\usepackage{tabularx}
\newcolumntype{L}{>{\raggedright\arraybackslash}X}
\newcommand{\suptab}[1]{\renewcommand{\thetable}{#1}\renewcommand{\theHtable}{#1}}
\newcommand{\suptabend}{\addtocounter{table}{-1}}
\newcommand{\supfig}[1]{\renewcommand{\thefigure}{#1}\renewcommand{\theHfigure}{#1}}
\newcommand{\supfigend}{\addtocounter{figure}{-1}}

\makeatletter\AtBeginDocument{\def\theHALG@line{\thealgorithm.\arabic{ALG@line}}}\makeatother
\begin{document}
\maketitle

\begin{abstract}
Video generators now produce realistic clips of scenes and objects, which makes generated video an attractive source of 3D content. Yet generated video is not 3D consistent: static content slowly deforms or slides, 3D reconstruction degrades, and the errors are too subtle to label by hand. To make generators 3D consistent, or to build rewards and guidance that push them there, we need data that say where and by how much a generated video breaks rigidity, and these data must come from generated video itself. We obtain per-pixel ground-truth labels from real--generated twins: a real clip paired with its re-creation by a keyframe-conditioned video generator. After inserting real keyframes so that the twin camera roughly follows the real one, we measure per pixel, in degrees, how far the twin departs from a rigid scene seen along its own camera path. We release \underline{N}on-\underline{R}igidly \underline{G}enerated Twins (NRG-Twins), 871 twins, the code that builds them for new generators, and a benchmark with an official test split. A detector trained on these data, the \underline{N}on-\underline{R}igidly \underline{G}enerated \underline{P}ixel \underline{D}etector (NRG-PD), predicts the per-pixel angle from RGB alone. It separates inconsistent from consistent pixels with AUROC 0.85, against 0.40--0.74 for seven existing consistency and motion-segmentation methods, and its predictions improve 3DGS of video that contains generated content.
\end{abstract}

\section{Introduction}
\begin{table*}[t]\centering\small\setlength{\tabcolsep}{3pt}
\resizebox{\textwidth}{!}{%
\begin{tabular}{lcccccccc}\toprule
Method & Gen.\ video & Video only & Clip score & Per-pixel map & Static 3D & Degrees & Per-pixel GT labels & AUROC \\\midrule
VBench-2.0 (multi-view consistency) & \cmark & \cmark & \cmark & \xmark & \cmark & \xmark & \xmark & --\textsuperscript{*} \\
MEt3R & \cmark & \cmark & \cmark & \cmark\textsuperscript{\dag} & \cmark & \xmark & \xmark & 0.595 / 0.630\textsuperscript{\ddag} \\
GeCo & \cmark & \cmark & \cmark & \cmark\textsuperscript{\dag} & \cmark & \xmark & \xmark & 0.585 \\
MonST3R & \xmark & \cmark & \xmark & \cmark\textsuperscript{\S} & \xmark & \xmark & \xmark & 0.580 \\
Easi3R & \xmark & \cmark & \xmark & \cmark\textsuperscript{\S} & \xmark & \xmark & \xmark & 0.735 \\
MegaSaM & \xmark & \cmark & \xmark & \cmark\textsuperscript{\S} & \xmark & \xmark & \xmark & 0.398 \\
RoMo & \xmark & \cmark & \xmark & \cmark\textsuperscript{\S} & \xmark & \xmark & \xmark & 0.652 \\
VGGT4D & \xmark & \cmark & \xmark & \cmark\textsuperscript{\S} & \xmark & \xmark & \xmark & 0.562 \\
\textbf{NRG-PD (Ours)} & \cmark & \cmark & \cmark & \cmark & \cmark & \cmark & \cmark & $0.851\pm0.036$\textsuperscript{\P} \\
\bottomrule
\end{tabular}}
\caption{What existing methods provide; a check means the method has the property. Gen.\ video: designed for generated video; Video only: needs nothing but the video; Clip score: one score per video; Per-pixel map: a map for every frame; Static 3D: targets static content, not moving objects; Degrees: error as an angle; Per-pixel GT labels: checked against per-pixel ground-truth labels on generated video. AUROC: separating clearly inconsistent from consistent pixels on our test set (Sec.~\ref{sec:exp}); 0.5 is chance. \textsuperscript{*}No per-pixel output. \textsuperscript{\dag}Map used inside its score; GeCo: per-frame map, evaluated on its WarpBench / OccluBench (warps and occlusions injected into real clips). \textsuperscript{\ddag}Frame $t$ vs.\ $t{+}4$ / vs.\ frame 0. \textsuperscript{\S}Map of moving objects. \textsuperscript{\P}Mean $\pm$ SD over 3 seeds.}\label{tab:capabilities}
\end{table*}
Video generators now produce realistic clips of scenes and objects, including sequences in which the camera orbits an object. This makes generated video a tempting source of 3D content: video is plentiful and easy to train on, a prompt can describe almost any object or scene, and 3D data are scarce in comparison (the largest public 3D object collection, Objaverse-XL~\cite{deitke2023objaverse}, has about 10M objects). Recent 3D generation already builds on video and multi-view diffusion priors (e.g., SV3D~\cite{voleti2024sv3d}, CAT3D~\cite{gao2024cat3d}).

\begin{figure}[t]
\centering
\includegraphics[width=\columnwidth]{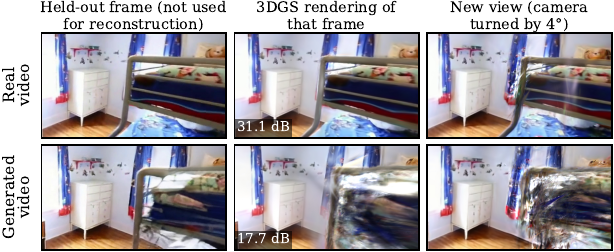}
\caption{\textbf{A generated video cannot (yet) be reconstructed in 3D as well as a real video.} Top: a real RealEstate10K clip; bottom: a video generated from its first and last frames. We fit 3DGS to each video with frames 33--47 held out. Columns: held-out frame 40, its 3DGS rendering (PSNR), and a new view with the camera turned by 4$^\circ$. The generated video looks right frame by frame, but its 3D reconstruction breaks.}
\label{fig:motivation}
\end{figure}
\begin{figure}[t]
\centering
\includegraphics{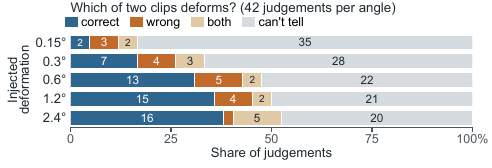}\vspace{-6pt}
\caption{Human study (7 raters): the same real clip shown twice, one side with a known rigid 2D deformation of one region; share of correct, wrong, both and can\textquoteright t-tell answers per deformation angle (42 judgements per angle; counts on the bars). %
}
\label{fig:human}
\end{figure}

Generated video, however, looks right frame by frame but suffers from non-rigid inconsistency: static content slowly deforms or slides over time, and 3D reconstruction pays for it (see, \emph{e.g.}, Fig.~\ref{fig:motivation}). We let a video generator re-create 29 real clips, each conditioned on a few of the original frames: 3DGS~\cite{kerbl20233d} of a real clip reaches 37.6\,dB on its held-out frames, but only 32.4\,dB with its generated clips. These errors are hard to see: in a study with seven people, an injected deformation of 0.15--2.4$^\circ$ was identified in only 2 to 16 of 42 judgements per angle (see Fig.~\ref{fig:human} and  Supp.~H), so such data cannot be labelled by hand.

To make generators 3D consistent and build reward models or guidance toward this goal, we need data showing where and how much generated videos break rigidity. These data must come from generated videos themselves: deformations simulated on synthetic or real videos do not match generators' errors. The difficulty is ground truth, since generated videos have no real scenes behind them.

Existing work does not provide such data (Table~\ref{tab:capabilities}). 3D-consistency evaluators (L3DE~\cite{chang2025far}, GeCo~\cite{gu2025geco}, MEt3R~\cite{asim2025met3r}, SGC~\cite{dou2026measuring}) report video- or pair-level scores; GeCo and L3DE also localize errors (GeCo evaluates its maps on its own WarpBench and OccluBench, built by injecting warps and occlusions into real video; L3DE against human-marked regions), but none checks it against per-pixel ground-truth labels on generated video. Motion-segmentation and 4D methods (RoMo~\cite{goli2025romo}, VGGT4D~\cite{hu2026vggt4d}, MonST3R~\cite{zhang2025monst3r}, Easi3R~\cite{chen2025easi3r}, MegaSaM~\cite{li2025megasam}) find moving objects in real video, not inconsistency in static content. On generated videos with per-pixel ground-truth labels (Sec.~\ref{sec:exp}), the dense maps produced by these seven methods leave room for more precise localization of inconsistent pixels. No dataset gives per-pixel ground-truth labels of 3D inconsistency on \textbf{generated videos}; BrokenVideos~\cite{lin2025brokenvideos} marks visual artifacts in generated video with human masks, without measuring 3D error.

\begin{figure}[t]
\centering
\includegraphics[width=\linewidth]{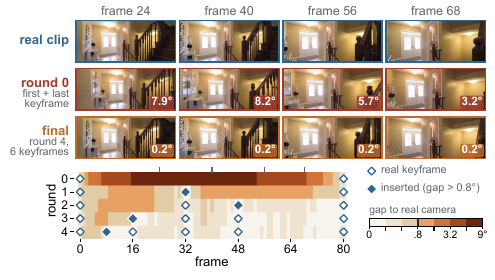}
\caption{Keyframe insertion pulls the twin camera onto the real camera (one training twin). Top: frames of the real clip, of the round-0 twin (keyframes: first and last real frame) and of the final twin (round 4, six keyframes); number on each frame: its rotation gap to the real camera. Bottom: gap per frame in each round; open diamonds: keyframes used; filled diamonds: keyframe inserted in that round; right: legend and colour scale of the gap.}
\label{fig:keyframes}
\end{figure}
\begin{figure*}[t]
\centering
\includegraphics[width=.9\textwidth,trim=0 12bp 0 3bp,clip]{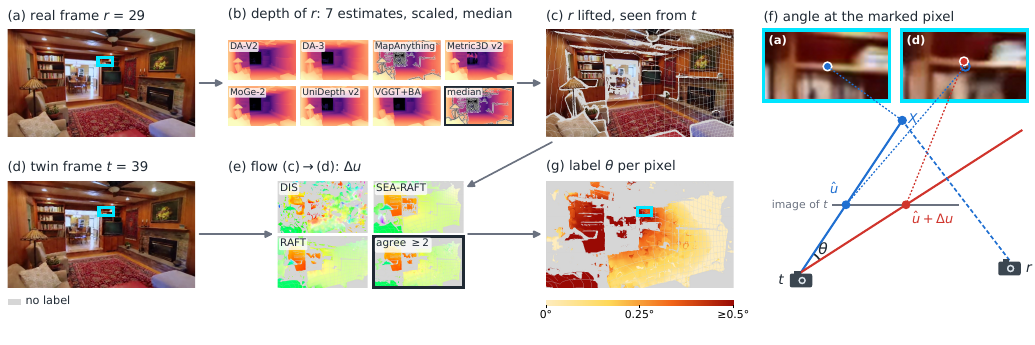}\vspace{-4pt}
\caption{Per-pixel ground-truth label for one twin frame (training twin, frame $t=39$). (a) The fitting frame $r$ with the smallest median $t_\perp/Z$ for the twin camera (Sec.~3.3). (b) Its depth from six single-image methods and VGGT+BA, each scaled to the sparse 3D points of the camera reconstruction (Sec.~3.2) and fused by the per-pixel median. (c) $r$ lifted with the fused depth and projected into the twin camera: where a rigid scene would put the content. (d) The twin frame. (e) Optical flow from (c) to (d) by DIS, SEA-RAFT and RAFT, kept where at least two methods pass the forward--backward check and agree: the residual $\Delta u$. (f) At the marked pixel, $X$ is the lifted pixel of $r$, $\hat u$ its projection into $t$ and $\hat u+\Delta u$ the same content in the twin; the NRG angle $\theta$ is the angle between the two viewing rays. (g) NRG-angle label after subtracting the noise measured on the real keyframes (Algorithm~2 in the supplement); grey: no label.}
\label{fig:truth}
\end{figure*}
We obtain per-pixel ground-truth labels by pairing each RealEstate10K~\cite{zhou2018stereo} clip with its re-creation by a keyframe-conditioned video generator using a few of the clip's real frames, forming what we call a real--generated twin (Sec.~\ref{sec:dataset}). Such pairs are used to train a geometry corrector (VGC~\cite{lyu2026learning}), but its generated camera only approximately follows the real one, without per-pixel ground-truth labels. A twin shows the same content as its real clip, but its camera can deviate considerably; assessing rigidity thus requires aligning cameras to avoid mistaking camera differences for deformation. We therefore add real keyframes so the twin camera roughly follows the real one, then measure, per pixel, the angle $\theta$ between where content would appear under scene rigidity and where it actually appears in the twin camera. \underline{N}on-\underline{R}igidly \underline{G}enerated (NRG) pixels thus have a large $\theta$, which we call the NRG angle (Sec.~\ref{sec:dataset}). We validate this measurement by injecting known motion. Requiring only real videos and a keyframe-conditioned generator, the procedure allows rebuilding the dataset for any new generator.

With these data we train a detector, the \underline{N}on-\underline{R}igidly \underline{G}enerated \underline{P}ixel \underline{D}etector (NRG-PD), that predicts the per-pixel angle from the generated video alone (Sec.~\ref{sec:method}). On generated videos with per-pixel ground-truth labels, it separates clearly inconsistent from consistent pixels with pixel AUROC 0.85 (0.5 is chance), against 0.40--0.74 for seven existing methods (Sec.~\ref{sec:exp}). Its predictions also help 3D reconstruction: when part of a clip deforms, masking the pixels flagged by NRG-PD (+injected), the same architecture trained also on injected deformations (Sec.~\ref{sec:recon}), improves 3DGS by $+3.07$\,dB over a random mask of the same size; when generated frames are inserted into a real clip, dropping the frames NRG-PD flags improves 3DGS by $+4.36$\,dB over random dropping. On prompt-generated video without ground-truth labels, the tracks it flags have larger reprojection error than the tracks it calls clean (track AUROC 0.85, the highest among the seven methods).

In summary, we make the following contributions:
\begin{itemize}
\item \textbf{A dataset of generated video with per-pixel 3D-inconsistency ground truth labeling.} \underline{N}on-\underline{R}igidly \underline{G}enerated Twins (NRG-Twins): 871 real--generated twins whose per-pixel NRG angle is given in degrees. We will release the data and the code that builds it, so new generators can be added.
\item \textbf{A benchmark.} An official test split, whose labels are public but not for model selection or tuning, and a pixel-level protocol, on which we evaluate seven existing consistency and motion-segmentation methods against per-pixel ground-truth labels on generated video.
\item \textbf{A detector and its application to 3D reconstruction.} NRG-PD predicts the per-pixel angle from RGB alone; its predictions identify inconsistent content in prompt-generated video and we show how this can be used to improve 3D reconstructions via 3DGS on generated videos. 
\end{itemize}
\section{Related Work}
\paragraph{3D consistency of generated video.}

Video benchmarks score 3D consistency per video or frame pair: VBench~\cite{huang2024vbench}, VBench-2.0~\cite{zheng2025vbench}, WorldScore~\cite{duan2025worldscore}, SGC~\cite{dou2026measuring}, MEt3R~\cite{asim2025met3r}. L3DE~\cite{chang2025far} localizes errors using Grad-CAM and human annotations, while GeCo~\cite{gu2025geco} predicts per-frame inconsistency maps evaluated on synthetic warps and occlusions injected into real video. VideoGPA~\cite{du2026videogpa} fine-tunes a generator using a geometric consistency score; BrokenVideos~\cite{lin2025brokenvideos} uses human masks to mark generated-video artifacts. Unlike these methods, NRG-Twins provides per-pixel geometric ground-truth labels on generated video, measured in degrees.

\paragraph{Moving objects and 4D reconstruction.}
RoMo~\cite{goli2025romo}, VGGT4D~\cite{hu2026vggt4d}, MonST3R~\cite{zhang2025monst3r}, Easi3R~\cite{chen2025easi3r} and MegaSaM~\cite{li2025megasam} separate moving objects from the static scene in real video. However, Non-rigid inconsistency is a different target: the content is meant to be static, and its motion is small and spread over the frame. We use the dense maps of these methods, together with those of GeCo and MEt3R, as baselines (Sec.~\ref{sec:exp}).

\paragraph{Inconsistency in generated content for 3D.}
GaussVideoDreamer~\cite{hao2025gaussvideodreamer}, RIGI~\cite{wang2026rigi}, and VDEGaussian~\cite{xiao2025vdegaussian} address inconsistent views in 3D reconstruction, while World Reconstruction from Inconsistent Views~\cite{hollein2026world} aligns per-frame geometry non-rigidly before fitting 2D Gaussian splats. VGC~\cite{lyu2026learning} trains a geometry corrector on real–generated pairs using real keyframes for segment-wise generation. We adopt this pairing, add real keyframes to align the generated camera, and measure per-pixel NRG angles along its own camera path instead of correcting or masking errors.

\paragraph{Robust reconstruction.}
NeRF-W~\cite{martin2021nerf}, RobustNeRF~\cite{sabour2023robustnerf}, SpotLessSplats~\cite{sabour2025spotlesssplats}, WildGaussians~\cite{kulhanek2024wildgaussians} and RobustSplat~\cite{fu2025robustsplat} reconstruct scenes from images with transient content by down-weighting pixels that the fit cannot explain. Non-rigid inconsistency in generated video is a similar obstacle for 3D reconstruction; we test whether NRG-PD masks help RobustSplat on video with generated content (Sec.~\ref{sec:exp}).

\section{Twin Dataset with Ground-Truth Labels}\label{sec:dataset}
We build NRG-Twins using Wan [47] to obtain per-pixel ground-truth labels of 3D inconsistency. However, the procedure readily generalizes to other keyframe-conditioned video generators.

The main challenge is ground truth: a generated video has no real scene from which its distortions could be measured. Thus, we start from real videos. Following VGC~\cite{lyu2026learning}, which pairs real clips with their keyframe-conditioned recreations to train a geometry corrector, a keyframe-conditioned generator recreates a real clip from a few of its frames; the keyframes control how far the generated video departs from the real one, and the real clip, with its cameras and depth, serves as the reference. Optical flow then locates the same content in both videos, and the NRG angle measures its difference to a valid rigid 3D camera motion, namely the generated video's own camera path. It may differ from the real one. In the following we describe each part: twin construction, cameras, depth, optical flow and the NRG angle.

\subsection{Twin construction}\label{sec:twin}
We take real clips from RealEstate10K~\cite{zhou2018stereo} (the pixelSplat~\cite{charatan2024pixelsplat} version, plus earlier YouTube originals) and re-create each 81-frame clip ($848\times480$) with Wan~\cite{wan2025wan}. The generator works segment by segment, each call conditioned on its start and end keyframe, as in VGC~\cite{lyu2026learning}; it never receives the camera trajectory. We start from keyframes 0 and 80. After each generation round we estimate the per-frame camera rotations of the real clip, $R^{r}_t$, and of the twin, $R^{w}_t$, with VGGT~\cite{wang2025vggt} applied directly to the frames (Supp.~A.1). The two estimates differ by an unknown global rotation $Q$, which we fit on the current keyframes set $\mathcal{K}$, where twin and real frame are identical:
\begin{equation} Q=\operatorname*{arg\,min}_{Q\in SO(3)}\sum_{k\in\mathcal{K}}\big\lVert R^{r}_k-Q\,R^{w}_k\big\rVert_F^2, \end{equation}
a Wahba problem~\cite{wahba1965least} with a closed-form solution (Supp.~A.1). The rotation gap of frame $t$ is the angle of the remaining relative rotation,
\begin{equation} \delta_t=\arccos\Big(\tfrac{1}{2}\big(\operatorname{tr}\big((R^{r}_t)^{\top}Q\,R^{w}_t\big)-1\big)\Big). \end{equation}
If $\max_t\delta_t\le0.8^\circ$, about twice the error of the camera estimate (Supp.~A.1), we stop; otherwise we insert the real frame at the largest gap as a further keyframe (a frame index divisible by 4, as the generator's segment length requires, at least 8 frames from existing keyframes) and regenerate the two adjacent segments, at most 6 times (Fig.~\ref{fig:keyframes}; Algorithm~1 in the supplement). Keyframes are copied back into the twin byte for byte; they are excluded from training and measure the noise of each twin's labels (Sec.~\ref{sec:noise}). 
On 40 training twins, keyframe insertion reduces the median camera rotation gap from 1.20° to 0.30°, while the median Sampson error remains 0.42 px, twice that of real video (0.21 px; Supp.~A.3). Thus,  keyframe insertion improves camera alignment while the deformation of the content remains.

\subsection{Cameras}

The real clip and twin share fixed dataset intrinsics $K$ (Supp.~A.5). We estimate their poses using both VGGT and COLMAP~\cite{schonberger2016structure,schonberger2016pixelwise}, each followed by bundle adjustment. We use COLMAP poses where registration succeeds and the solution agrees with VGGT across the clip, and VGGT poses otherwise. A similarity transform fitted on the 73 fitting frames F (81 frames minus 8 held-out real frames for label-error evaluation; Supp.~A.2) aligns the twin to the real clip. Across 702 training pairs, the median residual rotation gap is $0.25^\circ$ on non-keyframes and $0.22^\circ$ on keyframes, where it should be zero, indicating camera-estimation error.

\subsection{Depth and reprojection}\label{sec:depth}
We take as reference the fitting frame $r\in\mathcal{F}$ with the smallest median $t_\perp/Z$ (defined below) for the twin camera of frame $t$.

\emph{Fused depth.} Every frame $f$ has seven depth maps $Z_{m,f}$: six single-image estimators (Depth Anything V2~\cite{yang2024depthanythingv2}, Depth Anything 3~\cite{lin2025depthanything3}, MapAnything~\cite{keetha2026mapanything}, Metric3D v2~\cite{hu2024metric3dv2}, MoGe-2~\cite{wang2025moge2}, UniDepth v2~\cite{piccinelli2025unidepthv2}) and the depth of the VGGT reconstruction of the real clip; dense depth from COLMAP or 3DGS is not used (Supp.~A.5). Each map is scaled to the sparse 3D points $k\in\mathcal{P}_f$ of the camera reconstruction (Sec.~3.2) seen in $f$ (depth $z_k$, pixel $\mathbf{u}_k$), and the scaled maps are fused by the per-pixel median:
\begin{gather} s_{m,f}=\operatorname*{median}_{k\in\mathcal{P}_f}\frac{z_k}{Z_{m,f}(\mathbf{u}_k)},\\ D_f(\mathbf{u})=\operatorname*{median}_{m}\; s_{m,f}\,Z_{m,f}(\mathbf{u}). \end{gather}

\emph{Reprojection.} Each reference pixel $\mathbf{u}$ is lifted to the 3D point $\mathbf{X}$ and projected point by point (not rendered) into the twin camera; $\hat{\mathbf{u}}$ is where the content would appear if the scene were rigid (Fig.~\ref{fig:truth}c):
\begin{equation} \mathbf{X}=D_r(\mathbf{u})\,K^{-1}\tilde{\mathbf{u}},\qquad \hat{\mathbf{u}}=\pi\big(K(R_{t\leftarrow r}\mathbf{X}+\mathbf{t}_{t\leftarrow r})\big), \end{equation}
with intrinsics $K$, relative pose $(R,\mathbf{t})$, homogeneous coordinates $\tilde{\cdot}$ and perspective division $\pi$. Where several points land on one pixel, the nearest is kept; points outside the twin image $\Omega_t$ get no label.

\emph{Where depth is reliable.} A relative depth error $\delta Z/Z$ moves $\hat{\mathbf{u}}$ by a ray angle of about $(t_\perp/Z)(\delta Z/Z)$, with $t_\perp$ the camera translation perpendicular to the ray and $Z=D_r(\mathbf{u})$. A pixel is labeled only if all seven depth maps are valid there and $t_\perp/Z\le0.01$ (in the next bin, $0.01$--$0.03$, the 95th percentile of the depth-induced error is $0.11$--$0.14^\circ$, close to the error floor of the labels; Supp.~A.5). On LiDAR (ARKitScenes~\cite{baruch2021arkitscenes}) and laser-scan depth (ScanNet++~\cite{yeshwanth2023scannet++}), this keeps the 95th percentile of the depth-induced error at $0.032^\circ$ and $0.034^\circ$, about a fifth of the noise floor of the labels and well below the optical-flow error (Supp.~A.5).

\subsection{Optical flow}\label{sec:flow}
Optical flow from the reprojected reference to the twin frame locates the actual content. We run three methods, DIS~\cite{kroeger2016fast}, SEA-RAFT~\cite{wang2024sea} and RAFT~\cite{teed2020raft}. A method passes at a pixel if its forward--backward error is small, and two methods agree if their flows are close (both in pixels):
\begin{gather} \big\lVert \mathbf{f}_m(\mathbf{u})+\mathbf{b}_m\big(\mathbf{u}+\mathbf{f}_m(\mathbf{u})\big)\big\rVert\le 1+0.05\,\lVert\mathbf{f}_m(\mathbf{u})\rVert,\\ \lVert\mathbf{f}_i(\mathbf{u})-\mathbf{f}_j(\mathbf{u})\rVert\le 1+0.05\,\lVert\mathbf{f}(\mathbf{u})\rVert. \end{gather}
Let $n(\hat{\mathbf{u}})$ be the size of the largest subset of passing methods that agree pairwise; if $n(\hat{\mathbf{u}})\ge2$, its component-wise median is the residual displacement $\Delta\mathbf{u}$; otherwise, and for occluded pixels, the pixel gets no label. On real held-out frames, where the true residual is zero, this flow has a 95th-percentile error of $0.116^\circ$ and measures only 0.052\% of pixels wrongly as $\geq0.6^\circ$ (Supp.~A.5).

\subsection{NRG angle}\label{sec:nrgangle}
For a pixel $\mathbf{u}$ of the reference frame $r$, $\hat{\mathbf{u}}$ is its projection into twin frame $t$ under a rigid scene (Sec.~\ref{sec:depth}) and $\Delta\mathbf{u}$ is the flow residual at $\hat{\mathbf{u}}$ (Sec.~\ref{sec:flow}), so $\hat{\mathbf{u}}+\Delta\mathbf{u}$ is where the same content actually appears in the twin. In homogeneous coordinates $\tilde{\mathbf{x}}$, the vector $K^{-1}\tilde{\mathbf{x}}$ is the direction of the viewing ray through image point $\mathbf{x}$ in the twin camera. The NRG angle is the angle between the two rays (Fig.~\ref{fig:truth}f),
\begin{gather} \theta(\hat{\mathbf{u}})=\arccos\frac{\mathbf{a}^{\top}\mathbf{b}}{\lVert\mathbf{a}\rVert\,\lVert\mathbf{b}\rVert},\\ \mathbf{a}=K^{-1}\tilde{\hat{\mathbf{u}}},\qquad \mathbf{b}=K^{-1}\widetilde{(\hat{\mathbf{u}}+\Delta\mathbf{u})}, \nonumber\end{gather}
in degrees, stored at the twin pixel $\hat{\mathbf{u}}$; $\theta=0$ means that the content appears where a rigid scene would put it. Because it is measured between viewing rays, $\theta$ does not depend on image resolution, and the errors of depth (Sec.~\ref{sec:depth}) and flow (Sec.~\ref{sec:flow}) are stated in the same unit. A twin pixel receives a label only if all checks pass:
\begin{equation}\begin{gathered} \ell(\hat{\mathbf{u}})=\mathbf{1}\big[\hat{\mathbf{u}}\in\Omega_t\big]\;\mathbf{1}\big[n(\hat{\mathbf{u}})\ge2\big]\\ \cdot\,\mathbf{1}\big[t_\perp/Z\le0.01\big], \end{gathered}\end{equation}
with $\mathbf{1}[\cdot]$ the indicator function.

\subsection{Label noise from keyframes}\label{sec:noise}
The keyframes are real frames inside the twin, so their true NRG angle is zero and the angle measured there is the noise of the labelling pipeline. For keyframe $k$ we measure $\hat\sigma_k^2=\operatorname{mean}_{\mathbf{u}}\min(\theta_k(\mathbf{u}),3^\circ)^2$ over its labelled pixels and interpolate linearly for frames between keyframes $k_1<t<k_2$ (Supp.~A.4):
\begin{equation} \hat\sigma^2(t)=\frac{(k_2-t)\,\hat\sigma_{k_1}^2+(t-k_1)\,\hat\sigma_{k_2}^2}{k_2-k_1}. \end{equation}
If the measured residual is the true one plus independent zero-mean noise, $\mathbb{E}[\theta^2]\approx\theta_{\mathrm{true}}^2+\sigma^2$, so evaluation uses the noise-subtracted angle
\begin{equation} \theta_c=\sqrt{\max\!\big(\min(\theta,3^\circ)^2-\hat\sigma^2,\;0\big)}, \end{equation}
where the clip at $3^\circ$ limits the long-tail variance (Supp.~A.4). Training uses the same quantity without the square root, in units of $0.6^\circ$, the angle above which a measured value is unlikely to be measurement error alone (Supp.~A.4): $T=(\min(\theta,3)^2-\hat\sigma^2)/0.6^2$, so a squared-error loss regresses the noise-subtracted target directly and needs no $\hat\sigma^2$ at inference. A twin is excluded if its raw keyframe RMS exceeds $0.6^\circ$, its noise-subtracted value exceeds $0.3^\circ$, or the noise cannot be computed (Supp.~A.4). A rotation of known angle applied to a region of twin frames, with cameras, depth and references fixed, is measured with slope 1.001 and median absolute error $0.020^\circ$ (Supp.~A.6).

\subsection{NRG-Twins}
NRG-Twins contains 871 twins: 702 for training, 82 for validation and 87 for testing, split by whole YouTube video (8:1:1) so that no YouTube video is shared between training and the other splits. After keyframe exclusion and noise subtraction the splits have 53,501 / 6,303 / 6,639 frames and 10.42\,B / 1.20\,B / 1.31\,B labelled pixels; NRG angles $\geq0.6^\circ$ cover 1.49\% / 1.37\% / 1.64\% of the pixels and angles below $0.075^\circ$ about 67\%. The benchmark uses 83 test twins with labelled keyframes (Sec. ~\ref{sec:exp}). We will release the twins, the list of real clips, the label maps, cameras, depth and the construction code on Hugging Face for research use. Building one twin costs at least 0.35 GPU-hours and 3.5 CPU core-hours; twins from LTX-2.3~\cite{hacohen2026ltx2}, ScanNet++~\cite{yeshwanth2023scannet++} and DL3DV~\cite{ling2024dl3dv} are not part of this version (Supp.~A.5).

\section{NRG-PD}\label{sec:method}
\begin{table*}[t]
\centering\footnotesize
\setlength{\tabcolsep}{4pt}
\begin{tabular*}{\textwidth}{@{\extracolsep{\fill}}@{}lcccc@{\hspace{10pt}}ccc@{}}\toprule
 & \multicolumn{4}{c}{Ranking by NRG angle} & \multicolumn{3}{c}{Retrieval of NRG pixels} \\
\cmidrule(lr){2-5}\cmidrule(l){6-8}
Method & AUROC & Kendall $\tau$ & Clip corr. & MAE ($^\circ$) & Average precision & Precision (top 5\%) & Recall (top 5\%) \\
\midrule
\textbf{NRG-PD (ours)} & \textbf{0.851 $\pm$ 0.036} & \textbf{0.646 $\pm$ 0.053} & 0.112 & \textbf{0.106} & \textbf{0.185 $\pm$ 0.075} & \textbf{0.089} & \textbf{0.285} \\
Easi3R & 0.735 & 0.505 & \textbf{0.139} & --- & 0.056 & 0.041 & 0.130 \\
RoMo & 0.652 & 0.498 & $-$0.026 & --- & 0.038 & 0.055 & 0.176 \\
MEt3R & 0.595 & 0.393 & 0.130 & --- & 0.031 & 0.044 & 0.141 \\
GeCo & 0.585 & 0.287 & $-$0.130 & --- & 0.034 & 0.045 & 0.144 \\
MonST3R & 0.580 & 0.337 & 0.061 & --- & 0.033 & 0.037 & 0.117 \\
VGGT4D & 0.562 & 0.140 & $-$0.061 & --- & 0.028 & 0.019 & 0.060 \\
MegaSaM & 0.398 & $-$0.375 & 0.082 & --- & 0.020 & 0.020 & 0.063 \\
\bottomrule
\end{tabular*}
\caption{Results on the 83 test twins (Sec.~\ref{sec:protocol}). Left: each method on its own valid pixels. Right: only the $4.4\times10^8$ pixels scored by all methods. Ours: mean over 3 seeds; $\pm$: SD over seeds (left), half-width of the 95\% interval (right). Chance: AUROC 0.5, Kendall $\tau$ 0, average precision 0.023; the precision ceiling is 0.31.}
\label{tab:s5.7}
\end{table*}
\begin{figure}[t]
\centering
\includegraphics[width=.8\linewidth]{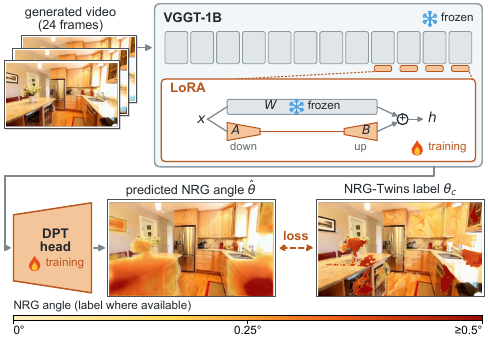}
\caption{NRG-PD. A 24-frame generated clip enters VGGT-1B, whose 24 frame/global attention layer pairs stay frozen; LoRA (rank 16) adds trainable low-rank branches to the last 8 pairs. A DPT head initialised from VGGT's depth head predicts the per-pixel NRG angle and is trained against the NRG-Twins labels where they exist. Example: one frame of a validation twin.}
\label{fig:model}
\end{figure}
NRG-PD predicts the NRG angle of every pixel from the generated video alone, without cameras, depth, optical flow or a reference video (Fig.~\ref{fig:model}).

\paragraph{Architecture.} A clip $\{I_t\}_{t=1}^{24}$ of 24 frames enters VGGT-1B~\cite{wang2025vggt}, whose 24 alternating frame and global attention layer pairs stay frozen. In the last 8 pairs, LoRA~\cite{hu2021lora} adds a trainable low-rank branch of rank $16$ to the query--key--value, output-projection and both MLP layers: a frozen weight $W\in\mathbb{R}^{m\times n}$ becomes
\begin{equation} h=Wx+BAx,\qquad A\in\mathbb{R}^{16\times n},\; B\in\mathbb{R}^{m\times16}. \end{equation}
A DPT head~\cite{ranftl2021dpt}, initialised from VGGT's depth head, maps the features to one value $\hat{d}_t(\mathbf{u})$ per pixel and frame. NRG-PD has 36.85M trainable parameters.

\paragraph{Target and loss.} The target is the noise-subtracted squared NRG angle of Sec.~\ref{sec:noise}, in units of $0.6^\circ$:
\begin{equation} T_t(\mathbf{u})=\frac{\min(\theta_t(\mathbf{u}),3^\circ)^2-\hat\sigma^2(t)}{(0.6^\circ)^2}. \end{equation}
Over the labelled pixels $\Lambda$ of the non-keyframes, the model is trained with the squared error
\begin{equation} \mathcal{L}=\frac{1}{|\Lambda|}\sum_{(t,\mathbf{u})\in\Lambda}\big(\hat{d}_t(\mathbf{u})-T_t(\mathbf{u})\big)^2 \end{equation}
plus an ordinal auxiliary output that classifies the noise-subtracted angle into 7 ordered bins. Because the noise is subtracted in the target, the squared-error minimiser estimates the noise-free squared angle under the conditions of Supp.~A.4 and needs no $\hat\sigma^2$ at inference; the predicted NRG angle is
\begin{equation} \hat\theta_t(\mathbf{u})=0.6^\circ\sqrt{\max\big(\hat{d}_t(\mathbf{u}),0\big)}. \end{equation}

\paragraph{Training.} We train on 702 training twins with 24-frame windows of random start and temporal stride 1, 2 or 3, half of them reversed in time, using AdamW with a linear warm-up over the first 300 updates and a constant learning rate afterwards, for 3 seeds. 

\section{Experiments}\label{sec:exp}
\begin{figure*}[t]
\centering
\includegraphics[width=.8\textwidth]{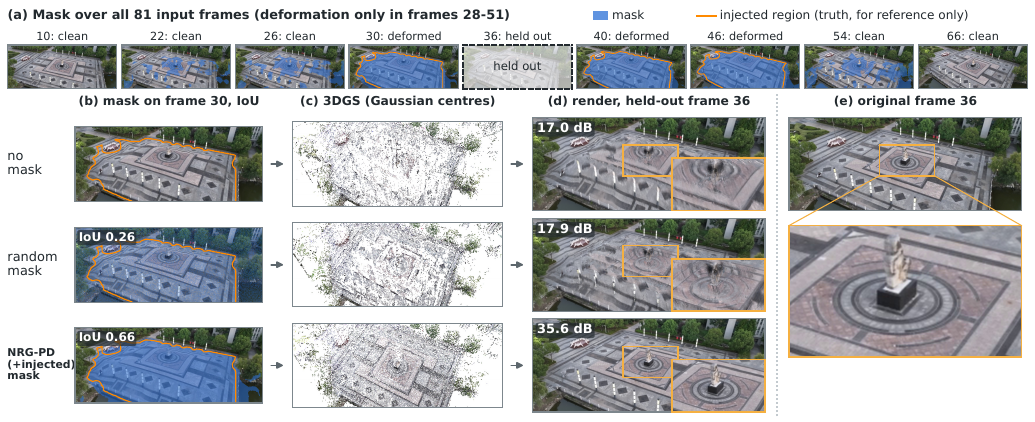}
\caption{Masked 3DGS on the best of the 17 clips warped in frames 28--51. (a) The mask of NRG-PD (+injected) (blue) over the 81 frames; orange: warped region, for reference only. (b)--(d) One row per mask (none, random, NRG-PD (+injected)): (b) the mask on training frame 30 and its IoU with the warped region; (c) centres of the fitted Gaussians (opacity $>0.5$) from a raised viewpoint; (d) render at held-out frame 36; dB: PSNR against the clean original (e) inside the warped region over held-out frames 28, 36 and 44.}
\label{fig:mask}
\end{figure*}
We evaluate NRG-PD and seven existing methods on the NRG-Twins test set (Sec.~\ref{sec:test}), test whether the predictions help 3D reconstruction of video with generated content (Sec.~\ref{sec:recon}), and test on prompt-generated video without labels whether the tracks the predictions flag fit a rigid reconstruction worse than the tracks they call consistent (Sec.~\ref{sec:track}). Detection of synthetic 2D deformation and the protocol of the human study are in Supp.~G and~H.

\subsection{Protocol}\label{sec:protocol}
The test set has 83 twins; each has 81 frames, keyframes are excluded, and the label is the noise-subtracted angle $\theta_c$. NRG pixels have $\theta_c\ge0.6^\circ$ and rigid pixels $\theta_c<0.075^\circ$; Supp.~A.4 derives both thresholds from the measurement noise. Pixels in between are left out of AUROC and average precision. With $s$ the score of a method, $\mathcal{N}$ the NRG pixels and $\mathcal{R}$ the rigid pixels, pooled over all clips, the main metric is
\begin{equation}\begin{aligned} \mathrm{AUROC}&=P\big(s(p)>s(q)\big)+\tfrac12P\big(s(p)=s(q)\big),\\ &\qquad p\in\mathcal{N},\;q\in\mathcal{R}, \end{aligned}\end{equation}
with 0.5 for chance. Kendall $\tau$ measures how often pixel pairs whose labels differ by more than twice the noise are ordered correctly (0 for chance, 1 for perfect). Average precision ranks the NRG and rigid pixels by score (chance 0.023). Precision and recall use the top 5\% scored pixels of each clip; NRG pixels are about 1.6\% of the labelled pixels, so precision is at most 0.31. The clip correlation is the Spearman correlation over clips between the 90th percentiles of score and label. Supp.~K gives the formulas. The angle error (MAE, in degrees) applies to NRG-PD, the only method that outputs degrees; the baselines are scored with the rank metrics. Intervals come from a bootstrap over whole clips. The baselines are GeCo and MEt3R (consistency) and RoMo, VGGT4D, MonST3R, Easi3R and MegaSaM (motion segmentation), each run once on the same twin input with official weights and settings; we use the continuous score before the method's own threshold, oriented so that higher means inconsistent, and map it back to the image grid (Supp.~C).

\subsection{NRG-PD on the test set}\label{sec:test}

Table~\ref{tab:s5.7} shows that NRG-PD outperforms all seven baselines in pixel-level ranking and retrieval, achieving 0.851 AUROC versus 0.735 for the strongest baseline and 3.3× higher average precision. All paired AUROC improvements are statistically positive. Clip-level ranking remains difficult for every method, while angular prediction errors increase with the ground-truth NRG angle.

\subsection{3D reconstruction of generated video}\label{sec:recon}
We test whether the predictions help 3DGS reconstruct video that contains generated content, with RobustSplat~\cite{fu2025robustsplat} and the same settings throughout; masked pixels or dropped frames are left out of the loss, and each mask is compared with an equal-size random mask and with no mask (Supp.~Table~6). RobustSplat's own learned mask leaves almost no pixel out of the loss and separates pixels with large and small NRG angles only weakly (AUROC 0.640, against 0.851 for NRG-PD; Supp.~J).

\begin{figure*}[t]
\centering
\includegraphics[width=0.8\textwidth]{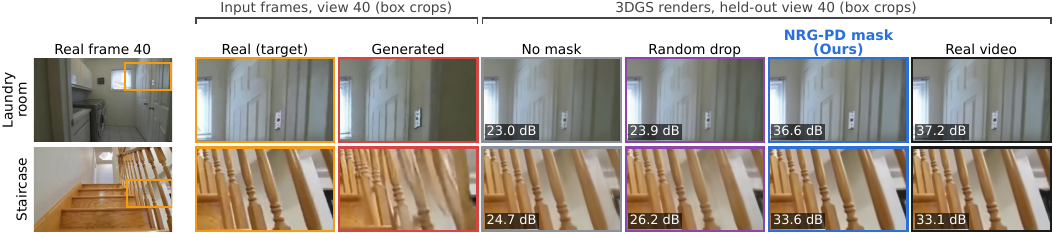}
\caption{Dropping the frames NRG-PD flags (Sec.~\ref{sec:recon}): the two of the 11 test clips with the largest gain over the better of no mask and random. Left: real frame 40 with the $280\times158$ box, searched at an 8-pixel stride, of largest summed error, taking at each pixel the larger error of the no-mask and NRG-PD renders. Right: the box in the real frame, in the generated input frame and in the renders at held-out view 40 (seed 0); dB: PSNR of the box against the real frame. All 11 clips are in the supplement.}
\label{fig:hyb}
\end{figure*}

\paragraph{Injected deformation.} We warp one region in 2D in real clips, either in frames 28--51 or in all frames (17 clips each), fit RobustSplat with 10 held-out frames and score the renders against the clean original inside the warped region. Since this deformation is artificial, the masks come from NRG-PD (+injected), the NRG-PD architecture trained also on injected clips (Supp.~G). With $\hat\theta_t(\mathbf{u})$ the predicted NRG angle of pixel $\mathbf{u}$ in frame $t$, the pixel is masked if it exceeds the median over the clip by a margin $\delta$:
\begin{equation} m_t(\mathbf{u})=\mathbf{1}\Big[\hat\theta_t(\mathbf{u})-\operatorname{median}_{t'}\hat\theta_{t'}(\mathbf{u})\ge\delta\Big],\qquad\delta=0.06^\circ, \end{equation}
so at most half of a pixel's frames are masked; $\delta$ is chosen on validation clips (Supp.~G). With deformation in frames 28--51, the masks beat random by $+3.07$\,dB (interval above 0, better on 82\% of clips) and recover about a third of the ground-truth-mask gain ($+8.54$\,dB). On the best clip (Fig.~\ref{fig:mask}) the masked fit renders deformed frames as well as clean ones (whole-frame PSNR 34.1 vs.\ 34.3\,dB), while the unmasked and random fits drop to 21--23\,dB. With deformation in every frame the rule masks little, because each pixel's own median is already high.

\begin{figure}[t]
\centering
\includegraphics[width=\linewidth]{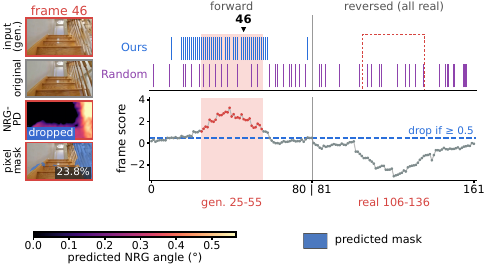}
\caption{Frames dropped in one test clip (fit seed 0). Left: frame 46 as generated (input) and in the original video, the NRG-PD prediction and the pixel mask (blue; corner: share of masked pixels). Right, top: frames dropped by our rule (blue) and by the random control (purple) in the 162-frame input, where the views of the generated frames 25--55 reappear as real frames 106--136; ours drops all 28 generated training frames and 15 of the 122 real ones, random 8 and 35. Right, bottom: the frame score $r_t/\sigma_r$; frames scoring $\geq0.5$ are dropped.}
\label{fig:hybdrop}
\end{figure}

\paragraph{Generated frames inside a real clip.} For 11 test clips $\times$ 2 seeds, we replace RealEstate10K frames 25--55 with Wan frames generated only from frames 24 and 56, then append the same real clip in reverse, so every generated view also appears as a real frame. Our rule drops whole frames and uses only the detector and the cameras. With $s_t$ the mean predicted NRG angle of frame $t$, $N(t)$ the 6 frames with the nearest cameras and
\begin{equation} r_t=s_t-\operatorname{median}_{u\in N(t)}s_u, \end{equation}
frame $t$ is dropped if $r_t\ge c\,\sigma_r$, where $\sigma_r$ is the robust standard deviation of $r$ over the clip and $c=0.5$ is chosen on 8 validation clips (Supp.~F, Fig.~\ref{fig:hybdrop}). Renders at held-out views are scored against the real frames. Dropping the frames NRG-PD flags beats random dropping by $+4.36$\,dB and no masking by $+3.40$\,dB, and closes a per-pair median of 0.70 of the gap between no masking and the fit on the real video (Supp.~Table~6, Fig.~\ref{fig:hyb}). The generated frames are wrong as whole views, since the generator's camera path departs from the real one, so the frame is the unit to mask; pixel masks from the same detector do not beat random here. Fig.~9 in Supp.~F shows the per-pixel predictions of all methods on the same clips.

\begin{table}[t]
\centering\footnotesize
\setlength{\tabcolsep}{4pt}
\begin{tabular*}{\linewidth}{@{\extracolsep{\fill}}lcc@{}}\toprule
Method & All tracks & Reliable tracks \\
\midrule
\textbf{NRG-PD (ours)} & \textbf{0.848 $\pm$ 0.025} & \textbf{0.914 $\pm$ 0.019} \\
RoMo & 0.811 $\pm$ 0.022 & 0.827 $\pm$ 0.024 \\
MonST3R & 0.800 $\pm$ 0.021 & 0.835 \\
Easi3R & 0.749 & 0.815 \\
GeCo & 0.749 & 0.747 \\
VGGT4D & 0.722 & 0.776 \\
MEt3R & 0.689 & 0.754 \\
MegaSaM (own 168 clips) & 0.568 & 0.630 \\
\midrule
\multicolumn{3}{@{}l}{\emph{40 labelled twins, tracks grouped by}}\\
Ground-truth labels & 0.857 $\pm$ 0.082 & --- \\
Shuffled labels & 0.498 & --- \\
\bottomrule
\end{tabular*}
\caption{Track AUROC (0.5 = chance). Top: prompt videos; 144 of the 192 clips are scored by all six original methods (NRG-PD and five baselines), RoMo on 143 of them; reliable tracks: forward--backward consistent tracks only. Bottom: the same test on 40 labelled twins with tracks grouped by their ground-truth labels ($\theta_c\ge0.6^\circ$ vs.\ $<0.075^\circ$) or by shuffled labels. $\pm$: half-width of the 95\% interval, clip-level bootstrap.}
\label{tab:s5.17tab}
\end{table}
\subsection{Prompt-generated video}\label{sec:track}
Prompt-generated videos lack ground-truth labels. We test whether tracks flagged as inconsistent have larger rigid-reconstruction errors than clean tracks. Per clip, each method labels its top and bottom 30\% scored tracks as inconsistent and clean. We estimate cameras from clean tracks and score all remaining tracks by their \underline{T}ime-\underline{N}ormalised \underline{90}th-percentile reprojection error (TN90; Supp.~I). We report TN90 AUROC between the two groups (0.5 = chance). TN90 uses no method score, and the scored tracks are not used to fit the cameras; a method enters only through its clean group, which fixes the cameras, and its two groups, which are compared.

We validate this test on 40 labelled twins: ground-truth track labels yield AUROC $0.857\pm0.082$, versus $0.498$ for shuffled labels (Table~\ref{tab:s5.17tab}). We then test 192 81-frame clips from 48 static-scene prompts (8 scenes $\times$ 3 camera motions $\times$ 2 motion scales) and four generators: LTX-2.5~\cite{hacohen2026ltx2}, CogVideoX1.5-5B~\cite{yang2025cogvideox}, Wan2.2-T2V-A14B~\cite{wan2025wan}, and HunyuanVideo-1.5~\cite{tencent2025hunyuanvideo15}.

NRG-PD achieves AUROC $0.848\pm0.025$, highest among eight methods (Table~\ref{tab:s5.17tab}). Paired gains are $+0.038\pm0.025$ over RoMo, $+0.048\pm0.029$ over MonST3R, and $0.10$–$0.16$ over four other baselines, all with intervals above zero; MegaSaM uses its own 168 clips. NRG-PD also leads on forward–backward consistent tracks ($0.914$). Shuffled labels yield chance-level AUROC ($0.50$), and NRG-PD gives $0.845$ when it reads the baselines' re-encoded inputs.

\section{Conclusion}
Generated video is not rigid, and no evaluator has been checked against per-pixel ground truth on generated video. Real--generated twins provide it: the NRG angle measures, per pixel and in degrees, how far a twin departs from a rigid scene. NRG-PD, trained on NRG-Twins, predicts this angle from RGB alone and ranks pixels better than seven existing methods (AUROC 0.85 vs.\ at most 0.74); on prompt-generated video the tracks it flags have larger reprojection error (track AUROC 0.85), and its predictions improve 3DGS of video with generated content. The twins need only real videos and a keyframe-conditioned generator, so the data can follow new generators.

\section*{LLM Usage Statement}
LLM-based assistants were used extensively in this work, in both the research and the writing. Tools: Claude Code (Anthropic; Claude Opus and Claude Haiku models) and Codex (OpenAI; GPT-6 family models). Uses: (i) writing, debugging and running code for the dataset pipeline, detector training, baselines, evaluation and figures, including launching and monitoring jobs on a compute cluster; (ii) summarising experiment outputs and checking them against the raw result files; (iii) drafting and editing the text of all sections from the authors' notes and results, including wording and grammar; (iv) searching for and summarising related work and preparing the bibliography; (v) building the web page used for the human study. Experimental designs and analyses were proposed by the authors or by the assistants and decided by the authors; the reported numbers were checked against the result files, and the authors take full responsibility for the content of the paper. The video generators studied in this paper (Secs.~3 and~5) are generative models used as objects of study, not as writing aids.

\bibliographystyle{ieeenat_fullname}
\bibliography{refs}

\begin{thebibliography}{64}
\providecommand{\natexlab}[1]{#1}
\providecommand{\url}[1]{\texttt{#1}}
\expandafter\ifx\csname urlstyle\endcsname\relax
  \providecommand{\doi}[1]{doi: #1}\else
  \providecommand{\doi}{doi: \begingroup \urlstyle{rm}\Url}\fi

\bibitem[Asim et~al.(2025)Asim, Wewer, Wimmer, Schiele, and
  Lenssen]{asim2025met3r}
Mohammad Asim, Christopher Wewer, Thomas Wimmer, Bernt Schiele, and Jan~Eric
  Lenssen.
\newblock {MEt3R}: Measuring multi-view consistency in generated images.
\newblock In \emph{Proceedings of the IEEE/CVF Conference on Computer Vision
  and Pattern Recognition (CVPR)}, pages 6034--6044, 2025.

\bibitem[Barath et~al.(2019)Barath, Matas, and Noskova]{barath2019magsac}
Daniel Barath, Jiri Matas, and Jana Noskova.
\newblock Magsac: Marginalizing sample consensus.
\newblock In \emph{Proceedings of the IEEE/CVF conference on computer vision
  and pattern recognition}, pages 10197--10205, 2019.

\bibitem[Baruch et~al.(2021)Baruch, Chen, Dehghan, Dimry, Feigin, Fu, Gebauer,
  Joffe, Kurz, Schwartz, et~al.]{baruch2021arkitscenes}
Gilad Baruch, Zhuoyuan Chen, Afshin Dehghan, Tal Dimry, Yuri Feigin, Peter Fu,
  Thomas Gebauer, Brandon Joffe, Daniel Kurz, Arik Schwartz, et~al.
\newblock {ARKitScenes}: A diverse real-world dataset for {3D} indoor scene
  understanding using mobile {RGB-D} data.
\newblock In \emph{Advances in Neural Information Processing Systems (NeurIPS)
  Datasets and Benchmarks Track}, 2021.

\bibitem[Caron et~al.(2021)Caron, Touvron, Misra, J{\'e}gou, Mairal,
  Bojanowski, and Joulin]{caron2021emerging}
Mathilde Caron, Hugo Touvron, Ishan Misra, Herv{\'e} J{\'e}gou, Julien Mairal,
  Piotr Bojanowski, and Armand Joulin.
\newblock Emerging properties in self-supervised vision transformers.
\newblock In \emph{Proceedings of the IEEE/CVF International Conference on
  Computer Vision (ICCV)}, 2021.

\bibitem[Chang et~al.(2025)Chang, Liu, Liu, Lyu, Huang, Tao, Wan, Zhang, and
  Qi]{chang2025far}
Chirui Chang, Jiahui Liu, Zhengzhe Liu, Xiaoyang Lyu, Yi-Hua Huang, Xin Tao,
  Pengfei Wan, Di Zhang, and Xiaojuan Qi.
\newblock How far are ai-generated videos from simulating the 3d visual world:
  A learned 3d evaluation approach.
\newblock In \emph{Proceedings of the IEEE/CVF International Conference on
  Computer Vision (ICCV)}, pages 10307--10317, 2025.

\bibitem[Charatan et~al.(2024)Charatan, Li, Tagliasacchi, and
  Sitzmann]{charatan2024pixelsplat}
David Charatan, Sizhe~Lester Li, Andrea Tagliasacchi, and Vincent Sitzmann.
\newblock {pixelSplat}: 3d gaussian splats from image pairs for scalable
  generalizable 3d reconstruction.
\newblock In \emph{Proceedings of the IEEE/CVF Conference on Computer Vision
  and Pattern Recognition (CVPR)}, pages 19457--19467, 2024.

\bibitem[Chen et~al.(2025)Chen, Chen, Xiu, Geiger, and Chen]{chen2025easi3r}
Xingyu Chen, Yue Chen, Yuliang Xiu, Andreas Geiger, and Anpei Chen.
\newblock {Easi3R}: Estimating disentangled motion from {DUSt3R} without
  training.
\newblock In \emph{Proceedings of the IEEE/CVF International Conference on
  Computer Vision (ICCV)}, pages 9158--9168, 2025.

\bibitem[Deitke et~al.(2023)Deitke, Liu, Wallingford, Ngo, Michel, Kusupati,
  Fan, Laforte, Voleti, Gadre, et~al.]{deitke2023objaverse}
Matt Deitke, Ruoshi Liu, Matthew Wallingford, Huong Ngo, Oscar Michel, Aditya
  Kusupati, Alan Fan, Christian Laforte, Vikram Voleti, Samir~Yitzhak Gadre,
  et~al.
\newblock {Objaverse-XL}: A universe of 10m+ 3d objects.
\newblock In \emph{Advances in Neural Information Processing Systems (NeurIPS),
  Datasets and Benchmarks Track}, 2023.

\bibitem[Dou et~al.(2026)Dou, Zheng, Chen, Zheng, Zhou, and
  Lu]{dou2026measuring}
Weijia Dou, Wenzhao Zheng, Weiliang Chen, Yu Zheng, Jie Zhou, and Jiwen Lu.
\newblock Measuring 3d spatial geometric consistency in dynamic generated
  videos.
\newblock \emph{Proceedings of the IEEE/CVF European Conference on Computer
  Vision (ECCV)}, 2026.

\bibitem[Du et~al.(2026)Du, Ye, Cong, Li, Ni, Agarwal, Zhou, Li, Balestriero,
  and Wang]{du2026videogpa}
Hongyang Du, Junjie Ye, Xiaoyan Cong, Runhao Li, Jingcheng Ni, Aman Agarwal,
  Zeqi Zhou, Zekun Li, Randall Balestriero, and Yue Wang.
\newblock {VideoGPA}: Distilling geometry priors for 3d-consistent video
  generation.
\newblock \emph{International Conference on Machine Learning (ICML)}, 2026.

\bibitem[Duan et~al.(2025)Duan, Yu, Chen, Fei-Fei, and Wu]{duan2025worldscore}
Haoyi Duan, Hong-Xing Yu, Sirui Chen, Li Fei-Fei, and Jiajun Wu.
\newblock {WorldScore}: A unified evaluation benchmark for world generation.
\newblock In \emph{Proceedings of the IEEE/CVF International Conference on
  Computer Vision (ICCV)}, pages 27713--27724, 2025.

\bibitem[Efron(2010)]{efron2010large}
Bradley Efron.
\newblock \emph{Large-scale inference: empirical Bayes methods for estimation,
  testing, and prediction}.
\newblock Cambridge University Press, 2010.

\bibitem[Fu et~al.(2025)Fu, Zhang, Yao, Chen, Xiong, Huang, Cui, and
  Cao]{fu2025robustsplat}
Chuanyu Fu, Yuqi Zhang, Kunbin Yao, Guanying Chen, Yuan Xiong, Chuan Huang,
  Shuguang Cui, and Xiaochun Cao.
\newblock {RobustSplat}: Decoupling densification and dynamics for
  transient-free 3dgs.
\newblock In \emph{Proceedings of the IEEE/CVF International Conference on
  Computer Vision (ICCV)}, pages 27126--27136, 2025.

\bibitem[Fu et~al.(2024)Fu, Hamilton, Brandt, Feldmann, Zhang, and
  Freeman]{fu2024featup}
Stephanie Fu, Mark Hamilton, Laura~E Brandt, Axel Feldmann, Zhoutong Zhang, and
  William Freeman.
\newblock {FeatUp}: A model-agnostic framework for features at any resolution.
\newblock In \emph{International Conference on Learning Representations
  (ICLR)}, 2024.

\bibitem[Gao et~al.(2024)Gao, Holynski, Henzler, Brussee, Martin-Brualla,
  Srinivasan, Barron, and Poole]{gao2024cat3d}
Ruiqi Gao, Aleksander Holynski, Philipp Henzler, Arthur Brussee, Ricardo
  Martin-Brualla, Pratul Srinivasan, Jonathan~T Barron, and Ben Poole.
\newblock {CAT3D}: Create anything in 3d with multi-view diffusion models.
\newblock \emph{Advances in Neural Information Processing Systems(NeurIPS)},
  2024.

\bibitem[Goli et~al.(2025)Goli, Sabour, Matthews, Brubaker, Lagun, Jacobson,
  Fleet, Saxena, and Tagliasacchi]{goli2025romo}
Lily Goli, Sara Sabour, Mark Matthews, Marcus~A Brubaker, Dmitry Lagun, Alec
  Jacobson, David~J Fleet, Saurabh Saxena, and Andrea Tagliasacchi.
\newblock {RoMo}: Robust motion segmentation improves structure from motion.
\newblock In \emph{Proceedings of the IEEE/CVF International Conference on
  Computer Vision (ICCV)}, pages 6155--6164, 2025.

\bibitem[Gu et~al.(2026)Gu, Hur, Herrmann, Zhan, Zickler, Sun, and
  Pfister]{gu2025geco}
Leslie Gu, Junhwa Hur, Charles Herrmann, Fangneng Zhan, Todd Zickler, Deqing
  Sun, and Hanspeter Pfister.
\newblock {GeCo}: Evaluating geometric consistency for video generation via
  motion and structure.
\newblock \emph{Proceedings of the IEEE/CVF European Conference on Computer
  Vision (ECCV)}, 2026.

\bibitem[HaCohen et~al.(2026)HaCohen, Brazowski, Chiprut, Bitterman, Kvochko,
  Berkowitz, Shalem, Lifschitz, Moshe, Porat, et~al.]{hacohen2026ltx2}
Yoav HaCohen, Benny Brazowski, Nisan Chiprut, Yaki Bitterman, Andrew Kvochko,
  Avishai Berkowitz, Daniel Shalem, Daphna Lifschitz, Dudu Moshe, Eitan Porat,
  et~al.
\newblock {LTX-2}: Efficient joint audio-visual foundation model.
\newblock \emph{arXiv preprint arXiv:2601.03233}, 2026.

\bibitem[Hao et~al.(2025)Hao, Wang, Wang, Zhang, and
  Guo]{hao2025gaussvideodreamer}
Junlin Hao, Peiheng Wang, Haoyang Wang, Xinggong Zhang, and Zongming Guo.
\newblock {GaussVideoDreamer}: 3d scene generation with video diffusion and
  inconsistency-aware gaussian splatting.
\newblock \emph{arXiv preprint arXiv:2504.10001}, 2025.

\bibitem[Hartley and Zisserman(2004)]{hartley2004multiple}
Richard Hartley and Andrew Zisserman.
\newblock \emph{Multiple View Geometry in Computer Vision}.
\newblock Cambridge University Press, Cambridge, 2nd edition, 2004.

\bibitem[H{\"o}llein and Nie{\ss}ner(2026)]{hollein2026world}
Lukas H{\"o}llein and Matthias Nie{\ss}ner.
\newblock World reconstruction from inconsistent views.
\newblock In \emph{European Conference on Computer Vision (ECCV)}, 2026.

\bibitem[Hu et~al.(2022)Hu, Shen, Wallis, Allen-Zhu, Li, Wang, Wang, and
  Chen]{hu2021lora}
Edward~J Hu, Yelong Shen, Phillip Wallis, Zeyuan Allen-Zhu, Yuanzhi Li, Shean
  Wang, Lu Wang, and Weizhu Chen.
\newblock {LoRA}: Low-rank adaptation of large language models.
\newblock \emph{International Conference on Learning Representations (ICLR)},
  2022.

\bibitem[Hu et~al.(2024)Hu, Yin, Zhang, Cai, Long, Chen, Wang, Yu, Shen, and
  Shen]{hu2024metric3dv2}
Mu Hu, Wei Yin, Chi Zhang, Zhipeng Cai, Xiaoxiao Long, Hao Chen, Kaixuan Wang,
  Gang Yu, Chunhua Shen, and Shaojie Shen.
\newblock {Metric3D v2}: A versatile monocular geometric foundation model for
  zero-shot metric depth and surface normal estimation.
\newblock \emph{IEEE Transactions on Pattern Analysis and Machine
  Intelligence}, 46\penalty0 (12):\penalty0 10579--10596, 2024.

\bibitem[Hu et~al.(2026)Hu, Cheng, Yu, Guo, and Wang]{hu2026vggt4d}
Yu Hu, Chong Cheng, Sicheng Yu, Xiaoyang Guo, and Hao Wang.
\newblock {VGGT4D}: Mining motion cues in visual geometry transformers for 4d
  scene reconstruction.
\newblock In \emph{Proceedings of the IEEE/CVF Conference on Computer Vision
  and Pattern Recognition (CVPR) Findings}, 2026.

\bibitem[Huang et~al.(2024)Huang, He, Yu, Zhang, Si, Jiang, Zhang, Wu, Jin,
  Chanpaisit, Wang, Chen, Wang, Lin, Qiao, and Liu]{huang2024vbench}
Ziqi Huang, Yinan He, Jiashuo Yu, Fan Zhang, Chenyang Si, Yuming Jiang, Yuanhan
  Zhang, Tianxing Wu, Qingyang Jin, Nattapol Chanpaisit, Yaohui Wang, Xinyuan
  Chen, Limin Wang, Dahua Lin, Yu Qiao, and Ziwei Liu.
\newblock {VBench}: Comprehensive benchmark suite for video generative models.
\newblock In \emph{Proceedings of the IEEE/CVF Conference on Computer Vision
  and Pattern Recognition (CVPR)}, pages 21807--21818, 2024.

\bibitem[Jung et~al.(2026)Jung, Zholus, Sundermeyer, Doersch, Goroshin, Tan,
  Chandar, Triebel, and Tombari]{jung2026tapnextpp}
Sebastian Jung, Artem Zholus, Martin Sundermeyer, Carl Doersch, Ross Goroshin,
  David~Joseph Tan, Sarath Chandar, Rudolph Triebel, and Federico Tombari.
\newblock {TAPNext++}: What's next for tracking any point ({TAP})?
\newblock \emph{Proceedings of the IEEE/CVF Conference on Computer Vision and
  Pattern Recognition (CVPR) Findings}, pages 8429--8438, 2026.

\bibitem[Karaev et~al.(2025)Karaev, Makarov, Wang, Neverova, Vedaldi, and
  Rupprecht]{karaev2024cotracker3}
Nikita Karaev, Yuri Makarov, Jianyuan Wang, Natalia Neverova, Andrea Vedaldi,
  and Christian Rupprecht.
\newblock {CoTracker3}: Simpler and better point tracking by pseudo-labeling
  real videos.
\newblock In \emph{2025 IEEE/CVF International Conference on Computer Vision
  (ICCV)}. IEEE, 2025.

\bibitem[Keetha et~al.(2026)Keetha, M{\"u}ller, Sch{\"o}nberger, Porzi, Zhang,
  Fischer, Knapitsch, Zauss, Weber, Antunes, et~al.]{keetha2026mapanything}
Nikhil Keetha, Norman M{\"u}ller, Johannes Sch{\"o}nberger, Lorenzo Porzi,
  Yuchen Zhang, Tobias Fischer, Arno Knapitsch, Duncan Zauss, Ethan Weber,
  Nelson Antunes, et~al.
\newblock {MapAnything}: Universal feed-forward metric 3d reconstruction.
\newblock In \emph{International Conference on 3D Vision (3DV)}, 2026.

\bibitem[Kerbl et~al.(2023)Kerbl, Kopanas, Leimk{\"u}hler, and
  Drettakis]{kerbl20233d}
Bernhard Kerbl, Georgios Kopanas, Thomas Leimk{\"u}hler, and George Drettakis.
\newblock 3d gaussian splatting for real-time radiance field rendering.
\newblock \emph{ACM Transactions on Graphics}, 42\penalty0 (4), 2023.

\bibitem[Kroeger et~al.(2016)Kroeger, Timofte, Dai, and
  Van~Gool]{kroeger2016fast}
Till Kroeger, Radu Timofte, Dengxin Dai, and Luc Van~Gool.
\newblock Fast optical flow using dense inverse search.
\newblock In \emph{European Conference on Computer Vision (ECCV)}, pages
  471--488, 2016.

\bibitem[Kulhanek et~al.(2024)Kulhanek, Peng, Kukelova, Pollefeys, and
  Sattler]{kulhanek2024wildgaussians}
Jonas Kulhanek, Songyou Peng, Zuzana Kukelova, Marc Pollefeys, and Torsten
  Sattler.
\newblock {WildGaussians}: 3d gaussian splatting in the wild.
\newblock In \emph{Advances in Neural Information Processing Systems
  (NeurIPS)}, 2024.

\bibitem[Leroy et~al.(2024)Leroy, Cabon, and Revaud]{leroy2024grounding}
Vincent Leroy, Yohann Cabon, and J{\'e}r{\^o}me Revaud.
\newblock Grounding image matching in 3d with {MASt3R}.
\newblock In \emph{European Conference on Computer Vision (ECCV)}, 2024.

\bibitem[Li et~al.(2025)Li, Tucker, Cole, Wang, Jin, Ye, Kanazawa, Holynski,
  and Snavely]{li2025megasam}
Zhengqi Li, Richard Tucker, Forrester Cole, Qianqian Wang, Linyi Jin, Vickie
  Ye, Angjoo Kanazawa, Aleksander Holynski, and Noah Snavely.
\newblock {MegaSaM}: Accurate, fast, and robust structure and motion from
  casual dynamic videos.
\newblock In \emph{Proceedings of the IEEE/CVF Conference on Computer Vision
  and Pattern Recognition (CVPR)}, pages 10486--10496, 2025.

\bibitem[Lin et~al.(2026)Lin, Chen, Liew, Chen, Li, Shi, Feng, and
  Kang]{lin2025depthanything3}
Haotong Lin, Sili Chen, Junhao Liew, Donny~Y Chen, Zhenyu Li, Guang Shi, Jiashi
  Feng, and Bingyi Kang.
\newblock {Depth Anything 3}: Recovering the visual space from any views.
\newblock \emph{The International Conference on Learning Representations
  (ICLR)}, 2026.

\bibitem[Lin et~al.(2025)Lin, Peng, Zi, Gao, Qi, Ma, and
  Jiang]{lin2025brokenvideos}
Jiahao Lin, Weixuan Peng, Bojia Zi, Yifeng Gao, Xianbiao Qi, Xingjun Ma, and
  Yu-Gang Jiang.
\newblock {BrokenVideos}: A benchmark dataset for fine-grained artifact
  localization in {AI}-generated videos.
\newblock \emph{arXiv preprint arXiv:2506.20103}, 2025.

\bibitem[Ling et~al.(2024)Ling, Sheng, Tu, Zhao, Xin, Wan, Yu, Guo, Yu, Lu,
  et~al.]{ling2024dl3dv}
Lu Ling, Yichen Sheng, Zhi Tu, Wentian Zhao, Cheng Xin, Kun Wan, Lantao Yu,
  Qianyu Guo, Zixun Yu, Yawen Lu, et~al.
\newblock {DL3DV-10K}: A large-scale scene dataset for deep learning-based 3d
  vision.
\newblock In \emph{Proceedings of the IEEE/CVF Conference on Computer Vision
  and Pattern Recognition (CVPR)}, pages 22160--22169, 2024.

\bibitem[Lowe(2004)]{lowe2004distinctive}
David~G Lowe.
\newblock Distinctive image features from scale-invariant keypoints.
\newblock \emph{International journal of computer vision}, 60\penalty0
  (2):\penalty0 91--110, 2004.

\bibitem[Lyu et~al.(2026)Lyu, Shen, Li, Wang, Wang, Yuan, Zhao, and
  Yang]{lyu2026learning}
Weijie Lyu, Tiancheng Shen, Jason Li, Yujing Wang, Xiaoyuan Wang, Haobo Yuan,
  Yizhou Zhao, and Ming-Hsuan Yang.
\newblock Learning to correct geometry in generated videos.
\newblock In \emph{Advances in Neural Information Processing Systems
  (NeurIPS)}, 2026.

\bibitem[Martin-Brualla et~al.(2021)Martin-Brualla, Radwan, Sajjadi, Barron,
  Dosovitskiy, and Duckworth]{martin2021nerf}
Ricardo Martin-Brualla, Noha Radwan, Mehdi~SM Sajjadi, Jonathan~T Barron,
  Alexey Dosovitskiy, and Daniel Duckworth.
\newblock {NeRF} in the wild: Neural radiance fields for unconstrained photo
  collections.
\newblock In \emph{Proceedings of the IEEE/CVF Conference on Computer Vision
  and Pattern Recognition (CVPR)}, 2021.

\bibitem[Piccinelli et~al.(2025)Piccinelli, Sakaridis, Yang, Segu, Li,
  Abbeloos, and Van~Gool]{piccinelli2025unidepthv2}
Luigi Piccinelli, Christos Sakaridis, Yung-Hsu Yang, Mattia Segu, Siyuan Li,
  Wim Abbeloos, and Luc Van~Gool.
\newblock {UniDepthV2}: Universal monocular metric depth estimation made
  simpler.
\newblock \emph{IEEE Transactions on Pattern Analysis and Machine Intelligence
  (TPAMI)}, 2025.

\bibitem[Ranftl et~al.(2021)Ranftl, Bochkovskiy, and Koltun]{ranftl2021dpt}
Ren{\'e} Ranftl, Alexey Bochkovskiy, and Vladlen Koltun.
\newblock Vision transformers for dense prediction.
\newblock In \emph{2021 IEEE/CVF International Conference on Computer Vision
  (ICCV)}, pages 12159--12168. IEEE, 2021.

\bibitem[Ravi et~al.(2025)Ravi, Gabeur, Hu, Hu, Ryali, Ma, Khedr, R{\"a}dle,
  Rolland, Gustafson, et~al.]{ravi2025sam}
Nikhila Ravi, Valentin Gabeur, Yuan-Ting Hu, Ronghang Hu, Chaitanya Ryali,
  Tengyu Ma, Haitham Khedr, Roman R{\"a}dle, Chloe Rolland, Laura Gustafson,
  et~al.
\newblock {SAM} 2: Segment anything in images and videos.
\newblock In \emph{International Conference on Learning Representations
  (ICLR)}, 2025.

\bibitem[Sabour et~al.(2023)Sabour, Vora, Duckworth, Krasin, Fleet, and
  Tagliasacchi]{sabour2023robustnerf}
Sara Sabour, Suhani Vora, Daniel Duckworth, Ivan Krasin, David~J Fleet, and
  Andrea Tagliasacchi.
\newblock {RobustNeRF}: Ignoring distractors with robust losses.
\newblock In \emph{Proceedings of the IEEE/CVF Conference on Computer Vision
  and Pattern Recognition (CVPR)}, pages 20626--20636, 2023.

\bibitem[Sabour et~al.(2025)Sabour, Goli, Kopanas, Matthews, Lagun, Guibas,
  Jacobson, Fleet, and Tagliasacchi]{sabour2025spotlesssplats}
Sara Sabour, Lily Goli, George Kopanas, Mark Matthews, Dmitry Lagun, Leonidas
  Guibas, Alec Jacobson, David Fleet, and Andrea Tagliasacchi.
\newblock {SpotLessSplats}: Ignoring distractors in 3d gaussian splatting.
\newblock \emph{ACM Transactions on Graphics}, 2025.

\bibitem[Schonberger and Frahm(2016)]{schonberger2016structure}
Johannes~L Schonberger and Jan-Michael Frahm.
\newblock Structure-from-motion revisited.
\newblock In \emph{Proceedings of the IEEE Conference on Computer Vision and
  Pattern Recognition (CVPR)}, pages 4104--4113, 2016.

\bibitem[Sch{\"o}nberger et~al.(2016)Sch{\"o}nberger, Zheng, Frahm, and
  Pollefeys]{schonberger2016pixelwise}
Johannes~L Sch{\"o}nberger, Enliang Zheng, Jan-Michael Frahm, and Marc
  Pollefeys.
\newblock Pixelwise view selection for unstructured multi-view stereo.
\newblock In \emph{European Conference on Computer Vision (ECCV)}, pages
  501--518, 2016.

\bibitem[Sturm et~al.(2012)Sturm, Engelhard, Endres, Burgard, and
  Cremers]{sturm2012benchmark}
J{\"u}rgen Sturm, Nikolas Engelhard, Felix Endres, Wolfram Burgard, and Daniel
  Cremers.
\newblock A benchmark for the evaluation of rgb-d slam systems.
\newblock In \emph{2012 IEEE/RSJ international conference on intelligent robots
  and systems}, pages 573--580. IEEE, 2012.

\bibitem[{Team Wan} et~al.(2025){Team Wan}, Wang, Ai, Wen, Mao, Xie, Chen, Yu,
  Zhao, Yang, Zeng, Wang, Zhang, Zhou, Wang, Chen, Zhu, Zhao, Yan, Huang, Feng,
  Zhang, Li, Wu, Chu, Feng, Zhang, Sun, Fang, Wang, Gui, Weng, Shen, Lin, Wang,
  Wang, Zhou, Wang, Shen, Yu, Shi, Huang, Xu, Kou, Lv, Li, Liu, Wang, Zhang,
  Huang, Li, Wu, Liu, Pan, Zheng, Hong, Shi, Feng, Jiang, Han, Wu, and
  Liu]{wan2025wan}
{Team Wan}, Ang Wang, Baole Ai, Bin Wen, Chaojie Mao, Chen-Wei Xie, Di Chen,
  Feiwu Yu, Haiming Zhao, Jianxiao Yang, Jianyuan Zeng, Jiayu Wang, Jingfeng
  Zhang, Jingren Zhou, Jinkai Wang, Jixuan Chen, Kai Zhu, Kang Zhao, Keyu Yan,
  Lianghua Huang, Mengyang Feng, Ningyi Zhang, Pandeng Li, Pingyu Wu, Ruihang
  Chu, Ruili Feng, Shiwei Zhang, Siyang Sun, Tao Fang, Tianxing Wang, Tianyi
  Gui, Tingyu Weng, Tong Shen, Wei Lin, Wei Wang, Wei Wang, Wenmeng Zhou, Wente
  Wang, Wenting Shen, Wenyuan Yu, Xianzhong Shi, Xiaoming Huang, Xin Xu, Yan
  Kou, Yangyu Lv, Yifei Li, Yijing Liu, Yiming Wang, Yingya Zhang, Yitong
  Huang, Yong Li, You Wu, Yu Liu, Yulin Pan, Yun Zheng, Yuntao Hong, Yupeng
  Shi, Yutong Feng, Zeyinzi Jiang, Zhen Han, Zhi-Fan Wu, and Ziyu Liu.
\newblock {Wan}: Open and advanced large-scale video generative models.
\newblock \emph{arXiv preprint arXiv:2503.20314}, 2025.

\bibitem[Teed and Deng(2020)]{teed2020raft}
Zachary Teed and Jia Deng.
\newblock {RAFT}: Recurrent all-pairs field transforms for optical flow.
\newblock In \emph{European Conference on Computer Vision (ECCV)}, pages
  402--419, 2020.

\bibitem[Voleti et~al.(2024)Voleti, Yao, Boss, Letts, Pankratz, Tochilkin,
  Laforte, Rombach, and Jampani]{voleti2024sv3d}
Vikram Voleti, Chun-Han Yao, Mark Boss, Adam Letts, David Pankratz, Dmitry
  Tochilkin, Christian Laforte, Robin Rombach, and Varun Jampani.
\newblock {SV3D}: Novel multi-view synthesis and 3d generation from a single
  image using latent video diffusion.
\newblock In \emph{European Conference on Computer Vision (ECCV)}, 2024.

\bibitem[Wahba(1965)]{wahba1965least}
Grace Wahba.
\newblock A least squares estimate of satellite attitude.
\newblock \emph{SIAM review}, 7\penalty0 (3):\penalty0 409, 1965.

\bibitem[Wang et~al.(2025)Wang, Chen, Karaev, Vedaldi, Rupprecht, and
  Novotny]{wang2025vggt}
Jianyuan Wang, Minghao Chen, Nikita Karaev, Andrea Vedaldi, Christian
  Rupprecht, and David Novotny.
\newblock {VGGT}: Visual geometry grounded transformer.
\newblock In \emph{Proceedings of the IEEE/CVF Conference on Computer Vision
  and Pattern Recognition (CVPR)}, pages 5294--5306, 2025.

\bibitem[Wang et~al.(2026{\natexlab{a}})Wang, Zheng, Xu, and Liu]{wang2026rigi}
Jiacheng Wang, Zhedong Zheng, Wei Xu, and Ping Liu.
\newblock {RIGI}: Rectifying image-to-3d generation inconsistency via
  uncertainty-aware learning.
\newblock \emph{in IEEE Transactions on Image Processing, vol. 35, pp.
  6900-6914}, 2026{\natexlab{a}}.

\bibitem[Wang et~al.(2026{\natexlab{b}})Wang, Xu, Dong, Deng, Xiang, Lv, Sun,
  Tong, and Yang]{wang2025moge2}
Ruicheng Wang, Sicheng Xu, Yue Dong, Yu Deng, Jianfeng Xiang, Zelong Lv,
  Guangzhong Sun, Xin Tong, and Jiaolong Yang.
\newblock {MoGe-2}: Accurate monocular geometry with metric scale and sharp
  details.
\newblock \emph{Advances in Neural Information Processing Systems},
  38:\penalty0 35928--35959, 2026{\natexlab{b}}.

\bibitem[Wang et~al.(2024)Wang, Lipson, and Deng]{wang2024sea}
Yihan Wang, Lahav Lipson, and Jia Deng.
\newblock {SEA-RAFT}: Simple, efficient, accurate {RAFT} for optical flow.
\newblock In \emph{European Conference on Computer Vision (ECCV)}, pages
  36--54, 2024.

\bibitem[Wu et~al.(2025)Wu, Zou, Li, Huang, Yang, Tan, Peng, Wu, Xiong, Jiang,
  et~al.]{tencent2025hunyuanvideo15}
Bing Wu, Chang Zou, Changlin Li, Duojun Huang, Fang Yang, Hao Tan, Jack Peng,
  Jianbing Wu, Jiangfeng Xiong, Jie Jiang, et~al.
\newblock {HunyuanVideo 1.5} technical report.
\newblock \emph{arXiv preprint arXiv:2511.18870}, 2025.

\bibitem[Xiao et~al.(2025)Xiao, Lin, Lu, Zhai, Jiang, Zhao, Zhang, Jiang, Wang,
  and Liu]{xiao2025vdegaussian}
Yuru Xiao, Zihan Lin, Chao Lu, Deming Zhai, Kui Jiang, Wenbo Zhao, Wei Zhang,
  Junjun Jiang, Huanran Wang, and Xianming Liu.
\newblock {VDEGaussian}: Video diffusion enhanced 4d gaussian splatting for
  dynamic urban scenes modeling.
\newblock \emph{arXiv preprint arXiv:2508.02129}, 2025.

\bibitem[Yang et~al.(2024)Yang, Kang, Huang, Zhao, Xu, Feng, and
  Zhao]{yang2024depthanythingv2}
Lihe Yang, Bingyi Kang, Zilong Huang, Zhen Zhao, Xiaogang Xu, Jiashi Feng, and
  Hengshuang Zhao.
\newblock {Depth Anything V2}.
\newblock \emph{Advances in Neural Information Processing Systems},
  37:\penalty0 21875--21911, 2024.

\bibitem[Yang et~al.(2025)Yang, Teng, Zheng, Ding, Huang, Xu, Yang, Hong,
  Zhang, Feng, et~al.]{yang2025cogvideox}
Zhuoyi Yang, Jiayan Teng, Wendi Zheng, Ming Ding, Shiyu Huang, Jiazheng Xu,
  Yuanming Yang, Wenyi Hong, Xiaohan Zhang, Guanyu Feng, et~al.
\newblock {CogVideoX}: Text-to-video diffusion models with an expert
  transformer.
\newblock In \emph{International Conference on Learning Representations
  (ICLR)}, 2025.

\bibitem[Yeshwanth et~al.(2023)Yeshwanth, Liu, Nie{\ss}ner, and
  Dai]{yeshwanth2023scannet++}
Chandan Yeshwanth, Yueh-Cheng Liu, Matthias Nie{\ss}ner, and Angela Dai.
\newblock {ScanNet++}: A high-fidelity dataset of 3d indoor scenes.
\newblock In \emph{Proceedings of the IEEE/CVF International Conference on
  Computer Vision (ICCV)}, pages 12--22, 2023.

\bibitem[Zhang et~al.(2025{\natexlab{a}})Zhang, Herrmann, Hur, Jampani,
  Darrell, Cole, Sun, and Yang]{zhang2025monst3r}
Junyi Zhang, Charles Herrmann, Junhwa Hur, Varun Jampani, Trevor Darrell,
  Forrester Cole, Deqing Sun, and Ming-Hsuan Yang.
\newblock {MonST3R}: A simple approach for estimating geometry in the presence
  of motion.
\newblock In \emph{International Conference on Learning Representations
  (ICLR)}, 2025{\natexlab{a}}.

\bibitem[Zhang et~al.(2025{\natexlab{b}})Zhang, Keetha, Lyu, Jhamb, Chen, Qiu,
  Karhade, Jha, Hu, Ramanan, Scherer, and Wang]{zhang2025ufm}
Yuchen Zhang, Nikhil Keetha, Chenwei Lyu, Bhuvan Jhamb, Yutian Chen, Yuheng
  Qiu, Jay Karhade, Shreyas Jha, Yaoyu Hu, Deva Ramanan, Sebastian Scherer, and
  Wenshan Wang.
\newblock {UFM}: A simple path towards unified dense correspondence with flow.
\newblock \emph{Conference on Neural Information Processing Systems (NeurIPS)},
  2025{\natexlab{b}}.

\bibitem[Zheng et~al.(2025)Zheng, Huang, Liu, Zou, He, Zhang, Gu, Zhang, He,
  Zheng, et~al.]{zheng2025vbench}
Dian Zheng, Ziqi Huang, Hongbo Liu, Kai Zou, Yinan He, Fan Zhang, Lulu Gu,
  Yuanhan Zhang, Jingwen He, Wei-Shi Zheng, et~al.
\newblock {VBench-2.0}: Advancing video generation benchmark suite for
  intrinsic faithfulness.
\newblock \emph{arXiv preprint}, 2025.

\bibitem[Zhou et~al.(2018)Zhou, Tucker, Flynn, Fyffe, and
  Snavely]{zhou2018stereo}
Tinghui Zhou, Richard Tucker, John Flynn, Graham Fyffe, and Noah Snavely.
\newblock Stereo magnification: Learning view synthesis using multiplane
  images.
\newblock \emph{ACM Transactions on Graphics (Proc. SIGGRAPH 2018)}, 2018.

\end{thebibliography}

\clearpage
\appendix
\renewcommand{\floatpagefraction}{0.6}\renewcommand{\dblfloatpagefraction}{0.6}
\begin{center}{\Large\bf Supplementary Material}\end{center}
\vspace{6pt}
\section{Ground-truth label construction}
\subsection{Keyframe insertion: alignment, threshold and number of insertions}\label{app:keyframes}
\begin{itemize}
\item \textbf{Alignment.} Let $R^{r}_t$ and $R^{w}_t$ be the camera rotations of frame $t$ in the real clip and in the twin (Algorithm~\ref{alg:twin}), and $\mathcal{K}$ the current keyframes. The rotation $Q$ that best maps the twin cameras onto the real ones over the keyframes follows from a singular value decomposition:
\begin{equation}
\begin{aligned}
M&=\sum_{k\in\mathcal{K}}R^{r}_k\,(R^{w}_k)^{\top}=U S V^{\top},\\
Q&=U\operatorname{diag}\big(1,1,\det(UV^{\top})\big)V^{\top},
\end{aligned}
\end{equation}
where $U$ and $V$ are orthogonal, $S$ is diagonal, and the determinant keeps $Q$ a proper rotation. One keyframe already fixes $Q$, so the first and last frame suffice. The rotation gap of frame $t$ is the angle of the relative rotation in camera coordinates,
\begin{equation}
\begin{aligned}
\delta_t&=\angle\big((R^{r}_t)^{\top}Q\,R^{w}_t\big),\\
\angle(R)&=\arccos\Big(\frac{\operatorname{tr}R-1}{2}\Big),
\end{aligned}
\end{equation}
which does not depend on the choice of world frame. The code also fits scale and translation from the keyframe camera centres; these affect only camera positions and do not enter $\delta_t$. On keyframes the twin frame is the real frame, so $\delta_k$ contains only estimation error.
\begin{table}[t]
\centering\footnotesize\suptab{A1}
\begin{tabularx}{\columnwidth}{@{}lL@{}}\toprule
Model & Wan2.1-Fun-1.3B-InP (first- and last-frame inpainting) \\
Resolution & $848\times480$ \\
Sampling & 50 steps, guidance 6.0, shift 3.0, TeaCache threshold 0.10 \\
Prompt & ``A static real-world scene of a real-estate property. The camera moves smoothly; nothing in the scene moves.'' \\
Negative prompt & the model's default \\
Seed & MD5 hash of the source identifier, modulo $2^{31}$ \\
Output & H.264 (CRF 12), 30\,fps \\\bottomrule
\end{tabularx}
\caption{Settings of the twin generator.}\label{tab:gen}\suptabend
\end{table}
\begin{algorithm}[b]
\caption{Twin generation with adaptive keyframe insertion}\label{alg:twin}
\begin{algorithmic}[1]
\Require real clip $V=(I_0,\dots,I_{80})$; keyframe-conditioned generator $G$; gap threshold $0.8^\circ$; at most $N=6$ insertions
\State $\mathcal{K}\gets\{0,80\}$ \Comment{start from the first and last real frame}
\For{$i=0,1,\dots,N$}
  \State $\hat V\gets$ concatenation of $G(I_a,I_b)$ over consecutive keyframes $a<b$ in $\mathcal{K}$ \Comment{each call sees only its two end frames; segments whose end keyframes are unchanged are reused}
  \State $\hat I_k\gets I_k$ for all $k\in\mathcal{K}$ \Comment{keyframes copied back byte-identically}
  \State $R^{r}_t,R^{w}_t\gets$ camera rotations of $V$ and $\hat V$ from VGGT applied directly to the frames (no bundle adjustment)
  \State $Q\gets$ rotation that best aligns $R^{w}_k$ to $R^{r}_k$ over $k\in\mathcal{K}$ (closed form, Supp.~A.1)
  \State $\delta_t\gets\angle\big(R^{r}_t,\,Q R^{w}_t\big)$ for $t=0,\dots,80$ \Comment{rotation gap, Sec.~3.1 of the main paper}
  \State $t^\ast\gets\arg\max_{t} \delta_t$
  \If{$\delta_{t^\ast}\le0.8^\circ$ \textbf{or} $i=N$} \textbf{break} \EndIf
  \State $\mathcal{E}\gets\{4,8,\dots,76\}$ minus frames closer than 8 to some $k\in\mathcal{K}$ \Comment{generator segments need indices divisible by 4}
  \If{$\mathcal{E}=\emptyset$} \textbf{break} \EndIf
  \State $\mathcal{K}\gets\mathcal{K}\cup\{\arg\min_{e\in\mathcal{E}}|e-t^\ast|\}$ \Comment{real frame nearest to the largest gap}
\EndFor
\State \Return twin $\hat V$, keyframes $\mathcal{K}$
\end{algorithmic}
\end{algorithm}
\begin{table}[t]
\centering\footnotesize\suptab{A2}
\setlength{\tabcolsep}{3pt}
\begin{tabular*}{\columnwidth}{@{\extracolsep{\fill}}lcccc@{}}\toprule
Camera estimate on the twin & Frames & p50 & p90 & p99 \\
\midrule
\textbf{VGGT, direct (loop)} & 467 & 0.52 & 1.67 & 3.51 \\
VGGT+BA & 461 & 0.73 & 2.95 & 9.20 \\
3DGS registration + BA$^\ast$ & 467 & 0.07 & 0.18 & 0.30 \\
VGGT+BA $\oplus$ 3DGS & 467 & 0.42 & 1.71 & 5.96 \\
VGGT+BA $\oplus$ 3DGS $\oplus$ COLMAP & 467 & 0.58 & 6.11 & 25.65 \\
\bottomrule
\end{tabular*}
\caption{Cameras on generated twins: each pasted keyframe is held out in turn and the camera of the twin at that frame is compared with the real camera (rotation error in degrees; 30 training sources, 119 loop rounds). $\oplus$: fused estimates. $^\ast$All held-out keyframes had been used to fit the 3DGS model. VGGT+BA takes 10.9 (32 CPU cores) to 24.8\,min (8 cores) per side; direct VGGT takes 1.0--3.3\,min per loop round across GPU types.}\label{tab:a0}\suptabend
\end{table}
\begin{table}[t]
\centering\footnotesize\suptab{A3}
\setlength{\tabcolsep}{3pt}
\begin{tabular*}{\columnwidth}{@{\extracolsep{\fill}}lcccccc@{}}\toprule
 & \multicolumn{3}{c}{Copied keyframes} & \multicolumn{3}{c@{}}{Known offsets} \\
\cmidrule(lr){2-4}\cmidrule(l){5-7}
Camera & Med. & p99 & $\le0.8^\circ$ & Med. & p99 & Correct \\
\midrule
VGGT, crop (loop) & 0.243 & 0.816 & 39/40 & 0.060 & 0.357 & 97.4\% \\
VGGT, official & 0.241 & 0.966 & 35/40 & 0.066 & 0.341 & 97.8\% \\
VGGT+BA & -- & -- & -- & 0.015$^\dagger$ & 0.252$^\dagger$ & 99.6\%$^\dagger$ \\
\bottomrule
\end{tabular*}
\caption{Camera estimates for the insertion loop on ScanNet++, in degrees. Copied keyframes (40 twins, 240 keyframes): rotation gap, whose true value is zero; $\le0.8^\circ$: twins whose largest keyframe gap stays at or below the threshold. Known offsets (20 renderings with nominal peak rotation offsets of 0.15--6$^\circ$, 1,580 frames): absolute difference between measured and true gap in degrees, and the share of frames with the correct $0.8^\circ$ decision. $^\dagger$18 of 20 renderings (1,422 frames); --: not measured. Near the threshold (nominal peak $1^\circ$) the loop camera decides 213 of 237 frames correctly (3 false positives, 21 false negatives). Registering the cameras to a 3DGS model gives a median rotation error of $0.55^\circ$ on 80 probe frames with offsets up to $1^\circ$ and $0.64^\circ$ on all 136 scored probe frames for a model fitted on 73 views, and $0.17^\circ$ (95 frames) and $0.22^\circ$ (160 frames) for the 3DGS model the frames were rendered from.}\label{tab:a1}\suptabend
\end{table}
\begin{table}[t]
\centering\footnotesize\suptab{A4}
\setlength{\tabcolsep}{3pt}
\begin{tabular*}{\columnwidth}{@{\extracolsep{\fill}}lc@{}}\toprule
TUM RGB-D clip & Max.\ gap, rounds 0--6 ($^\circ$) \\
\midrule
freiburg2 xyz, window 0 & 0.343--0.398 \\
freiburg2 xyz, window 6 & 0.332--0.383 \\
freiburg3 structure texture far, window 2 & 0.350--0.447 \\
freiburg3 structure texture far, window 4 & 0.438--0.549 \\
\bottomrule
\end{tabular*}
\caption{Largest rotation difference over frames between VGGT cameras and globally bundle-adjusted cameras of the same real video, for four windows of two TUM RGB-D sequences; range over the seven rounds (0--6) of the pilot loop. The 28 values span $0.332$--$0.549^\circ$ with median $0.38^\circ$; the threshold $0.8^\circ$ is about twice this median.}\label{tab:a2}\suptabend
\end{table}
\item \textbf{Generator.} Each segment between two keyframes is generated from its first and last frame with the settings of Table~\ref{tab:gen}, the same for every twin. Generation is not bit-reproducible across GPU types: in a test, the same source ended with keyframes 0, 48 and 80 on an RTX 4090 and 0, 32 and 80 on an H100, and the frames and the estimated cameras differed as well; the released twins and their keyframe lists are therefore the reference.
\item \textbf{Cameras for the loop (Table~\ref{tab:a1}).} The loop compares cameras estimated by the same method on two videos, so the estimate must give a gap close to zero on keyframes and reproduce the true gap elsewhere. On generated twins, with each pasted keyframe held out in turn (Table~\ref{tab:a0}), VGGT~\cite{wang2025vggt} applied directly to the frames, with its crop preprocessing (frames resized to $518\times294$; landscape frames are not cropped), has a median rotation error of $0.52^\circ$, and VGGT with bundle adjustment $0.73^\circ$ at several times the run time; the loop therefore uses direct VGGT. On renderings it also meets both requirements (next item, Table~\ref{tab:a1}). Registering cameras through a 3DGS~\cite{kerbl20233d} fit on ScanNet++ gives a median rotation error of $0.55^\circ$ on the 80 scored probe frames with known rotation perturbations of at most $1^\circ$, and $0.64^\circ$ over all 136 scored probe frames. With VGGT's official preprocessing instead of crop, the 99th-percentile keyframe gap is $0.97^\circ$ (35 of 40 twins $\le0.8^\circ$) against $0.82^\circ$ (39 of 40).
\item \textbf{Threshold of $0.8^\circ$ (Tables~\ref{tab:a1} and~\ref{tab:a2}).} The threshold must lie above the error of the camera estimate, otherwise the loop cannot stop. On four windows of two real TUM RGB-D~\cite{sturm2012benchmark} sequences, the maximum rotation difference between VGGT cameras and globally bundle-adjusted cameras of the same video is $0.33$--$0.55^\circ$; we take about twice this, $0.8^\circ$. Check on keyframes, where the true gap is zero (240 keyframes of 40 ScanNet++ twins): median $0.24^\circ$, 99th percentile $0.82^\circ$, 3 of 240 above $0.8^\circ$. Check on 20 ScanNet++ renderings with known rotation offsets of $0.15$--$6^\circ$ (1,580 frames): the absolute difference between measured and true gap has median $0.06^\circ$ and 99th percentile $0.36^\circ$, and the $0.8^\circ$ decision is correct for 97\% of the frames, with the errors near $0.8^\circ$.
\item \textbf{Cap of 6 insertions.} Twins that reach the cap of 6 insertions are kept, with their final maximum gap recorded. On 39 ScanNet++ pilot twins that still exceeded $0.8^\circ$ after 6 insertions, continuing the loop brought none below the threshold, and from the 6th to the 8th insertion the median largest gap fell by $0.039^\circ$ (28 twins).
\item \textbf{Camera position.} The loop constrains only the camera orientation. The remaining difference in camera position (divided by depth) has a median of about 0.012 and a 95th percentile of about 0.038 (30 pilot sources); the labelling handles it through the $t_\perp/Z$ rule of Sec.~3.3 of the main paper.
\end{itemize}
\subsection{Camera selection and alignment}\label{app:cameras}
The loop and the labels use cameras for different purposes. The loop compares rotations of the same estimator on two videos at every round, so it uses direct VGGT (Supp.~\ref{app:keyframes}). The labels also need camera positions and the sparse points that fix the depth scale, so they use bundle-adjusted cameras, COLMAP+BA where it registers and VGGT+BA otherwise. The remaining camera error is part of the label noise $\hat\sigma^2$ measured on the keyframes.
Each side $s\in\{r,w\}$ (real, twin) is solved twice: with VGGT plus bundle adjustment, giving poses $P^{\mathrm{V}}_{s,t}=(R^{\mathrm{V}}_{s,t},\mathbf{t}^{\mathrm{V}}_{s,t})$, and with COLMAP~\cite{schonberger2016structure,schonberger2016pixelwise} plus bundle adjustment, giving $P^{\mathrm{C}}_{s,t}$ on its registered frames $\mathcal{T}^{\mathrm{C}}_s$. With $\angle(R,R')=\arccos\big(\tfrac12(\operatorname{tr}(R^{\top}R')-1)\big)$ and the COLMAP solution aligned to the VGGT one by a similarity transform on their common frames, COLMAP is usable on side $s$ if
\begin{equation} u_s=\Big[\,|\mathcal{T}^{\mathrm{C}}_s|\ge41\Big]\wedge\Big[\operatorname*{median}_{t\in\mathcal{T}^{\mathrm{C}}_s}\angle\big(R^{\mathrm{C}}_{s,t},R^{\mathrm{V}}_{s,t}\big)\le1^\circ\Big], \end{equation}
and each frame takes
\begin{equation} P_{s,t}=\begin{cases}P^{\mathrm{C}}_{s,t}, & u_s\wedge t\in\mathcal{T}^{\mathrm{C}}_s,\\ P^{\mathrm{V}}_{s,t}, & \text{otherwise}.\end{cases} \end{equation}
Against the dataset's own poses (Table~\ref{tab:colvggt}), on the 17 sources where both return cameras, COLMAP+BA has the lower median error ($0.047^\circ$ against $0.167^\circ$) but its worst source is off by $16.6^\circ$, it returns no usable cameras for 3 of the 20 sources, and VGGT+BA returns cameras for all 20; COLMAP+BA is therefore used only where it registers and agrees with VGGT+BA.
\begin{table}[t]
\centering\footnotesize\setlength{\tabcolsep}{3pt}\suptab{A5}
\begin{tabular*}{\columnwidth}{@{\extracolsep{\fill}}lcc@{}}\toprule
Orientation error per source & COLMAP+BA & VGGT+BA \\
\midrule
Sources with usable cameras & 17 / 20 & 20 / 20 \\
\multicolumn{3}{@{}l}{\emph{The 17 sources with both}} \\
Median over sources & $0.047^\circ$ & $0.167^\circ$ \\
Largest & $16.6^\circ$ & $2.95^\circ$ \\
Sources above $0.3^\circ$ & 3 / 17 & 3 / 17 \\
\bottomrule
\end{tabular*}
\caption{Real-side cameras of 20 RealEstate10K sources against the dataset's own poses, aligned on all 81 frames; per source, the median orientation error over its frames. COLMAP+BA returns no usable cameras for the other 3 sources: two fail to reconstruct and one registers 2 of 81 frames.}\label{tab:colvggt}\suptabend
\end{table}
The twin is mapped into the frame of the real clip by a similarity transform $(Q,\lambda,\mathbf{o})$ on the fitting frames $\mathcal{F}$: $Q$ is obtained by the closed form of Supp.~\ref{app:keyframes} over $\mathcal{F}$, and scale and translation from the camera centres $\mathbf{c}$,
\begin{equation} (\lambda,\mathbf{o})=\operatorname*{arg\,min}_{\lambda>0,\,\mathbf{o}}\sum_{t\in\mathcal{F}}\big\lVert \mathbf{c}^{r}_t-(\lambda Q\,\mathbf{c}^{w}_t+\mathbf{o})\big\rVert^2 . \end{equation}
A pair is excluded if VGGT fails on either side, if fewer than 99\% of the observations of the VGGT+BA solution of either side lie in front of its cameras, or if the real cameras differ from the dataset's own poses by more than $1^\circ$ in median. The positive-depth check is needed because bundle adjustment initialised from VGGT can converge to a solution in which the scene lies behind the cameras while the reprojection error stays low and the rotations stay proper, so neither reveals it. In 10 training pairs with this failure, exactly one of the two clips (the real one or the twin) had fewer than 1\% of its observations at positive depth, each time in its VGGT+BA solution and never in COLMAP+BA; in 5 unaffected pairs both clips had more than 99.99\%. A flip of one clip also makes the least-squares scale negative, so the requirement $\lambda>0$ had already removed these pairs; the 99\% check also catches a flip of both clips. A COLMAP+BA solution that fails the 99\% check is replaced by VGGT+BA for that clip. The cameras of both sides are fitted on all 81 frames.

\noindent\textbf{Held-out frames.} The fitting frames $\mathcal{F}$ are the 81 frames minus the real frames 5, 15, 25, 35, 45, 55, 65 and 75, the same 8 frames for every clip. These frames are never used as reference frames or in the alignment above. Each is labelled like a twin frame from the remaining real frames; its true angle is zero, so the measured angle is the error of the labelling pipeline. Pooled over the held-out frames of 20 training sources, the 95th percentile of this angle is $0.1525^\circ$; this is the noise floor $\varepsilon$ of the labels and the calibration line of the keyframe check. Noise is removed per frame with the keyframe estimate $\hat\sigma^2$ of Algorithm~\ref{alg:truth}.

\noindent\textbf{Camera gap after alignment.} We pool frame-level rotation gaps across the training pairs separately for keyframes and non-keyframes before computing each median. Over the 702 training pairs, the median rotation gap between aligned twin and real cameras is $0.25^\circ$ on non-keyframes and $0.22^\circ$ on keyframes; the gap of the camera centres, divided by the mean distance of the cameras from the first frame, is 2.9\% and 2.7\%. On keyframes the twin frame is the real frame, so the keyframe gap is the error of the camera estimate; non-keyframes add about $0.03^\circ$ and 0.2\%.

\subsection{Rotation gap and epipolar error before and after keyframe insertion}\label{app:sampson}
\begin{table}[t]
\centering\footnotesize\suptab{A6}
\setlength{\tabcolsep}{3pt}
\begin{tabular*}{\columnwidth}{@{\extracolsep{\fill}}lcc@{}}\toprule
Clip & Rotation gap ($^\circ$) & Sampson error (px) \\
\midrule
Real clip & -- & 0.212 \\
Round-0 twin (keyframes 0, 80) & 1.196 & 0.429 \\
Final twin & 0.296 & 0.424 \\
\bottomrule
\end{tabular*}
\caption{Keyframe insertion aligns the camera but not the content: medians over 40 training sources. Rotation gap: median over frames of $\delta_t$, then over twins. Sampson error: square root of the Sampson distance of a SIFT match, median over the matches of a clip, then over clips. The final twin has a larger Sampson error than its real clip in 40 of 40 clips.}\label{tab:a3}\suptabend
\end{table}
Table~\ref{tab:a3} summarises this check.
\begin{itemize}
\item \textbf{Data.} 40 training sources, each with its real clip, its round-0 twin (keyframes 0 and 80 only) and its final twin after keyframe insertion.
\item \textbf{Camera.} The rotation gap $\delta_t$ of the insertion loop (Sec.~3.1 of the main paper); per twin we take the median over frames. Median over the 40 twins: $1.20^\circ$ for the round-0 twins and $0.30^\circ$ for the final twins.
\item \textbf{Content.} An epipolar check that uses neither VGGT nor our labels. SIFT~\cite{lowe2004distinctive} keypoints (at most 4000 per frame) are matched with a ratio test of 0.8 and a mutual-consistency check, in frame pairs 8 and 16 frames apart, one pair every 4 frames (36 pairs per clip). For each pair a fundamental matrix is fitted to the pair's own matches with MAGSAC~\cite{barath2019magsac} (threshold 1\,px), and the error of a match is the square root of its Sampson distance, over all matches that pass the ratio test. Per clip we take the median over its matches, then the median over clips: 0.43\,px for the round-0 twins, 0.42\,px for the final twins and 0.21\,px for the real clips; each final twin has a larger error than its real clip in 40 of 40 clips.
\end{itemize}
\subsection{Label noise, unit and exclusion rule}\label{app:noise}
\begin{algorithm}[t]
\caption{Per-pixel ground-truth label for one twin frame $t$}\label{alg:truth}
\begin{algorithmic}[1]
\Require real frames $I_r$ with fitted cameras; twin frame $\hat I_t$ with aligned camera $(R_t,\mathbf{t}_t)$; intrinsics $K$; depth maps $Z_{m,r}$ of $M=7$ methods; keyframes $\mathcal{K}$
\State $r\gets$ the fitting frame with the smallest median $t_\perp/Z$ over its valid depth pixels
\State $s_{m,r}\gets\operatorname{median}_{k\in\mathcal{P}_r} z_k/Z_{m,r}(\mathbf{u}_k)$ \Comment{scale each map to the sparse 3D points of the camera reconstruction}
\State $D_r(\mathbf{u})\gets\operatorname{median}_m s_{m,r}Z_{m,r}(\mathbf{u})$
\For{each pixel $\mathbf{u}$ of $I_r$}
  \State $\mathbf{X}\gets D_r(\mathbf{u})K^{-1}\tilde{\mathbf{u}}$;\quad $\hat{\mathbf{u}}\gets\pi\big(K(R_{t\leftarrow r}\mathbf{X}+\mathbf{t}_{t\leftarrow r})\big)$ \Comment{z-buffer keeps the nearest}
\EndFor
\State $I^{\mathrm{ref}}_t\gets$ $I_r$ warped to the projected positions $\hat{\mathbf{u}}$ \Comment{where a rigid scene would put the content}
\State $\Delta\mathbf{u}\gets$ component-wise median of the flows $I^{\mathrm{ref}}_t\to\hat I_t$ of the largest set of DIS, SEA-RAFT and RAFT that pass the forward--backward test and agree pairwise; no flow if fewer than two
\State $\theta(\hat{\mathbf{u}})\gets\angle\big(K^{-1}\tilde{\hat{\mathbf{u}}},\,K^{-1}\widetilde{(\hat{\mathbf{u}}+\Delta\mathbf{u})}\big)$
\State labelled $\gets$ projected $\wedge$ all $M$ depth maps valid $\wedge$ $t_\perp/Z\le0.01$ $\wedge$ flow defined $\wedge$ $\theta$ finite
\State $\hat\sigma^2(t)\gets$ linear interpolation of $\hat\sigma_k^2=\operatorname{mean}_{\mathbf{u}}\min(\theta_k(\mathbf{u}),3^\circ)^2$ at the neighbouring keyframes $k\in\mathcal{K}$
\State $\theta_c\gets\sqrt{\max\!\big(\min(\theta,3^\circ)^2-\hat\sigma^2(t),\,0\big)}$ \Comment{keyframes are real, so their $\theta$ is pure noise}
\State \Return $\theta_c$ on labelled pixels (others: no label)
\end{algorithmic}
\end{algorithm}
\begin{itemize}
\begin{table*}[t]
\centering\footnotesize\suptab{A7}
\setlength{\tabcolsep}{4pt}
\begin{tabular*}{\textwidth}{@{\extracolsep{\fill}}llc@{}}\toprule
\multicolumn{2}{@{}l}{Noise estimate $\hat\sigma^2$ of a held-out keyframe} & Mean $T$ [95\% CI] / mean $|T|$ \\
\midrule
 & Constant from real held-out frames & $+0.284$ [$+0.220$, $+0.370$] / 0.284 \\
 & First and last keyframe & $-0.218$ [$-0.328$, $-0.131$] / 0.289 \\
 & Mean of all other keyframes & $-0.142$ [$-0.222$, $-0.077$] / 0.194 \\
 & \textbf{Interpolated between neighbouring keyframes} & $-0.100$ [$-0.171$, $-0.044$] / \textbf{0.153} \\
 & Regression on camera and frame features & $-0.002$ [$-0.062$, $+0.076$] / 0.271 \\
\midrule
 & Interpolated; validation, 43 pairs, single check & $+0.013$ [$+0.000$, $+0.029$] / 0.023 \\
\midrule
\multicolumn{2}{@{}l}{Interpolated vs.\ fixed noise, trained models} & Scored vs.\ interpolated-noise truth \quad / \quad vs.\ fixed-noise truth \\
\midrule
 & AUROC & $-0.006$ [$-0.050$, $+0.031$] \quad / \quad $-0.007$ [$-0.054$, $+0.031$] \\
 & Kendall $\tau$ (filtered) & $+0.061$ [$+0.020$, $+0.106$] \quad / \quad $+0.068$ [$+0.024$, $+0.113$] \\
 & Spearman & $+0.020$ [$-0.015$, $+0.052$] \quad / \quad $+0.026$ [$-0.009$, $+0.057$] \\
 & Worst-segment Spearman & $-0.039$ [$-0.183$, $+0.096$] \quad / \quad $-0.055$ [$-0.190$, $+0.075$] \\
 & ECE (six thresholds) & $-0.019$ [$-0.025$, $-0.014$] \quad / \quad $-0.020$ [$-0.026$, $-0.012$] \\
 & Angular MAE ($^\circ$) & $+0.035$ [$+0.015$, $+0.058$] \quad / \quad $+0.035$ [$+0.016$, $+0.057$] \\
 & Low-angle bias ($^\circ$) & $+0.037$ [$+0.013$, $+0.066$] \quad / \quad $+0.038$ [$+0.014$, $+0.066$] \\
\bottomrule
\end{tabular*}
\caption{Noise estimators. The labels of each keyframe are withheld in turn (the cameras are fitted on all 81 frames); since a keyframe's true angle is zero, its dimensionless target $T$ should be zero. Top: training split, on the support all five estimators cover (501 pairs, 1,474 keyframes, 398\,M pixels, $t_\perp/Z\le0.01$); mean over pairs of the per-pair mean $T$ with a 95\% pair-bootstrap interval, and the mean of its absolute value, by which the interpolated estimate is selected (bold). Middle: the interpolated estimate on validation (43 pairs, 177 keyframes, 42.8\,M pixels). Bottom: models trained with the interpolated-noise target minus models trained with a fixed noise value, all 81 frames of 25 selection pairs, 3 seeds each, scored against the truth corrected with the interpolated noise or with the fixed noise value; a positive angular MAE means a larger absolute error, a positive low-angle bias a larger overestimate.}\label{tab:a5}\suptabend
\end{table*}
\item \textbf{Expectation of the squared target.} Write the measured angular displacement of a pixel as $\boldsymbol{\delta}+\boldsymbol{\epsilon}$, with $\boldsymbol{\delta}$ the true angular displacement (its norm is the NRG angle), $\boldsymbol{\epsilon}$ the measurement error of the pipeline and $\theta=\lVert\boldsymbol{\delta}+\boldsymbol{\epsilon}\rVert$ the measured angle in degrees. If $\boldsymbol{\epsilon}$ has zero mean and is independent of $\boldsymbol{\delta}$, then
\begin{equation}
\begin{aligned}
\mathbb{E}[\theta^2\mid\boldsymbol{\delta}]&=\lVert\boldsymbol{\delta}\rVert^2+\sigma^2,\\
T=\frac{\theta^2-\sigma^2}{0.6^2}\quad&\Rightarrow\quad\mathbb{E}[T\mid\boldsymbol{\delta}]=\frac{\lVert\boldsymbol{\delta}\rVert^2}{0.6^2},
\end{aligned}
\end{equation}
where $\sigma^2=\mathbb{E}\lVert\boldsymbol{\epsilon}\rVert^2$ is the noise variance, $T$ the training target and $0.6^\circ$ the unit (next item). A squared-error loss is minimised by the conditional mean, so the trained output estimates the expected noise-free squared angle when (i) $\boldsymbol{\epsilon}$ has zero mean and is independent of $\boldsymbol{\delta}$, (ii) $\hat\sigma^2$ equals the noise variance of the pixel, (iii) the angle lies below the clip at $3^\circ$ (next items), and (iv) whether a pixel is labelled does not depend on $\boldsymbol{\delta}$. Condition (iii) fails only for the clipped pixels, fewer than 0.15\% of the labelled pixels (next items). A clip at zero or an absolute or logarithmic loss would break this.
\begin{table}[p]
\centering\footnotesize\suptab{A8}
\setlength{\tabcolsep}{4pt}
\begin{tabular*}{\columnwidth}{@{\extracolsep{\fill}}lccc@{}}\toprule
Angle bin ($^\circ$) & $f_0$ (\%) & $f$ (\%) & $f_0/f$ \\
\midrule
0--0.01 & 4.97 & 0.51 & 9.75 \\
0.01--0.02 & 11.23 & 1.50 & 7.49 \\
0.02--0.03 & 12.28 & 2.40 & 5.12 \\
0.03--0.05 & 20.36 & 6.86 & 2.97 \\
0.05--0.075 & 16.79 & 10.56 & 1.59 \\
0.075--0.1 & 10.09 & 10.67 & 0.95 \\
0.1--0.15 & 10.21 & 17.28 & 0.59 \\
0.15--0.2 & 4.61 & 11.46 & 0.40 \\
0.2--0.3 & 4.03 & 12.93 & 0.31 \\
0.3--0.45 & 2.30 & 9.22 & 0.25 \\
0.45--0.6 & 1.01 & 4.67 & 0.22 \\
\textbf{0.6--1.0} & 1.11 & 5.73 & \textbf{0.19} \\
1.0--2.0 & 0.70 & 3.90 & 0.18 \\
$>2.0$ & 0.32 & 2.30 & 0.14 \\
\midrule
$\ge0.6$ & 2.13 & 11.93 & 0.18 \\
\midrule
\multicolumn{4}{@{}l}{\emph{Noise of pixels with zero NRG angle, by what is measured}} \\
\multicolumn{2}{@{}l}{Pilot held-out frames, raw $\theta$ (above)} & \multicolumn{2}{r@{}}{2.13\% $\ge0.6^\circ$} \\
\multicolumn{2}{@{}l}{Flow only, zero motion (Tab.~\ref{tab:a8})} & \multicolumn{2}{r@{}}{0.052\% $\ge0.6^\circ$} \\
\multicolumn{2}{@{}l}{Final keyframes, raw $\theta$ (258 twins)} & \multicolumn{2}{r@{}}{RMS $0.45^\circ$} \\
\multicolumn{2}{@{}l}{Final keyframes, $\theta_c$ (702 twins)} & \multicolumn{2}{r@{}}{1.0\% $\ge0.6^\circ$} \\
\bottomrule
\end{tabular*}
\caption{Origin of the unit: share of the measured angles of held-out real pixels ($f_0$, 291.7\,M pixels) and of twin pixels ($f$, 7.50\,G pixels) per angle bin, pooled over all rounds of 174 pilot sources (pilot pipeline, raw angle $\theta$ before noise subtraction), and the bound $f_0/f$ on the local false discovery rate under the two-group model. The bound first falls below 0.2 at $0.6^\circ$ (bold). Bottom: the noise figures quoted in this supplement measure different things: all error sources on pilot held-out frames; optical flow alone on rendered zero-motion pairs; all error sources on final-pipeline keyframes, raw and after the noise subtraction of Algorithm~\ref{alg:truth}.}\label{tab:a4}\suptabend
\end{table}
\item \textbf{Unit $0.6^\circ$ and the ordinal bins (Table~\ref{tab:a4}).} Held-out real frames contain no deformation, so their measured angles give the distribution $f_0$ of the measurement error. Twin pixels follow the two-group model~\cite{efron2010large}
\begin{equation}
\begin{aligned}
f&=\pi_0 f_0+(1-\pi_0)f_1,\\
\mathrm{lfdr}(b)&=\frac{\pi_0 f_0(b)}{f(b)}\le\frac{f_0(b)}{f(b)},
\end{aligned}
\end{equation}
where $f$ is the distribution of twin angles, $f_1$ that of pixels with a non-zero NRG angle, $\pi_0$ the share of twin pixels with measurement error only, and $\mathrm{lfdr}(b)$ the local false discovery rate of angle bin $b$: the probability that a twin pixel in $b$ shows only measurement error. The bound $f_0(b)/f(b)$ is the share of held-out pixels in the bin divided by the share of twin pixels. We set the unit to the smallest bin boundary above which this bound falls below $0.2$, which gives $0.6^\circ$: a twin pixel measured at $\ge0.6^\circ$ is measurement error with probability at most about 19\%. The pooled rounds of the 174 pilot sources (Table~\ref{tab:a4}) include the early ones; restricted to their last generation round, the bound in the $0.6$--$1.0^\circ$ bin is 0.29. The unit was fixed on the pooled rounds and not changed afterwards. The rigid threshold $0.075^\circ$ is the last bin boundary below which held-out pixels are more frequent than twin pixels ($f_0/f=1.59$ in $0.05$--$0.075^\circ$, $0.95$ in $0.075$--$0.1^\circ$). Of the held-out real pixels, 65.6\% lie below $0.075^\circ$, against 21.8\% of the twin pixels; from $0.6^\circ$ on, the bound is at most 0.19 (Table~\ref{tab:a4}). The 7 ordinal bins have boundaries $0.03$, $0.075$, $0.15$, $0.3$, $0.6$ and $1.2^\circ$, which include both thresholds. Both thresholds were set on these pilot sources and then frozen; the test split was not used. On the final pipeline they keep their meaning for the noise-corrected angle $\theta_c$: on the 3{,}361 keyframes of the 702 training twins ($8.6\times10^8$ labelled pixels, all with a true angle of zero), with $\hat\sigma^2$ interpolated from the other keyframes of the twin, 1.0\% of the pixels have $\theta_c\ge0.6^\circ$ and 67.7\% have $\theta_c<0.075^\circ$ (raw angle: 2.3\% and 12.5\%). The 19\% above comes from the pilot sources and the raw angle and bounds the share of measurement error among twin pixels at or above $0.6^\circ$; the 1.0\% here comes from the final pipeline and is how often noise alone reaches $0.6^\circ$ on a keyframe after noise subtraction.
\begin{table*}[t]
\centering\footnotesize\suptab{A9}
\setlength{\tabcolsep}{3.2pt}
\begin{tabular*}{\textwidth}{@{\extracolsep{\fill}}lcccccccccccc@{}}\toprule
 & & \multicolumn{10}{c}{$t_\perp/Z$ decile of the pair (smallest to largest)} \\
\cmidrule(l){3-12}
Method & Pairs & 1 & 2 & 3 & 4 & 5 & 6 & 7 & 8 & 9 & 10 \\
\midrule
\multicolumn{12}{@{}l}{\emph{ARKitScenes (LiDAR, 34 clips)}} \\
Depth Anything V2 & 8,525 & .026 & .024 & .022 & .022 & .023 & .024 & .028 & .031 & .034 & .042 \\
Depth Anything 3 & 8,525 & .018 & .017 & .016 & .016 & .016 & .017 & .020 & .020 & .020 & .022 \\
MapAnything & 8,525 & .020 & .021 & .021 & .020 & .021 & .023 & .025 & .025 & .026 & .031 \\
Metric3D v2 & 8,525 & .023 & .020 & .019 & .018 & .017 & .018 & .020 & .023 & .024 & .031 \\
MoGe-2 & 8,525 & .017 & .015 & .015 & .014 & .014 & .016 & .018 & .020 & .020 & .023 \\
UniDepth v2 & 8,525 & .023 & .019 & .017 & .015 & .015 & .016 & .018 & .021 & .022 & .030 \\
VGGT+BA & 6,015 & .041 & .028 & .023 & .020 & .019 & .019 & .020 & .022 & .023 & .028 \\
COLMAP MVS & 2,504 & .153 & .173 & .179 & .191 & .207 & .224 & .216 & .224 & .221 & .269 \\
3DGS & 6,015 & .221 & .231 & .213 & .210 & .196 & .193 & .189 & .187 & .196 & .205 \\
\midrule
\multicolumn{12}{@{}l}{\emph{ScanNet++ (laser-scan mesh, 20 clips)}} \\
VGGT+BA & 4,769 & .043 & .054 & .068 & .089 & .105 & .114 & .137 & .160 & .171 & .230 \\
COLMAP MVS & 4,518 & .096 & .131 & .164 & .194 & .197 & .203 & .220 & .227 & .226 & .217 \\
\bottomrule
\end{tabular*}
\caption{Depth-induced ray error against sensor depth (ARKitScenes LiDAR, ScanNet++ laser-scan mesh): angle (degrees) between the ray reprojected with true depth and with each method's median-scaled depth; 95th percentile over the evaluated pixels of a pair in each of its $t_\perp/Z$ deciles, median over pairs with a finite value. Pairs: frame pairs with a completed prediction, of 8,525 (ARKitScenes) and 5,020 (ScanNet++). ScanNet++ rows: the two multi-view methods.}\label{tab:a6}\suptabend
\end{table*}
\begin{table*}[t]
\centering\footnotesize\suptab{A10}
\setlength{\tabcolsep}{3.4pt}
\begin{tabular*}{\textwidth}{@{\extracolsep{\fill}}lcccccccccccccc@{}}\toprule
 & \multicolumn{7}{c}{ARKitScenes} & \multicolumn{7}{c@{}}{ScanNet++} \\
\cmidrule(lr){2-8}\cmidrule(l){9-15}
 & \multicolumn{3}{c}{Kept (\%)} & \multicolumn{4}{c}{p95 edge ($^\circ$)} & \multicolumn{3}{c}{Kept (\%)} & \multicolumn{4}{c@{}}{p95 edge ($^\circ$)} \\
\cmidrule(lr){2-4}\cmidrule(lr){5-8}\cmidrule(lr){9-11}\cmidrule(l){12-15}
$t_\perp/Z$ & .03 & .06 & .09 & .03 & .06 & .09 & cap & .03 & .06 & .09 & .03 & .06 & .09 & cap \\
\midrule
0--.001 & 98.6 & 99.8 & 100.0 & .003 & .003 & .003 & .003 & 99.5 & 100.0 & 100.0 & .004 & .004 & .004 & .004 \\
.001--.003 & 92.8 & 98.3 & 99.4 & .010 & .014 & .016 & .017 & 84.2 & 97.5 & 99.3 & .010 & .012 & .012 & .013 \\
.003--.01 & 64.6 & 85.1 & 91.9 & .016 & .022 & .026 & .041 & 25.2 & 61.9 & 80.4 & .021 & .026 & .030 & .041 \\
.01--.03 & 18.3 & 46.3 & 63.9 & .043 & .050 & .053 & .110 & 1.3 & 10.3 & 24.1 & .044 & .061 & .066 & .137 \\
.03--.1 & 1.0 & 7.8 & 18.3 & .062 & .075 & .094 & .346 & 0.0 & 0.2 & 1.2 & .092 & .105 & .112 & .515 \\
$>.1$ & 0.0 & 0.1 & 0.7 & .162 & .126 & .124 & .970 & 0.0 & 0.0 & 0.0 & -- & .167 & .317 & 1.823 \\
\midrule
$\le.01$ & 75.8 & 90.3 & 94.8 & .014 & .019 & .023 & \textbf{.032} & 45.3 & 73.7 & 86.6 & .016 & .022 & .026 & \textbf{.034} \\
\bottomrule
\end{tabular*}
\caption{Depth rule on ARKitScenes (24 clips, 259\,M pixels) and ScanNet++ (19 clips, 1.76\,G pixels): median of the seven depth maps with a spread threshold of $0.03$, $0.06$ or $0.09^\circ$ (column heads, degrees) or with the cap alone (cap, the rule of the released labels; all pixels kept); 0.0: below 0.05\%. Kept: share of pixels in the $t_\perp/Z$ bin that pass the spread test; p95 edge: upper edge of the histogram bin holding the 95th percentile of the depth-induced ray error of the kept pixels; last row: all pixels with $t_\perp/Z\le0.01$ (bold: released labels). Above $t_\perp/Z=0.01$ the error rises steeply in every column; hence the cap at $0.01$. --: no pixel kept.}\label{tab:a7}\suptabend
\end{table*}
\item \textbf{Clip at $3^\circ$.} Target and noise estimate use the same clipped angle $\min(\theta,3^\circ)$. Keyframes are real frames, so every angle measured there is measurement error: on 1{,}230 keyframes of 258 twins, the 0.15\% of labelled pixels above $3^\circ$ raise the mean $\theta^2$ from $0.202$ to $0.248\,\mathrm{deg}^2$. The clip keeps these rare large errors from dominating $\hat\sigma^2$ and the variance of $T$. For a pixel with $\theta\ge3^\circ$, the noise-corrected angle $\theta_c$ of Algorithm~\ref{alg:truth} takes the per-frame value
\begin{equation}
\begin{aligned}
\theta_c&=\sqrt{9-\hat\sigma^2}>1.2^\circ\\
&\text{for any }\hat\sigma^2<7.56\,\mathrm{deg}^2,
\end{aligned}
\end{equation}
where $\hat\sigma^2$ is the estimated noise variance of its frame and $1.2^\circ$ the lower boundary of the top bin. The ordinal bin of these pixels is unchanged; their continuous label and regression target take this common value. This affects fewer than 0.15\% of the labelled pixels: angles above $1.5^\circ$ make up 0.06--0.13\% (Supp.~E).
\item \textbf{Noise interpolated between keyframes (Table~\ref{tab:a5}).} Estimating the noise from the first and last keyframe only subtracts too much (243 pairs, mean target $-0.32$): the noise varies with the frame position. Of five estimators compared on the common training support of 501 pairs, linear interpolation between the neighbouring keyframes has the smallest mean absolute bias ($0.153$; mean bias $-0.100$, 95\% interval $-0.171$ to $-0.044$), and $+0.013$ on a single check on 43 validation pairs. Against a fixed noise value (25 selection pairs, 3 seeds each), its AUROC differs by $-0.006$ (95\% interval $-0.050$ to $+0.031$), while Kendall's $\tau$ improves by $+0.061$ ($+0.020$ to $+0.106$) and the calibration error drops by $0.019$ ($0.014$ to $0.025$).
\item \textbf{Check of the noise subtraction at zero NRG angle.} Keyframes have a true angle of zero, so the evaluation rule applied to them shows how often noise alone reaches the NRG threshold. On the 3{,}361 keyframes of the 702 training twins ($8.6\times10^8$ labelled pixels), $\theta_c\ge0.6^\circ$ holds for 0.47\% of the pixels with the keyframe's own $\hat\sigma^2$ and for 1.0\% with $\hat\sigma^2$ interpolated from the other keyframes, against 1.49\% of the non-key pixels (Table~\ref{tab:e1}). The rate is concentrated: the 10\% of twins with the highest keyframe rate contribute 58\% of these pixels, and 65\% of the twins stay below 0.5\%. Across twins, the keyframe rate correlates with the NRG share of the non-key frames (Spearman 0.57; 5.5\% NRG pixels in the 10\% noisiest twins, 1.2\% in the others). This is consistent with inconsistent content making the camera estimate harder; label noise in these twins is also possible. The subtraction assumes that generated frames have the noise measured on the keyframes; camera error specific to generated frames cannot be measured. All methods are scored against the same labels, and the per-frame $\hat\sigma^2$ of every twin is released so that noisy twins can be excluded.
\item \textbf{Exclusion rule.} Over the labelled pixels of all keyframes of a twin (angles above $3^\circ$ counted as $3^\circ$) we compute (i) the RMS of the keyframe angle and (ii) its RMS after subtracting, at each keyframe, the noise interpolated from the neighbouring keyframes. A twin is excluded if (i) $>0.6^\circ$: the label error exceeds one unit ($0.6^\circ$); if (ii) $>0.3^\circ$: one keyframe is clearly worse than its neighbours, a local failure; or if fewer than two keyframes can be checked. Rule (ii) compares keyframes with each other; twins that are noisy throughout are caught by (i).
\end{itemize}
\subsection{How the depth and flow rules were validated}\label{app:truthval}
\begin{itemize}
\item \textbf{Intrinsics.} Real and twin frames use the intrinsics of the RealEstate10K~\cite{zhou2018stereo} annotation, scaled to $848\times480$. The source frames are $640\times360$ for 845 of the 871 clips, including all validation and test clips, and $854\times480$ for 26 training clips; every frame is resized to $848\times480$ without cropping (scale $1.325\times1.333$ or $0.993\times1$). The intrinsics are constant within each clip ($c_x=424$, $c_y=240$). The generator outputs frames of the same size and field of view, and the keyframes are copied back byte for byte; LTX-2.3~\cite{hacohen2026ltx2} is not used because it rescales and crops the frames. For small angles
\begin{equation}
\theta\approx\frac{\lVert\Delta\mathbf{u}\rVert}{f},
\end{equation}
with $\theta$ in radians, $\Delta\mathbf{u}$ the image displacement and $f$ the focal length, both in pixels, so a relative error of the focal length changes $\theta$ by about the same relative amount. A resize by a $3\times3$ scaling matrix $S$, applied to both the homogeneous pixel coordinates $\tilde{\mathbf{x}}$ and the intrinsic matrix $K$, leaves every viewing ray unchanged,
\begin{equation}
(SK)^{-1}S\tilde{\mathbf{x}}=K^{-1}S^{-1}S\tilde{\mathbf{x}}=K^{-1}\tilde{\mathbf{x}},
\end{equation}
so the angle between two rays is unchanged as well. We therefore measure inconsistency as an angle rather than a pixel displacement: the same angle covers a different number of pixels at different resolutions and for different focal lengths (Table~\ref{tab:angpx}), whereas the angle does not change when frames are resized; and, unlike a displacement in metres, it needs no metric depth scale, which generated video does not provide. Content displaced along the viewing ray of the twin camera projects to the same pixel and has $\theta=0$.
\begin{table}[t]
\centering\footnotesize\setlength{\tabcolsep}{3pt}\suptab{A11}
\begin{tabular*}{\columnwidth}{@{\extracolsep{\fill}}lcc@{}}\toprule
Recorded as & Labels, $848\times480$ & Evaluation, $518\times294$ \\
\midrule
\multicolumn{3}{@{}l}{\emph{NRG threshold}} \\
Angle & $0.6^\circ$ & $0.6^\circ$ \\
Pixel displacement (px) & 4.34 [4.13, 4.81] & 2.65 [2.52, 2.94] \\
\midrule
\multicolumn{3}{@{}l}{\emph{Rigid threshold}} \\
Angle & $0.075^\circ$ & $0.075^\circ$ \\
Pixel displacement (px) & 0.54 [0.52, 0.60] & 0.33 [0.32, 0.37] \\
\bottomrule
\end{tabular*}
\caption{One deviation recorded as an angle and as a pixel displacement. The angle is the same at both resolutions; the pixel displacement shrinks with the frame and also differs between clips with different focal lengths. Pixel displacement at the image centre, $f\tan\theta$: median over the 784 training and validation twins, with the 5th--95th percentile over clips in brackets; focal length $f$ at $848\times480$: median 414.8\,px, 5th--95th percentile 394.4--459.3\,px, range 297.5--627.7\,px, scaled by $518/848$ for the evaluation grid. The thresholds apply to the noise-corrected angle $\theta_c$.}\label{tab:angpx}\suptabend
\end{table}
\item \textbf{Single-image vs.\ multi-view depth (Table~\ref{tab:a6}).} We compared six single-image methods (Depth Anything V2~\cite{yang2024depthanythingv2}, Depth Anything 3~\cite{lin2025depthanything3}, MapAnything~\cite{keetha2026mapanything}, Metric3D v2~\cite{hu2024metric3dv2}, MoGe-2~\cite{wang2025moge2} and UniDepth v2~\cite{piccinelli2025unidepthv2}), VGGT with bundle adjustment, COLMAP multi-view stereo with fixed cameras and 3DGS depth against LiDAR (ARKitScenes~\cite{baruch2021arkitscenes}, 34 clips) and laser-scan depth (ScanNet++~\cite{yeshwanth2023scannet++}, 20 clips), measuring the 95th percentile of the depth-induced ray error. At the smallest $t_\perp/Z$ decile on ARKitScenes it is $0.017$--$0.026^\circ$ for the single-image methods, $0.041^\circ$ for VGGT+BA, $0.153^\circ$ for COLMAP multi-view stereo (dense depth for 2{,}504 of 8{,}525 frame pairs) and $0.221^\circ$ for 3DGS; on ScanNet++ VGGT+BA gives $0.043^\circ$ and COLMAP $0.096^\circ$. Under the criterion fixed before this comparison (the 95th percentile over all pixels pooled in each $t_\perp/Z$ bin at most $0.03^\circ$ on both ARKitScenes and ScanNet++), no single method passed once $t_\perp/Z$ exceeded 0.003, so the depth maps are fused: the fused depth is the median of the six single-image maps and the multi-view VGGT+BA depth, each scaled to the multi-view sparse points; frames with too few 3D points fall back to the VGGT points and are flagged.
\begin{figure}[t]
\centering\supfig{A1}
\includegraphics[width=\columnwidth]{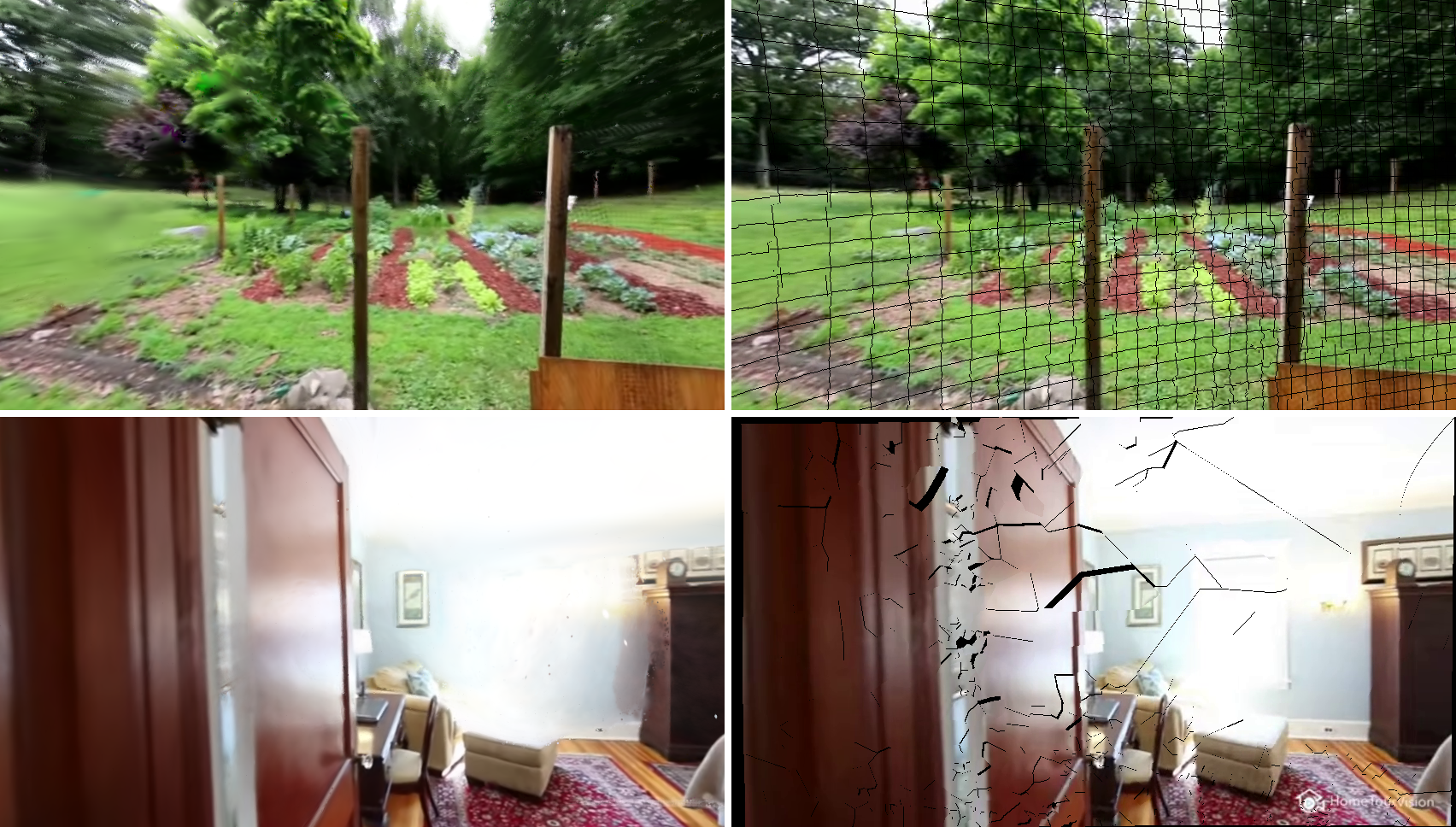}
\caption{The two references at frame 50 of the pilot twins Garden (top) and Bedroom (bottom). Left: G, the 3DGS fit of the real clip rendered at the twin camera. Right: R, the real fitting frame reprojected to the twin camera with the fused depth; black lines are pixels the reprojection does not reach.}\label{fig:tworef}\supfigend
\end{figure}
\begin{table}[t]
\centering\footnotesize\suptab{A12}
\begin{tabular*}{\columnwidth}{@{\extracolsep{\fill}}lcccc@{}}\toprule
\multicolumn{5}{@{}l}{Share of all twin pixels left after each step, 7 sources pooled} \\
\midrule
\multicolumn{4}{@{}l}{reached by the depth reprojection R} & 75.1\% \\
\multicolumn{4}{@{}l}{and $t_\perp/Z\le0.01$} & 44.1\% \\
\multicolumn{4}{@{}l}{and the depth-spread test} & 38.3\% \\
\multicolumn{4}{@{}l}{and G and R agree in flow} & 3.3\% \\
\multicolumn{4}{@{}l}{and noise floor $\le0.3^\circ$ (labelled)} & 0.19\% \\
\midrule
Scene & Spread & Flow & Labelled & Floor \\
\midrule
Garden & 62.3\% & 0.16\% & 0.12\% & $0.21^\circ$ \\
Bedroom & 31.9\% & 5.53\% & 0 & $0.91^\circ$ \\
Workbench & 46.9\% & 5.87\% & 0 & $0.38^\circ$ \\
Stove & 28.2\% & 7.42\% & 0 & $1.87^\circ$ \\
Kitchen island & 7.5\% & 0.00\% & 0 & $0.34^\circ$ \\
Dining room & 34.8\% & 3.45\% & 0.97\% & $0.25$--$0.27^\circ$ \\
Aerial view & 56.7\% & 0.40\% & 0.27\% & $0.11^\circ$ \\
\midrule
\multicolumn{5}{@{}l}{Depth / SIFT-triangulated depth, median per source} \\
\midrule
\multicolumn{4}{@{}l}{rendered 3DGS depth (6 sources)} & 0.59--0.98 \\
\multicolumn{4}{@{}l}{bundle-adjusted depth (6 sources)} & 0.98--1.05 \\
\bottomrule
\end{tabular*}
\caption{Pilot of the two-reference design on 7 RealEstate10K training sources ($2.3\times10^8$ pixel-frames).}\label{tab:tworef}\suptabend
\end{table}
\item \textbf{Depth reprojection and a rendered 3DGS reference (Table~\ref{tab:tworef}, Fig.~\ref{fig:tworef}).} An earlier design also rendered a 3DGS fit of the real clip at the twin camera and kept a pixel only where both references agreed. On 7 training sources ($2.3\times10^8$ pixel-frames) this labelled 0.19\% of the pixels. Most pixels were lost where the two references disagree in flow (38.3\% $\to$ 3.3\%). Four sources have a noise floor above $0.3^\circ$ and were labelled nowhere. The columns of Table~\ref{tab:tworef}: Spread, pixels passing the depth-spread test; Flow, pixels where G and R also agree in flow; Floor, the 95th-percentile noise floor of the source; frames without a finite floor are not labelled. The rendered 3DGS depth is shorter than SIFT-triangulated depth by a factor of 0.59--0.98 on 6 sources, against 0.98--1.05 for the bundle-adjusted depth. With depth reprojection alone about half of the twin pixels are labelled (47.9\% in the training split).
\item \textbf{Depth rule (Table~\ref{tab:a7}).} A pixel is labelled where all $M=7$ depth maps are valid, with their median as depth, and $t_\perp/Z\le0.01$, where $t_\perp$ is the camera baseline perpendicular to the viewing ray and $Z$ the depth; to first order a relative depth error $\delta Z/Z$ shifts the reprojected ray by $(t_\perp/Z)\,|\delta Z/Z|$ radians. $t_\perp/Z$ is computed on the reference grid and used at the same pixel position of the twin frame; looking it up at each twin pixel's reference pixel through the depth buffer instead would add 1.38\% and remove 0.10\% of the labelled pixels (every fourth non-key frame of the 702 training twins). Training, evaluation and the released labels all use the same rule. We measure the resulting ray error against sensor depth on ARKitScenes~\cite{baruch2021arkitscenes} (LiDAR, 24 clips) and ScanNet++ (laser-scan mesh, 19 clips), with scale set from sparse points only. With the cap, its 95th percentile is $0.032^\circ$ and $0.034^\circ$ (Table~\ref{tab:a7}, last row), about a fifth of the noise floor of the labels ($0.1525^\circ$); above $0.01$ several methods fail together and the error rises steeply, hence the cap. A spread test on
\begin{equation}
\rho=\frac{t_\perp}{Z}\cdot\frac{\max_m Z_m-\min_m Z_m}{\operatorname{median}_m Z_m}\quad\text{(radians)},
\end{equation}
with threshold $0.3\,\varepsilon$ ($0.046^\circ$), would lower this percentile to about $0.02^\circ$ (Table~\ref{tab:a7}, columns $0.03$--$0.09^\circ$) but remove 12.3\% of the labelled training pixels (every fourth non-key frame of the 702 training twins, $2.17\times10^9$ pixels; more than 38\% in the 10\% most affected twins). Since this gain is small against the noise floor, the released labels use the cap without the spread test. For each twin frame, we choose the fitting frame that minimizes the median $t_\perp/Z$ over its valid depth pixels.
\begin{table}[t]
\centering\footnotesize\suptab{A13}
\begin{tabular*}{\columnwidth}{@{\extracolsep{\fill}}lccc@{}}\toprule
Intercept & p95 error [95\% CI] & $\ge0.6^\circ$ & Labelled \\
\midrule
0.5\,px & $0.104^\circ$ [0.095, 0.115] & 0.040\% & 73.0\% \\
1\,px (used) & $0.116^\circ$ [0.107, 0.127] & 0.052\% & 81.7\% \\
\midrule
\multicolumn{4}{@{}l}{\emph{Forward--backward failures, share of supported pixels}} \\
\multicolumn{2}{@{}l}{DIS} & \multicolumn{2}{r@{}}{52.5\%} \\
\multicolumn{2}{@{}l}{SEA-RAFT} & \multicolumn{2}{r@{}}{15.8\%} \\
\multicolumn{2}{@{}l}{RAFT} & \multicolumn{2}{r@{}}{19.8\%} \\
\midrule
\multicolumn{4}{@{}l}{\emph{Outcome of the three-method rule (0.5\,px), share of supported pixels}} \\
\multicolumn{2}{@{}l}{Labelled} & \multicolumn{2}{r@{}}{73.0\%} \\
\multicolumn{2}{@{}l}{Fewer than two methods pass} & \multicolumn{2}{r@{}}{22.3\%} \\
\multicolumn{2}{@{}l}{At least two pass, no agreeing pair} & \multicolumn{2}{r@{}}{4.7\%} \\
\multicolumn{2}{@{}l}{Exactly one method fails} & \multicolumn{2}{r@{}}{at least 83\% of these labelled} \\
\bottomrule
\end{tabular*}
\caption{Flow rule in the zero-motion diagnostic (253 sources, 2,024 real held-out frames, 816\,M supported pixels): every measured angle is error. Top: both intercepts on the same recomputed flows; p95 error in degrees; $\ge0.6^\circ$: share of labelled pixels measured at NRG level; Labelled: share of reference-valid pixels. Bottom: a separate partition run at 0.5\,px: per-method failures and where the pixels go under the consensus rule; the last row is a subgroup, with the bound derived in Supp.~\ref{app:truthval}.}\label{tab:a8}\suptabend
\end{table}
\item \textbf{Flow rule (Table~\ref{tab:a8}).} The forward--backward test and the agreement test (Sec.~3.4 of the main paper) both use the tolerance
\begin{equation}
\tau=1\,\mathrm{px}+0.05\,\lVert\mathbf{f}\rVert,
\end{equation}
where $\mathbf{f}$ is the flow vector of the pixel; for the agreement test $\lVert\mathbf{f}\rVert$ is the shorter of the two flows. The constant treats a forward--backward error or a disagreement below one pixel as no error; the second term lets the tolerance grow with the length of the flow. The zero-motion diagnostic measures the optical-flow error alone: its references are 3DGS renders at real held-out views, compared with the real frames (253 sources, 2{,}024 frame pairs). The labels use the depth-reprojected references, whose depth error is bounded by the depth rule. The original 0.5\,px intercept gives a 95th-percentile error of $0.104^\circ$ ($0.095$--$0.115^\circ$), with 0.040\% of accepted pixels at $\ge0.6^\circ$ and 73\% of reference-valid pixels accepted. The replay applies both tolerance settings to the same recomputed flows on those frame pairs. At the production intercept of 1\,px, the corresponding values are $0.116^\circ$ ($0.107$--$0.127^\circ$), 0.052\% and 82\%; the error stays far below $0.6^\circ$. DIS~\cite{kroeger2016fast} uses the medium preset with a frozen reliability mask, RAFT~\cite{teed2020raft} the large model. We use all three flows if they pass the forward--backward test and agree pairwise. Otherwise, we take the first pair that passes and agrees, in the order DIS and SEA-RAFT~\cite{wang2024sea}, DIS and RAFT, SEA-RAFT and RAFT. The residual is the component-wise median of the selected flows. On the 2{,}024 frame pairs of the zero-motion diagnostic, the forward--backward test fails on 52.5\% of the supported pixels for DIS, 15.8\% for SEA-RAFT and 19.8\% for RAFT; of the pixels on which exactly one method fails, at least 83\% are still labelled because the other two agree.
\begin{table}[t]
\centering\footnotesize\suptab{A14}
\begin{tabularx}{\columnwidth}{@{}Lr@{}}\toprule
Stage & Cost \\
\midrule
Wan generation, all iterations (871 pairs) & 197.3 GPU-h \\
Camera inference in the insertion loop (775 pairs timed) & 30.4 GPU-h \\
Point tracking for label-camera BA (628) & 30.3 GPU-h \\
Six single-image depth methods (5,226 runs) & 43.6 GPU-h \\
\quad Metric3D v2 / MapAnything / DA2 & 20.7 / 7.8 / 5.3 \\
\quad MoGe-2 / UniDepth v2 / DA3 & 4.2 / 3.8 / 1.9 \\
\midrule
Total, timed GPU stages & 301.6 GPU-h \\
Label-camera BA optimisation (1,742 sides) & 3,053 CPU core-h \\
COLMAP with BA (459 pairs timed) & 87.5 h summed wall clock \\
Label assembly (871 pairs) & 94.6 h summed wall clock \\
\bottomrule
\end{tabularx}
\caption{Measured costs from the final receipts of 871 kept pairs. Counts in brackets give the coverage of each timer. The insertion-loop camera timer covers 775 pairs with additional adaptive calls. The other 96 pairs had no additional calls. The COLMAP sum covers 459 pairs with linked READY timers for both the real and twin sides. The other 412 pairs lack linked READY timers. Generation timing covers recorded successful iterations, including 2,366 adaptive calls. BA uses actual CPU time. Untimed stages include VGGT inference for label cameras, camera selection and optical flow. The measured totals are lower bounds.}\label{tab:a9}\suptabend
\end{table}
\item \textbf{Splits and cost (Table~\ref{tab:a9}).} Split seed 20261109; every label map's checksum is checked. Measured cost from the production logs of all 871 pairs: at least 301.6 GPU-hours (generation 197.3, camera inference for adaptive keyframe insertion 30.4, point tracking for the bundle adjustment of the label cameras 30.3, six depth methods 43.6; VGGT inference for the label cameras and the optical flow are not timed, so this is a lower bound); median 2 adaptive generator calls per twin (IQR 1--4; mean 2.7; 2366 in total); VGGT with bundle adjustment 3053 CPU core-hours.
\begin{table*}[t]
\centering\footnotesize\suptab{A15}
\setlength{\tabcolsep}{3pt}
\begin{tabular*}{\textwidth}{@{\extracolsep{\fill}}llrrrrrrrrr@{}}\toprule
Check & Nominal yaw & Labels & Slope & Bias ($^\circ$) & MAE ($^\circ$) & p95 ($^\circ$) & In bin & Inside lost & Outside lost & Feather folds \\
\midrule
1 & All angles  & 7{,}357{,}712 & 1.001 & $-0.0062$ & 0.020 & 0.119 &  & 16.75\% & 31.96\% &  \\
  & $0.15^\circ$ & 1{,}445{,}205 & 0.996 & $-0.005$ & 0.013 & 0.064 & 23.62\% & 8.08\% & 15.82\% &  \\
  & $0.3^\circ$  & 1{,}296{,}351 & 0.980 & $-0.012$ & 0.017 & 0.110 & 42.52\% & 14.59\% & 24.23\% &  \\
  & $0.6^\circ$  & 1{,}280{,}548 & 0.994 & $-0.008$ & 0.017 & 0.091 & 46.49\% & 20.27\% & 34.39\% &  \\
  & $1.2^\circ$  & 1{,}095{,}210 & 1.011 & $-0.003$ & 0.021 & 0.096 & 48.11\% & 19.70\% & 46.18\% &  \\
  & $3^\circ$    & 1{,}194{,}066 & 1.005 & $-0.004$ & 0.029 & 0.152 & $>99.99\%$ & 19.38\% & 36.34\% &  \\
  & $6^\circ$    & 1{,}046{,}332 & 0.980 & $-0.003$ & 0.037 & 0.193 & $>99.99\%$ & 19.29\% & 37.64\% &  \\
\addlinespace
2 & All angles  & 4{,}497{,}848 & 0.947 & $-0.0750$ & 0.024 & 0.155 &  & 22.62\% & 22.80\% &  \\
  & $0.15^\circ$ & 891{,}522 & 0.983 & $-0.017$ & 0.018 & 0.080 & 20.81\% & 7.74\% & 7.01\% &  \\
  & $0.3^\circ$  & 851{,}660 & 0.973 & $-0.032$ & 0.023 & 0.167 & 31.07\% & 13.48\% & 12.06\% &  \\
  & $0.6^\circ$  & 662{,}501 & 0.979 & $-0.021$ & 0.020 & 0.141 & 35.52\% & 28.05\% & 29.11\% &  \\
  & $1.2^\circ$  & 692{,}255 & 0.962 & $-0.019$ & 0.024 & 0.121 & 43.82\% & 28.90\% & 29.66\% &  \\
  & $3^\circ$    & 697{,}372 & 1.025 & $-0.022$ & 0.028 & 0.145 & 99.88\% & 26.90\% & 27.29\% & 17.64\% \\
  & $6^\circ$    & 702{,}538 & 1.063 & $-0.360$ & 0.041 & 5.599 & 99.80\% & 30.69\% & 31.74\% & 17.80\% \\
\midrule
\multicolumn{2}{@{}l}{Check 1, original angle $\le 0.03^\circ$} & 133{,}375 & 1.005 & $-0.0063$ & 0.026 & & & & &  \\
\bottomrule
\end{tabular*}
\caption{Known-motion injection (Supp.~\ref{app:knownmotion}). Labels, slope, mean signed bias, median absolute error (MAE), and the 95th percentile of absolute error (p95) use the original-flow-composed angle as the target. The original-angle subset in the last row uses the exact injected angle as the target. In-bin share compares the exact injected and measured bins on all labelled core pixels at each nominal yaw. Inside and outside loss are fractions of originally labelled pixels on their separately defined baseline support. Feather folds are valid feather pixels with a non-positive warp determinant, shown for the two largest yaws in Check~2. Check~1 used a hard edge.}\label{tab:a10}\suptabend
\end{table*}
\item \textbf{Data not included in this version.} LTX-2.3 twins: the generator rescales and crops its input frames, and after the exact inverse 40 RealEstate10K clips keep a residual zoom of 0.97\%, above our 0.5\% limit. ScanNet++ twins: in a real-frame flow check against mesh-projected references, the 95th-percentile angular error at a one-frame gap is $0.96^\circ$ (RealEstate10K, against 3DGS render references: $0.10^\circ$), above the $0.6^\circ$ that defines NRG pixels. DL3DV twins: not finished through the full labelling chain for this version.
\end{itemize}

\subsection{Known-motion injection}\label{app:knownmotion}
We apply a known image warp to a region of the twin frames with the cameras, depth and reference frames fixed. Rerunning the flow and label computation on these inputs checks the flow-to-angle scale with camera estimation held outside the experiment. This is the experiment behind the known-motion sentence in Sec.~3.6 of the main paper. Its slope 1.001 and median absolute error $0.020^\circ$ use all retained Check~1 labels against the original-flow-composed target; compared directly with the injection, Check~1's 133{,}375 labels with original angle $\le0.03^\circ$ give slope 1.005 and median absolute error $0.026^\circ$ (Table~\ref{tab:a10}).

\emph{Protocol.} Both checks use the same 20 RealEstate10K training twins (seed 20261108) and six non-keyframes per twin, with different frame selections. Each frame receives one nominal yaw $a\in\{0.15,0.3,0.6,1.2,3,6\}^\circ$. Check~1 applies the homography
\begin{equation}
H_a=K\,R_y(a)\,K^{-1}
\end{equation}
to the central region $[0.2W,0.8W]\times[0.2H,0.8H]$, where $K$ is the intrinsic matrix and $R_y(a)$ is rotation by $a$ about the vertical camera axis. Check~2 independently implements the warp with a central $[0.25W,0.75W]\times[0.25H,0.75H]$ region and a 16-pixel feathered border. Both recompute labels on valid projections with $t_\perp/Z\le0.01$. For the composed comparison we follow each originally measured flow by the known warp and convert that displacement to the expected angle.

\emph{Results.} The pooled composed comparisons give slopes 1.001 and 0.947 and median absolute errors $0.020^\circ$ and $0.024^\circ$ in Checks~1 and~2. Comparing all labelled core pixels directly with the injection gives median absolute errors $0.117^\circ$ and $0.131^\circ$, respectively. When the exact injected angle is at least $1.2^\circ$, 99.91\% and 99.81\% of labels fall in the injected bin. Bin agreement is lower at smaller angles. Both checks lose labels even outside the warped region. At $6^\circ$, absolute error in the composed comparison reaches $0.19^\circ$ and $5.60^\circ$ at the 95th percentile. Folding in the feathered warp can introduce ambiguous correspondences near the core.

\subsection{Source clips and their screening}\label{app:sources}
\begin{table*}[t]
\centering\footnotesize\suptab{A16}
\setlength{\tabcolsep}{4pt}
\begin{tabular*}{\textwidth}{@{\extracolsep{\fill}}lrrrrrrrrr@{}}\toprule
 & \multicolumn{3}{c}{pixelSplat} & \multicolumn{3}{c}{YouTube downloads} & \multicolumn{3}{c}{Combined} \\
\cmidrule(lr){2-4}\cmidrule(lr){5-7}\cmidrule(l){8-10}
 & Train & Val & Test & Train & Val & Test & Train & Val & Test \\
\midrule
Final twin on the formal split roster & 540 & 101 & 101 & 258 & 0 & 0 & 798 & 101 & 101 \\
Absent from the completed labelling roster & 18 & 8 & 5 & 0 & 0 & 0 & 18 & 8 & 5 \\
Completed labelling chain & 522 & 93 & 96 & 258 & 0 & 0 & 780 & 93 & 96 \\
\midrule
Raw keyframe RMS $>0.6^\circ$ & 52 & 8 & 7 & 9 & 0 & 0 & 61 & 8 & 7 \\
Noise-subtracted RMS $>0.3^\circ$ & 3 & 1 & 0 & 4 & 0 & 0 & 7 & 1 & 0 \\
Raw keyframe RMS not computable & 0 & 1 & 0 & 1 & 0 & 0 & 1 & 1 & 0 \\
Noise-subtracted RMS not computable & 7 & 1 & 2 & 0 & 0 & 0 & 7 & 1 & 2 \\
Positive-depth observations $<99\%$ & 1 & 0 & 0 & 0 & 0 & 0 & 1 & 0 & 0 \\
Positive-depth check not possible & 0 & 0 & 0 & 1 & 0 & 0 & 1 & 0 & 0 \\
\midrule
Kept in NRG-Twins & 459 & 82 & 87 & 243 & 0 & 0 & 702 & 82 & 87 \\
Test twins with every keyframe scorable & & & 83 & & & 0 & & & 83 \\
\bottomrule
\end{tabular*}
\caption{Formal twin cohort by source and split. The first row counts indexed finals on the formal split roster. The middle rows report the mutually exclusive exclusions applied to completed pairs. The last row gives the test evaluation cohort.}\label{tab:a12}\suptabend
\end{table*}
\paragraph{Intake.} Every source is an 81-frame window of a RealEstate10K training sequence. From the pixelSplat release, a sequence qualifies if it has 81 consecutive frames with timestamp steps of 33 or 34\,ms (30\,fps) whose frames all decode at the same size, with no two consecutive frames identical and no frame-to-frame difference above 5 times the clip median (a cut). Of 14{,}000 sequences drawn in a fixed order (sorted by the MD5 of the sequence id), 8{,}904 have such a window, 7{,}109 pass the frame checks and 6{,}495 pass the screen below. The 243 kept training twins from the earlier YouTube downloads passed the same screen. We assign clips from each original YouTube video to a single split, targeting an $8{:}1{:}1$ ratio for training, validation and test.
\paragraph{Screen.} Frames are converted to grey at 640\,px width; up to 400 Shi--Tomasi corners are tracked with pyramidal Lucas--Kanade (forward--backward error below 1\,px), and a similarity transform is fitted to each consecutive frame pair with RANSAC (2\,px). Accumulating the in-plane rotations gives the rotation path of the clip. The shake $s_{2D}$ is the RMS difference between this path and its 9-frame moving average, and the speed $v$ is the median per-frame translation in pixels. A source is kept if $s_{2D}<0.162^\circ$ and $v\ge0.66$\,px per frame; TUM RGB-D clips are never used.
\begin{table}[t]
\centering\footnotesize\suptab{A17}
\setlength{\tabcolsep}{4pt}
\begin{tabular*}{\columnwidth}{@{\extracolsep{\fill}}lcc@{}}\toprule
Vote batch & Clips & Agree with screen \\
\midrule
1--5, set the thresholds (in-sample) & 25 & 25/25 \\
6, extremes, TUM and random & 30 & 26/30 \\
7A, random, preregistered & 60 & 57/59 \\
7B, near the speed threshold & 7 & 3/6 \\
7B, near the shake threshold & 7 & 5/7 \\
7B, near the 3D-shake check & 6 & 6/6 \\
\midrule
\multicolumn{3}{@{}l}{Votes 6--7: screen keeps 98 (93 acceptable, 5 shaky or slow),} \\
\multicolumn{3}{@{}l}{excludes 10 (6 acceptable, 4 shaky or slow)} \\
\midrule
\multicolumn{3}{@{}l}{RealEstate10K intake} \\
\midrule
Sequences drawn (MD5 order) & \multicolumn{2}{c}{14{,}000} \\
81-frame window at 30\,fps & \multicolumn{2}{c}{8{,}904} \\
Frame checks passed & \multicolumn{2}{c}{7{,}109} \\
Shake and speed screen passed & \multicolumn{2}{c}{6{,}495} \\
Registered for generation & \multicolumn{2}{c}{3{,}664} \\
With a generation claim & \multicolumn{2}{c}{2{,}648} \\
Initial generation ready & \multicolumn{2}{c}{2{,}637} \\
Final twin indexed (10 November) & \multicolumn{2}{c}{2{,}554} \\
\bottomrule
\end{tabular*}
\caption{Source screen. Top: agreement of human votes with the screen (unsure votes left out); later votes were cast after the thresholds were fixed (Fig.~\ref{fig:votes}). Bottom: intake funnel.}\label{tab:a11}\suptabend
\end{table}
\paragraph{Generation and admission (Tables~\ref{tab:a11} and~\ref{tab:a12}).}
The pixelSplat sampler excluded earlier registered sequences, then screened in MD5 order. The generation roster opened with batch~001 and extended through batch~016. Workers claimed eligible rows in roster order as slots became available. The recorded claims span 5--7~November. The generation log records a disk-space hold on 5~November. The campaign's final stop rule, date or compute cap is not recorded. For each twin, the loop ends when the rotation gap is at most $0.8^\circ$ or reaches six insertions. Both outcomes are included in the final-twin count.
In the 10~November acceptance snapshot, 2{,}554 pixelSplat finals and 632 YouTube finals are indexed, with no shared sequence IDs between the two source groups. The earlier YouTube production readout counted 1{,}186 finals on 5~November. Its total dispatch and final campaign counts are not recorded in this census. The formal split roster includes 1{,}000 of the indexed finals and assigns every represented original YouTube video to one split. Of these, 969 appear in the completed labelling roster. Among the 18 pixelSplat training pairs absent from the completed roster, 12 were excluded for invalid alignment scale, three lacked pre-BA camera receipts, one reached labelling with an empty fit-frame depth input, and two had no source receipt in the census. The census does not record stage-level causes for the eight validation and five test pairs absent from the completed roster. Campaign-wide failure totals for generation, cameras, the external camera check, depth, flow, and truth are not recorded.

\begin{figure}[t]
\centering\supfig{A2}
\includegraphics[width=\columnwidth]{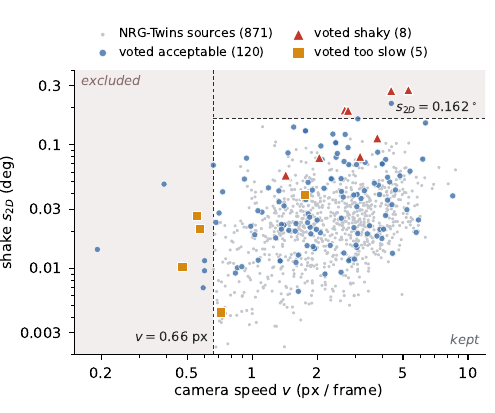}
\caption{Votes on real clips against the source screen of Supp.~\ref{app:sources}. Each marker is one voted clip at its measured camera speed $v$ and shake $s_{2D}$; grey dots are the 871 sources of NRG-Twins. Dashed lines: thresholds; the tinted region is excluded. Two votes of unsure are left out; logarithmic axes.}
\label{fig:votes}\supfigend
\end{figure}

\paragraph{Origin of the thresholds (Table~\ref{tab:a11}).} Both thresholds come from votes on real clips (RealEstate10K, DL3DV and TUM RGB-D): each clip was shown as a looping video and labelled acceptable, shaky, too slow or unsure. Batches 1--5 (25 clips) set the thresholds between the most extreme accepted and the mildest rejected clip: largest accepted $s_{2D}$ $0.1615^\circ$, smallest among clips rejected for shake $0.189^\circ$ (excluding the $0.057^\circ$ TUM clip described below); smallest accepted $v$ 0.661\,px, largest rejected 0.476\,px. All four TUM clips were voted shaky, one of them at $s_{2D}=0.057^\circ$, which the measure does not catch; TUM is therefore excluded as a source. Votes cast after the thresholds were fixed (Fig.~\ref{fig:votes}; unsure votes left out) agree with the screen as follows.
\begin{itemize}
\item 30 clips, 17 of them at the extremes of the two measures, the remaining TUM clip and 12 at random: 26 of 30.
\item 60 clips drawn at random, preregistered: 57 of 59, 96.6\% (Wilson 95\% interval 88.5--99.1\%).
\item 20 clips nearest to the thresholds, preregistered: 14 of 19.
\item Of the 98 clips the screen keeps, 5 were voted shaky or too slow; of the 10 it excludes, 6 were voted acceptable.
\end{itemize}
The 871 sources of NRG-Twins have $s_{2D}\le0.14522^\circ$ (median $0.023^\circ$) and $v\ge0.663$\,px (median 2.28\,px).
\paragraph{Twins kept (Table~\ref{tab:a12}).} 780, 93 and 96 twins (training, validation, test) completed the labelling chain. A twin is excluded if its raw keyframe RMS exceeds $0.6^\circ$ (76 twins), its noise-subtracted RMS exceeds $0.3^\circ$ (8), its keyframe noise cannot be computed (12), or it fails the selected-camera check, which requires at least 99\% of the finite point--camera observations to have positive depth (2: one below the threshold, one not checkable); 702, 82 and 87 remain. The test evaluation also requires every keyframe of a twin to be scorable, which removes 4 test twins and leaves the 83 of Sec.~5.1 of the main paper.
\paragraph{Release.} We release on Hugging Face, for research use, the generated twins and their real clips, the source-clip list, the label maps of all three splits, cameras, depth maps, the full per-frame truth archives and the construction code. The generated twins, labels and derived data are licensed under CC BY-NC 4.0 and the code under the MIT licence. The real frames come from RealEstate10K YouTube videos; we do not relicense them, and their rights remain with their owners. Removal requests are handled through the dataset's public discussion page. The test labels are public and are meant for final evaluation only, not for model selection, threshold choice or tuning.

\section{Ablation: noise-corrected vs.\ raw NRG angle}\label{app:ablation}

\paragraph{NRG-PD implementation.}
The VGGT-1B backbone alternates 24 frame-attention and 24 global-attention blocks. The pretrained backbone is frozen, and rank-16 LoRA~\cite{hu2021lora} updates the joint query--key--value projection, output projection, and both MLP linear layers in the last eight blocks of each stream. We initialize the DPT~\cite{ranftl2021dpt} head from VGGT's depth decoder and initialize its final output convolution to zero. The LoRA branches and DPT head together contain $36{,}849{,}031$ trainable parameters.

We sample 24-frame RGB windows and their noise-corrected labels from the same twins. We draw the stride uniformly from $\{1,2,3\}$, draw the start uniformly over starts that fit that stride, and reverse the frame order with probability $0.5$. Copied keyframes remain eligible as input frames. Both losses use the same valid non-keyframe pixels. The ordinal loss applies binary cross-entropy to six logits for whether the noise-corrected angle reaches each bin boundary. We use AdamW with linear warm-up to $10^{-4}$ over the first 300 updates, moment coefficients $(0.9,0.999)$, and weight decay $0.01$. Table~\ref{tab:train} lists all training and checkpoint-selection settings.
\begin{table}[h]\centering\footnotesize\suptab{B1}
\begin{tabularx}{\columnwidth}{@{}lL@{}}\toprule
Setting & Value \\\midrule
Input per step & one 24-frame window, frames resized to $518\times294$ \\
Steps & 3{,}000 \\
Optimiser & AdamW, learning rate $10^{-4}$, $(0.9,0.999)$, weight decay 0.01 \\
Schedule & linear warm-up over 300 steps, then constant \\
Gradient clipping & 1.0 \\
Precision & bfloat16 mixed precision \\
Seeds & 20261108, 20261109, 20261110 \\
Checkpoint & lowest loss on 20 validation twins; within 1\% of the lowest counts as tied, the later step is kept \\
Test split & not used for any choice \\\bottomrule
\end{tabularx}
\caption{Training settings of NRG-PD.}
\label{tab:train}\suptabend
\end{table}

\paragraph{Twin inference.}
NRG-PD takes the generated RGB frames as its sole input. We run 24-frame windows starting at frames 0, 19, 38, and 57, retaining predictions for frames 0--20, 21--39, 40--58, and 59--80, respectively. We convert the signed regression output $\hat d$, which estimates the target $T$ of Supp.~\ref{app:noise}, to an angle as
\begin{equation}
\hat\theta=0.6^\circ\sqrt{\max(\hat d,0)}.
\end{equation}
Since $\hat d$ estimates an expected squared angle under conditions (i)--(iv) of Supp.~\ref{app:noise}, $\hat\theta$ is a root-mean-square angle given the video: not the actual angle of an individual pixel and not the mean angle.

\begin{itemize}
\item NRG-PD bins the noise-corrected NRG angle into 7 classes; NRG-PD (raw angle) bins the raw angle into 6. The choice between the two targets was made on a selection set of 26 twins drawn at random (seed 20261108) from the 258 twins available on 8 November, with provisional models trained on the other 232; 25 of them pass the keyframe rule, and all belong to the final training split. There the noise-corrected version ranks better and is better calibrated (mean per-clip AUROC difference $+0.039$, Kendall $\tau$ $+0.107$, ECE $-0.096$; clip-bootstrap intervals exclude 0). On the test set the two final models give AUROC $0.851\pm0.036$ vs.\ $0.863\pm0.012$ and Kendall $\tau$ $0.646\pm0.053$ vs.\ $0.668\pm0.014$ (mean $\pm$ s.d.\ over 3 seeds). Table~\ref{tab:ablation_raw} repeats the test-set checks of Table~\ref{tab:s5.9tab} for both versions.
\end{itemize}
\begin{table}[h]
\centering\footnotesize
\begin{tabularx}{\columnwidth}{@{}Lcc@{}}\toprule
Check & NRG-PD & \begin{tabular}[b]{@{}c@{}}NRG-PD\\(raw angle)\end{tabular} \\
\midrule
AUROC, seed 1 & 0.865 & 0.851 \\
AUROC, seed 2 & 0.810 & 0.874 \\
AUROC, seed 3 & 0.877 & 0.864 \\
Rank corr.\ with depth & $-$0.22 to $-$0.32 & $-$0.22 to $-$0.33 \\
Depth as score, near & 0.575 & 0.575 \\
Depth as score, far & 0.425 & 0.425 \\
AUROC in 10 depth bins (median) & 0.74--0.85 & 0.80--0.85 \\
\bottomrule
\end{tabularx}
\caption{Test set, 83 twins; the checks of Table~\ref{tab:s5.9tab} for both versions.}
\label{tab:ablation_raw}
\end{table}

\section{Baseline protocol}\label{app:baselines}

The generated-video column of Table~1 in the main paper follows the evaluation task of the original methods. The clip-score and per-pixel-map columns distinguish aggregate outputs from the spatial maps used for pixel evaluation. VBench-2.0~\cite{zheng2025vbench} uses feature matching and camera-motion checks to return a single multi-view consistency score per eligible clip. Pixel AUROC uses methods with spatial score maps.

GeCo~\cite{gu2025geco} combines motion and structure errors into a per-frame map and averages the maps spatially and temporally for its video score. Its WarpBench injects temporally smooth foreground warps, and its OccluBench places an inconsistent object in a region revealed after occlusion.

\begin{itemize}
\item GeCo: official fused error (UFM~\cite{zhang2025ufm} flow + VGGT depth), 20-frame centred window per frame, with the official depth threshold 0.02, co-visibility threshold 0.5, confidence percentile 20 and minimum confidence 0.2; native output $480\times848$.
\item MEt3R~\cite{asim2025met3r}: MASt3R~\cite{leroy2024grounding} + DINO16~\cite{caron2021emerging}/FeatUp~\cite{fu2024featup} cosine feature inconsistency at $256\times256$, overlap only; MEt3R's pair-level score aggregates aligned feature similarity over the valid overlap. The four-frame variant pairs target $t$ with $t{+}4$ and uses $t{-}4$ for the final four targets. The two additional pairings use frame 0 as the reference for every other target or take the maximum over gaps 1, 2, 4 and 8 in both directions and frame 0. On the 83 test twins, each on its own valid pixels, the four-frame pairing gives AUROC 0.595, the frame-0 pairing 0.630 and the multi-gap maximum 0.620; Table~2 of the main paper uses the four-frame pairing, the official setting.
\item RoMo: per pixel, the larger of the forward and backward Sampson~\cite{hartley2004multiple} residuals in RoMo's partial-motion detection step, before its threshold; native output on the $480\times832$ centre crop.
\item VGGT4D: attention saliency before its adaptive threshold; four temporal windows; native output $480\times832$.
\item MonST3R~\cite{zhang2025monst3r}: per-pair normalised norm of RAFT flow minus camera-and-depth flow, averaged over pairs, before its threshold; native output $288\times512$.
\item Easi3R~\cite{chen2025easi3r}: refined dynamic attention map before SAM2~\cite{ravi2025sam} refinement; native output $18\times32$ patches.
\item MegaSaM~\cite{li2025megasam}: official per-pixel motion probability; native output $42\times72$ (its tracker grid).
\item All: same 81 twin frames, official weights, maps returned to $480\times848$, area-averaged to $294\times518$, scored with the same evaluator. The external wrappers apply a $480\times832$ center crop, use method-specific preprocessing and runtime adapters, and export and align the score maps. The MonST3R and Easi3R wrappers use the RAFT checkpoint specified by their repositories. The MegaSaM wrapper uses its specified Depth Anything ViT-L, UniDepth V2 ViT-L, and RAFT Things checkpoints. Inside the official code the wrappers change no computation: the only edits export the continuous value named above, set output paths, load local weights and adapt the code to the installed PyTorch version. Every score keeps the orientation its method defines (higher means more motion or more inconsistency); no orientation was chosen on the test set, so a method can fall below chance, as MegaSaM does in Table~2 of the main paper.
\end{itemize}
\section{Checks of NRG-PD on the test set}\label{app:checks}
Table~\ref{tab:s5.9tab} lists the checks of NRG-PD on the 83 test twins. The last four rows ask whether the test AUROC only reflects that near pixels are more often NRG pixels. Depth is the fused reference depth, in the scale of each clip's own reconstruction and pooled over the 83 twins. The predictions of NRG-PD are weakly correlated with depth and higher on nearer surfaces (rank correlation $-0.22$ to $-0.32$ over the three seeds). Depth alone as the score, with the same labels, gives AUROC 0.575 when near pixels rank higher ($-Z$) and 0.425 when far pixels rank higher ($+Z$); both orientations are reported and neither was chosen on the test set. With the pixels split into 10 depth bins, the median over bins of the AUROC of NRG-PD is 0.74--0.85 over the three seeds. Depth alone therefore does not account for the test AUROC.
\begin{table}[h]\centering\footnotesize
\begin{tabularx}{\columnwidth}{@{}Lc@{}}\toprule
Check & NRG-PD \\
\midrule
AUROC, seed 1 & 0.865 \\
AUROC, seed 2 & 0.810 \\
AUROC, seed 3 & 0.877 \\
Rank corr.\ with depth & $-$0.22 to $-$0.32 \\
Depth as score, near & 0.575 \\
Depth as score, far & 0.425 \\
AUROC in 10 depth bins (median) & 0.74--0.85 \\
\bottomrule
\end{tabularx}
\caption{Checks of NRG-PD on the 83 test twins; AUROC 0.5 = chance.}
\label{tab:s5.9tab}
\end{table}

\section{Reconstruction of training twins and label statistics}
\paragraph{Motivation example (Fig.~1 of the main paper).} Of eight candidate RealEstate10K test pairs, we show the one with the largest real-minus-generated gap in mean held-out PSNR. The generated clip is conditioned on the first and last real frames. We estimate VGGT cameras separately for each clip and refine each set by global bundle adjustment over all 81 views. We fit RobustSplat~\cite{fu2025robustsplat} for 30{,}000 iterations with frames 33--47 excluded from fitting and score renders against the corresponding frames of each input. The displayed midpoint, frame 40, reaches 31.1\,dB for the real fit and 17.7\,dB for the generated fit. The rotated view orbits the frame-40 camera by $4^\circ$ about its vertical axis around the point on its optical axis at the median depth of the clip's reconstructed points.

\begin{itemize}
\item Training-pair reconstruction comparison (Sec.~1 of the main paper). On 29 training-split pairs, we fit with each video's own VGGT cameras refined by global bundle adjustment and score each fit against its own held-out frames (frames 4, 12, \dots, 76). The median over clips is 37.6\,dB for the real fits and 32.4\,dB for the generated fits.
\item After excluding copied keyframes (Table~\ref{tab:e1}), the training, validation and test splits contain 53{,}501, 6{,}303 and 6{,}639 frames, with 10{,}420{,}347{,}997{,} 1{,}202{,}158{,}642 and 1{,}314{,}759{,}567 labelled pixels after noise correction, respectively (labelled: accepted flow, finite angle and $t_\perp/Z\le0.01$, the rule of training, evaluation and the released labels).
\item Pixels per label bin ($<0.075^\circ$ / 0.075--0.3 / 0.3--0.6 / 0.6--1.5 / $>1.5^\circ$): TRAIN 66.59 / 25.39 / 6.52 / 1.39 / 0.10\%; VAL 68.00 / 23.34 / 7.30 / 1.24 / 0.13\%; TEST 66.33 / 25.45 / 6.57 / 1.58 / 0.06\%.
\end{itemize}

\begin{table}[t]
\centering\footnotesize\suptab{E1}
\begin{tabular*}{\columnwidth}{@{\extracolsep{\fill}}lccc@{}}\toprule
 & Train & Val & Test \\
\midrule
Pairs & 702 & 82 & 87 \\
Frames & 53{,}501 & 6{,}303 & 6{,}639 \\
Labelled pixels & 10.42\,G & 1.20\,G & 1.31\,G \\
\midrule
\multicolumn{4}{@{}l}{\emph{Share per angle bin (\%)}} \\
$<0.075^\circ$ & 66.59 & 68.00 & 66.33 \\
$0.075$--$0.3^\circ$ & 25.39 & 23.34 & 25.45 \\
$0.3$--$0.6^\circ$ & 6.52 & 7.30 & 6.57 \\
$0.6$--$1.5^\circ$ & 1.39 & 1.24 & 1.58 \\
$>1.5^\circ$ & 0.10 & 0.13 & 0.06 \\
\bottomrule
\end{tabular*}
\caption{Label statistics per split after excluding copied keyframes: labelled pixels (accepted flow, finite angle, $t_\perp/Z\le0.01$) after noise correction and their share per angle bin.}\label{tab:e1}\suptabend
\end{table}
\section{Generated frames inside a real clip: inputs and rules}\label{app:hybrid}
\begin{table*}[t]
\centering\footnotesize\setlength{\tabcolsep}{2pt}
\begin{tabular*}{\textwidth}{@{\extracolsep{\fill}}lcccccc@{}}\toprule
 & \multicolumn{2}{c}{Injected 2D warp, frames 28--51} & \multicolumn{2}{c}{Injected 2D warp, all frames} & \multicolumn{2}{c@{}}{Generated frames 25--55} \\
\cmidrule(lr){2-3}\cmidrule(lr){4-5}\cmidrule(l){6-7}
Mask & PSNR (dB) & Detector $-$ row & PSNR (dB) & Detector $-$ row & PSNR (dB) & Detector $-$ row \\
\midrule
No mask & 21.06 & $+$3.21 [$+$0.45, $+$5.82] & 21.22 & $+$0.08 [$-$0.01, $+$0.45] & 30.91 & $+$3.40 [$+$0.31, $+$5.75] \\
Random (same count) & 21.18 & $+$2.92 [$+$0.46, $+$5.38] & 21.31 & $+$0.02 [$-$0.07, $+$0.38] & 29.95 & $+$4.36 [$+$1.02, $+$7.11] \\
Ground-truth mask / real video$^\ast$ & 32.60 & $-$4.40 [$-$8.11, $-$2.83] & 29.48 & $-$7.14 [$-$10.26, $-$2.54] & 36.82 & $-$2.50 [$-$4.82, $-$1.14] \\
\midrule
MEt3R & 21.26 & $+$3.04 [$+$0.40, $+$5.31] & 21.20 & $+$0.04 [$-$0.02, $+$0.20] & 31.34 & $+$3.00 [$-$0.92, $+$6.66] \\
GeCo & 21.37 & $+$2.04 [$+$0.59, $+$4.58] & 21.25 & $+$0.07 [$-$0.04, $+$0.42] & 29.95 & $+$4.40 [$-$0.19, $+$8.85] \\
MonST3R & 21.31 & $+$2.18 [$+$0.50, $+$5.35] & 20.32 & $+$0.10 [$-$0.02, $+$0.33] & 28.10 & $+$6.24 [$+$1.75, $+$10.55] \\
Easi3R & 21.18 & $+$2.95 [$+$0.36, $+$7.10] & 20.34 & $+$0.18 [$-$0.02, $+$0.68] & 30.94 & $+$3.41 [$-$0.78, $+$7.22] \\
MegaSaM & 21.10 & $+$3.31 [$+$0.46, $+$5.31] & 20.89 & $+$0.08 [$-$0.02, $+$0.18] & 23.34 & $+$11.01 [$+$7.15, $+$14.41] \\
RoMo & 21.31 & $+$1.81 [$+$0.38, $+$3.45] & 21.24 & $+$0.02 [$-$0.11, $+$0.20] & 29.16 & $+$5.19 [$+$0.38, $+$9.54] \\
VGGT4D & 21.97 & $+$2.57 [$+$0.37, $+$4.77] & 21.21 & $+$0.09 [$-$0.02, $+$0.60] & 26.69 & $+$7.65 [$+$4.04, $+$11.26] \\
\midrule
NRG-PD (+injected) & \textbf{24.02} & -- & \textbf{21.87} & -- & -- & -- \\
NRG-PD & -- & -- & -- & -- & \textbf{34.31} & -- \\
\bottomrule
\end{tabular*}
\caption{Masked 3DGS with RobustSplat for every mask (protocol in Supp.~\ref{app:hybrid} and~\ref{app:inject}). Bold: highest PSNR in each column, without the ground-truth row, which is an upper bound. Detector: NRG-PD (+injected), the NRG-PD architecture trained also on injected clips as in Sec.~5.3 of the main paper (Supp.~\ref{app:inject}), in the four injected-warp columns; NRG-PD in the two generated-frame columns. Table~\ref{tab:inj3090} gives NRG-PD on the injected warp. Detector $-$ row: paired difference with its 95\% bootstrap interval; $>0$: the detector is better. Generated frames: for the seven baselines, four of the 22 clip--seed cells pair the baseline with an NRG-PD refit on an RTX PRO 6000, so the NRG-PD mean behind their differences is 34.35\,dB; the NRG-PD row shows the fits of the main paper (34.31\,dB), which the random and no-mask rows pair with. $^\ast$Ground-truth mask for the injected warp; fit on the real video, without generated frames, for the generated frames. In the generated-frame columns, the intervals against MEt3R, GeCo and Easi3R include 0.}
\label{tab:mask}
\end{table*}

Across arms, we fit RobustSplat for 30{,}000 iterations with the learned mask disabled. The binary keep mask assigns zero weight to masked pixels or all pixels of a dropped frame, and unit weight throughout the unmasked control. The random control for a detector pixel mask samples pixels uniformly without replacement at the same count per frame.

\begin{figure*}[t]
\centering
\includegraphics[width=\textwidth]{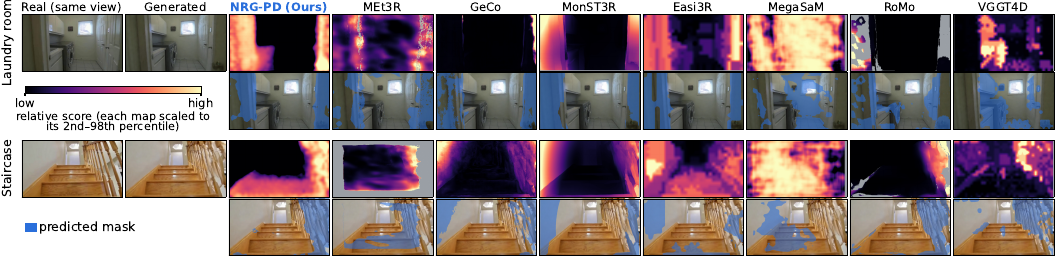}
\caption{Score maps (top row of each clip) and masks (bottom row, blue) of all methods at view 40, a generated frame, of the clips in Fig.~7 of the main paper; the first two columns are the real and the generated frame. Each map is scaled to its own 2nd--98th percentile in this frame; grey: no output. The mask of NRG-PD is the pixel rule of Sec.~5.3 of the main paper; every other method masks the same number of its highest-scored pixels.}
\label{fig:cmp}
\end{figure*}

Table~\ref{tab:mask} compares every mask in both experiments. Injected warp (Supp.~\ref{app:inject}): PSNR inside the warped region of the held-out frames, median over the 17 clips per input, with a paired clip bootstrap; the detector is NRG-PD (+injected), and each baseline masks per frame as many pixels as NRG-PD (+injected), its highest-scored ones. NRG-PD gives $+0.08$ [$-0.22$, $+0.91$]\,dB over random for frames 28--51 (Table~\ref{tab:inj3090}). Generated frames: whole-view PSNR at held-out views 36, 40 and 44 against the real frames, mean over 11 clips $\times$ 2 fit seeds, with a source-clustered bootstrap. The share of the gap closed in the main paper (0.70) compares, for each clip and fit seed, the gain of NRG-PD over no masking with the gain of the fit on the real video, $(\text{NRG-PD}-\text{no mask})/(\text{real video}-\text{no mask})$, and takes the median over the 22 clip--seed pairs; each baseline drops as many frames as NRG-PD, those with the highest frame score $r_t$ computed from its own score maps. Every paired difference compares fits on the same GPU type. For the injected warp, the main paper reports the original fits: in frames 28--51, NRG-PD (+injected) beats the random mask by $+3.07$\,dB and is better on 82\% of the clips, and the ground-truth mask beats it by $+8.54$\,dB. Some of these control fits had run on another GPU type than NRG-PD (+injected); refitting them on its type (42 fits), as in Table~\ref{tab:mask}, gives $+2.92$ [0.46, 5.38]\,dB and $+9.42$ [5.93, 11.71]\,dB. For the generated frames, the four clip--seed cells whose fits had run on an H200 were refit as complete cells on an RTX PRO 6000; the NRG-PD and random-drop rows keep the fits of the main paper. Pixel masks from NRG-PD on the generated frames give $-0.05$ [$-0.55$, 0.33]\,dB against random (3 clips $\times$ 1 seed). Fig.~\ref{fig:cmp} shows the score maps and masks of all methods on the clips of Fig.~7 of the main paper.

\begin{figure*}[p]
\centering
\includegraphics[width=\textwidth]{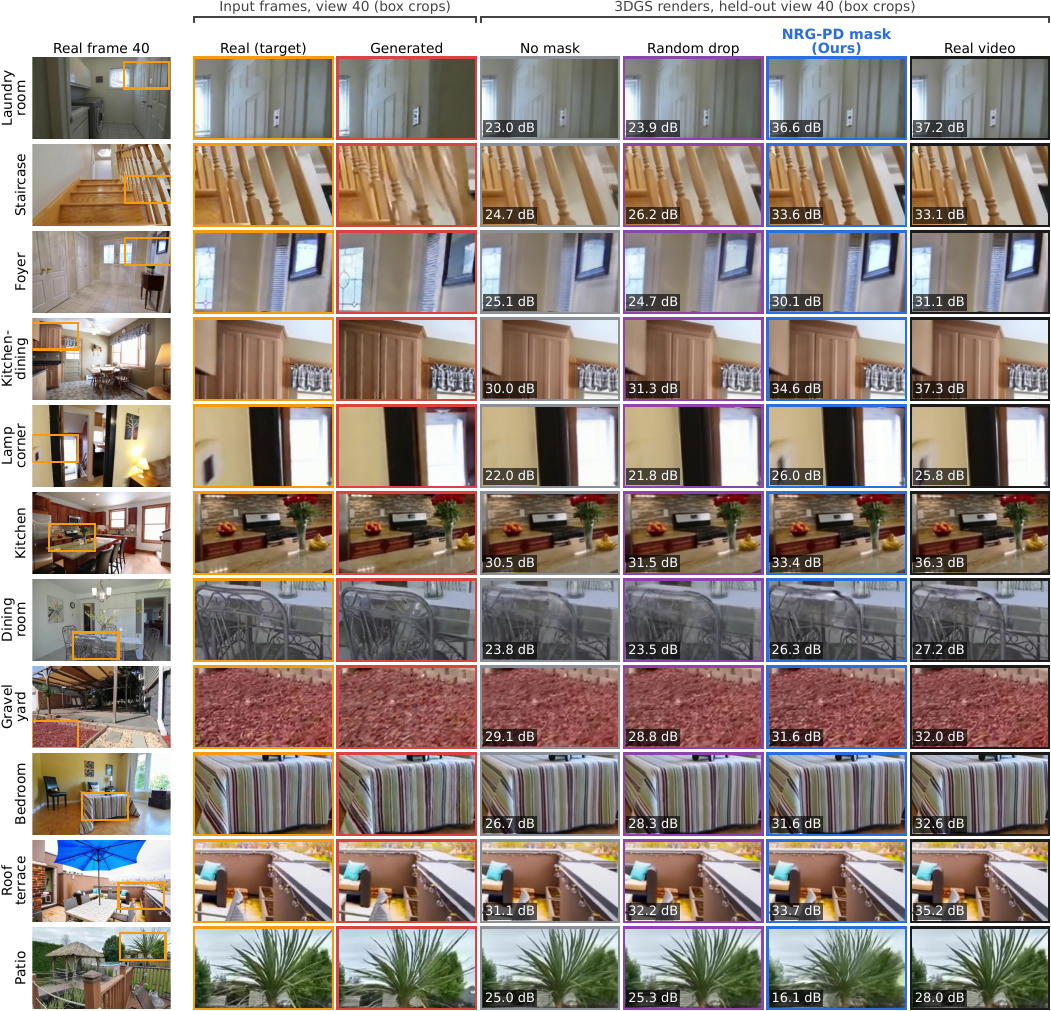}
\caption{All 11 test clips of the generated-frames experiment, ordered by the gain of NRG-PD frame dropping (Supp.~\ref{app:hybrid}); Fig.~7 of the main paper shows the first two. Columns as in Fig.~7 of the main paper; dB: PSNR of the box against the real frame.}
\label{fig:hyball}
\end{figure*}
\begin{figure*}[p]
\centering
\includegraphics[height=0.92\textheight]{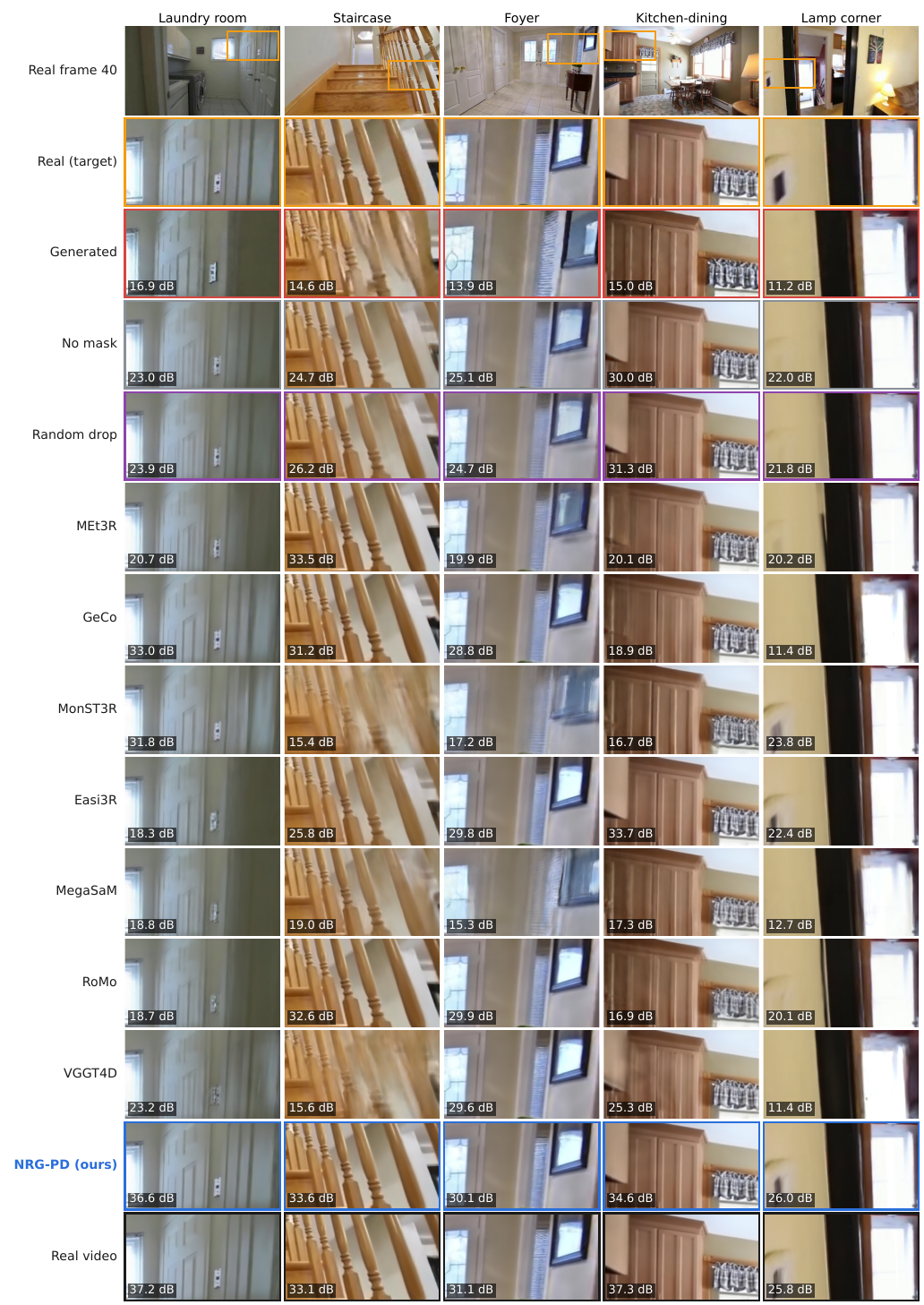}
\caption{Frame dropping by every method on the five test clips with the largest gain (Supp.~\ref{app:hybrid}; Fig.~7 of the main paper shows the first two). Top row: real frame 40 with the box; below, the box in the real frame, in the generated input frame and in the renders at held-out view 40 for no mask, random dropping, each baseline, NRG-PD and the fit on the real video. Each baseline drops as many frames as NRG-PD. dB: PSNR of the box against the real frame.}
\label{fig:hybbase}
\end{figure*}
\paragraph{Frame dropping with NRG-PD (+injected).} Table~\ref{tab:hybours2d} applies the same frame rule and threshold to the maps of NRG-PD (+injected) (Supp.~\ref{app:inject}) on the same 11 clips $\times$ 2 fit seeds. Its random control drops the same number of frames. Every compared fit of a clip--seed cell ran on one GPU type; the nine cells whose NRG-PD fits had run on an RTX PRO 6000 or an H200 were refit on an RTX 4090 with all their masks, so the NRG-PD row here differs slightly from Table~\ref{tab:mask}. NRG-PD (+injected) drops 2 of the 31 generated frames on one clip (NRG-PD: 19) and none on another (NRG-PD: 24); there it is 8.1 and 12.4\,dB below NRG-PD.
\begin{table}[t]
\centering\footnotesize\setlength{\tabcolsep}{3pt}
\begin{tabular*}{\columnwidth}{@{\extracolsep{\fill}}lcc@{}}\toprule
Frames dropped by & PSNR (dB) & NRG-PD (+injected) $-$ row \\
\midrule
No mask & 30.95 & $+$1.94 [$-$0.46, $+$4.10] \\
Random (same count) & 31.28 & $+$1.62 [$-$1.18, $+$4.26] \\
NRG-PD & 34.22 & $-$1.33 [$-$4.17, $+$1.14] \\
NRG-PD (+injected) & 32.89 & -- \\
\bottomrule
\end{tabular*}
\caption{Generated frames 25--55: frame dropping with NRG-PD (+injected). Whole-view PSNR at held-out views 36, 40 and 44, mean over 11 clips $\times$ 2 fit seeds; paired differences with 95\% source-clustered bootstrap intervals; NRG-PD (+injected) is better in 16, 16 and 11 of the 22 cells.}
\label{tab:hybours2d}
\end{table}

\begin{itemize}
\item Input: frames 0--80 of a real RE10K clip of the test split (11 clips), then the same 81 frames in reverse (162 frames); frames 25--55 of the forward half are replaced by Wan2.1-Fun-1.3B-InP~\cite{wan2025wan} frames generated from real frames 24 and 56 only. RobustSplat is trained with the real cameras; the held-out views 36, 40, 44 (copies 125, 121, 117) and the check frame 12 (copy 149) are not trained on, and frames 0, 80, 81 and 161 are never dropped or masked. A forward frame $t$ has reverse index $161-t$, so generated frames 25--55 have real counterparts at indices 106--136. Of the 126 real fitting frames, 122 are eligible for dropping after excluding the four protected boundary frames.
\item Frame rule. Let $s_t$ be the mean predicted NRG angle of frame $t$ over all its pixels, and $N(t)$ the 6 frames $u$ with $|u-t|>3$ whose cameras are closest to that of $t$ under the camera distance
\begin{equation}
d(t,u)=\angle(R_t,R_u)+10\,\frac{\lVert\mathbf{c}_t-\mathbf{c}_u\rVert}{D},
\end{equation}
where $\angle(R_t,R_u)$ is the rotation angle between the two cameras in degrees, $\mathbf{c}_t$ the camera centre of frame $t$, and $D$ the scene extent, the largest distance of a camera centre from the mean centre. Frame $t$ is dropped if
\begin{equation}
\begin{aligned}
r_t&=s_t-\operatorname*{median}_{u\in N(t)} s_u\ \ge\ c\,\sigma,\\
\sigma&=1.4826\cdot\operatorname*{median}_{t}\,\big|r_t-\operatorname{median}(r)\big|,
\end{aligned}
\end{equation}
where $r_t$ is the excess of frame $t$ over its camera neighbours and $\sigma$ the robust standard deviation of $r$ over the 162 frames. The rule thus uses only the detector's predictions on the 162 input frames and the given cameras. The frame score in Fig.~8 of the main paper is $r_t/\sigma$. $c=0.5$ was chosen on eight hybrids built from validation twins as the value that best separates generated from real training frames (Youden index of the drop rates 0.517, against 0.422 for $c=1$, 0.308 for $c=1.5$ and 0.214 for $c=2$).
\item Random control: drops the same number of frames as the rule in that clip, drawn uniformly from the droppable frames; one fixed draw per fit seed. Fig.~\ref{fig:hyball} shows all 11 test clips.
\item Per-clip results (Fig.~\ref{fig:hyball}): clips are ordered by the mean gain of NRG-PD frame dropping across fit seeds 0 and 1. Within each seed, the gain is the difference over the better of no masking and random dropping, using mean PSNR across held-out views 36, 40 and 44: laundry room $+10.2$, staircase $+6.4$, foyer $+5.7$, kitchen-dining $+4.6$, lamp corner $+4.6$, kitchen $+3.0$, dining room $+2.8$, gravel yard $+2.8$, bedroom $+2.7$, roof terrace $+1.8$, patio $-9.3$\,dB. On the patio clip the rule drops 5 generated and 39 real frames.
\item Figures~\ref{fig:hyball} and~\ref{fig:hybbase} use the columns of Fig.~7 of the main paper: real frame 40 with the $280\times158$ box of largest summed error, searched at an 8-pixel stride, taking at each pixel the larger error of the no-mask and NRG-PD renders; then the box in the real frame, in the generated input frame and in the renders at held-out view 40 (fit seed 0). In Fig.~\ref{fig:hybbase}, each baseline drops the same number of frames as NRG-PD, those with the highest frame score $r_t$ computed from its own score maps, and all fits of a column ran on the same GPU type.
\item Pixel rule, on the same detector output: a pixel of frame $t$ is masked if its predicted NRG angle exceeds its own median over the 162 frames by at least $0.06^\circ$ (the threshold $\delta$ of the injected-deformation experiment). Its random control masks the same number of pixels per frame. In Fig.~8 of the main paper, the generated frame and original are shown at the same source index. The blue overlay on the generated frame shows this pixel mask, and the annotation gives its masked fraction.
\end{itemize}

\section{Injected 2D deformation}\label{app:inject}
\begin{table*}[t]
\centering\footnotesize\setlength{\tabcolsep}{3pt}
\begin{tabular*}{\textwidth}{@{\extracolsep{\fill}}lccccccccc@{}}\toprule
Injected set & NRG-PD & NRG-PD (+injected) & GeCo & MEt3R & MonST3R & Easi3R & MegaSaM & RoMo$^\ddagger$ & VGGT4D$^\dagger$ \\
\midrule
2D warp, test split (62 clips) & 0.565 & \textbf{0.838} & 0.415 & 0.508 & 0.389 & 0.410 & 0.262 & 0.589 & 0.277 \\
\bottomrule
\end{tabular*}
\caption{Pixel AUROC on the injected test split (0.5 = chance; bold = best). 95\% scene-bootstrap intervals: NRG-PD [0.254, 0.645], NRG-PD (+injected) [0.663, 0.863]. RoMo$^\ddagger$: scored on its own valid pixels (NRG-PD there: 0.574). VGGT4D$^\dagger$: maps only for the 42 clips of the 228-clip set (NRG-PD there: 0.532).}
\label{tab:s5.11tab}
\end{table*}
\begin{figure*}[t]
\centering
\includegraphics[width=\textwidth]{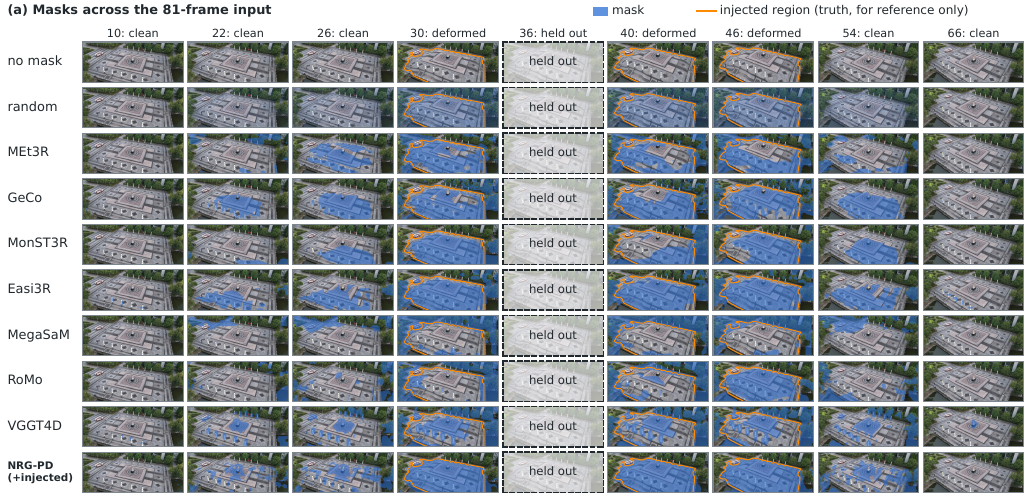}
\caption{Masks of every method over the 81 input frames of the clip of Fig.~6 of the main paper (deformation in frames 28--51; Supp.~\ref{app:inject}). Orange: warped region, for reference only.}
\label{fig:injmasks}
\end{figure*}
\begin{figure*}[p]
\centering
\includegraphics[width=\textwidth]{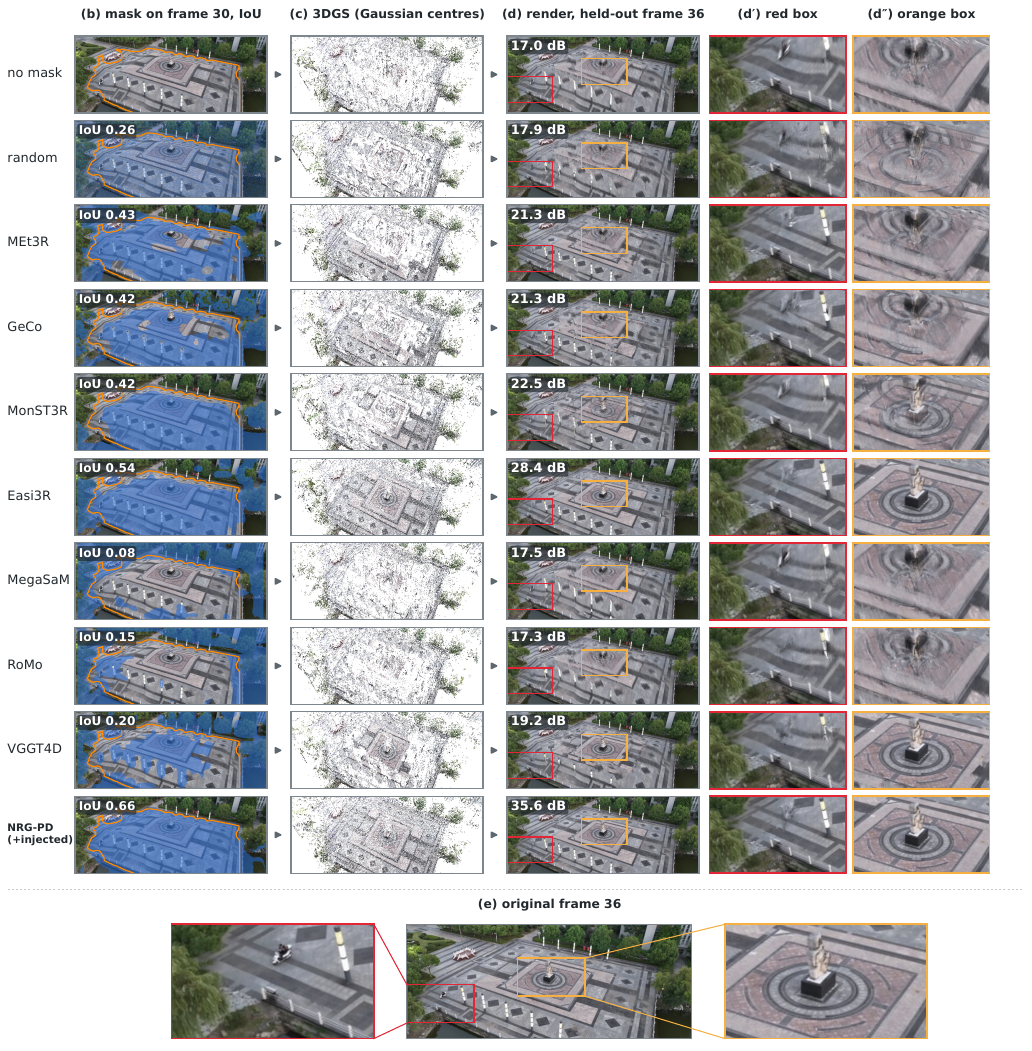}
\caption{Panels (b)--(e) of Fig.~6 of the main paper for every mask of Fig.~\ref{fig:injmasks} (Supp.~\ref{app:inject}): (b) the mask on training frame 30 and its IoU with the warped region, pooled over the fitting frames without the two endpoints; (c) centres of the fitted Gaussians (opacity $>0.5$) from a raised viewpoint; (d) render at held-out frame 36 with its regional PSNR; (d$'$), (d$''$) the red and the orange box of (d), enlarged; (e) the original frame 36 with both boxes enlarged. The mask marked as ours in Fig.~6 of the main paper is that of NRG-PD (+injected).}
\label{fig:injrecon}
\end{figure*}
\paragraph{Injected clips.} We take DL3DV~\cite{ling2024dl3dv} videos (81 frames, $848\times480$) and displace one segmented region (an object, the ground or the background) in 2D. With $\mathbf{d}$ a fixed unit direction, $f$ the focal length and a peak angle $\theta_{\max}\in\{1.2,2.4,3.6,4.8,6.0\}^\circ$, pixel $\mathbf{u}$ of frame $t$ moves by
\begin{equation} \Delta\mathbf{u}_t=f\tan(\theta_{\max})\,p(t)\,w(\mathbf{u})\,\mathbf{d}, \end{equation}
where $w\in[0,1]$ is non-zero only inside the region and falls smoothly to 0 at its border over 30--240\,px, and $p(t)$ is the time profile, one way or round trip:
\begin{equation}
\begin{aligned}
p_{\text{one way}}(t)&=\begin{cases}\tfrac12-\tfrac12\cos(\pi t/53), & t\le53,\\ 1, & t>53,\end{cases}\\
p_{\text{round trip}}(t)&=\tfrac12-\tfrac12\cos(2\pi t/80),
\end{aligned}
\end{equation}
so the round trip peaks at frame 40. The label of a pixel is the angle between its viewing ray and that of its source position; clean clips are the same videos without the warp and have label 0. Two sets were built: 135 warped clips from 25 scenes with 32 clean clips (32 scenes in all), and 190 warped clips with 38 clean clips (228 in total, 38 scenes); together they cover 38 scenes.
\paragraph{NRG-PD (+injected).} NRG-PD (+injected) is the NRG-PD architecture trained also on injected clips, as in Sec.~5.3 of the main paper. The 38 scenes are split at random into 23 for training, 8 for validation and 7 for testing, with all clips of a scene in one split: 245 / 88 / 62 clips, of which 201 / 74 / 50 are warped. The detector has the architecture, loss and target of NRG-PD (Sec.~4 of the main paper), with $\hat\sigma^2=0$ for injected clips. Each training batch holds 4 windows from the 702 training twins and 4 from the 245 injected training clips; we train 1{,}000 steps for each of 3 seeds, then continue to step 1{,}300 with linear learning-rate decay. The detection evaluation averages the predictions of the three validation-selected checkpoints (steps 1{,}300, 1{,}100 and 1{,}200 for seeds 20261110, 20261111 and 20261112). The injected-deformation reconstruction masks of Sec.~5.3 of the main paper use the checkpoint selected on the 74 injected validation clips before training was continued (seed 20261111, step 1{,}000), fixed before any reconstruction.
\paragraph{Detection.} The metric is the pixel AUROC with pixels of $\theta\geq0.6^\circ$ as positives and pixels of $\theta<0.075^\circ$ as negatives, sampling up to 4{,}000 pixels per clip from the support common to the compared methods. On the test split (62 clips, 7 scenes), NRG-PD reaches 0.565; NRG-PD (+injected) reaches 0.838 [0.663, 0.863], so this deformation type is learnable once it is in the training data (Table~\ref{tab:s5.11tab}). On the full injected sets, NRG-PD reaches 0.699 (135 warped and 32 clean clips) and 0.675 (190 warped and 38 clean clips; next Easi3R 0.60).
\paragraph{Reconstruction example (Fig.~6 of the main paper).} The IoU annotations beside frame 30 pool mask intersections and unions over the fitting frames, excluding the two endpoints. The displayed regional PSNR pools squared RGB errors over the warped pixels across the deformed held-out views (frames 28, 36 and 44) before conversion to decibels. Fig.~\ref{fig:injrecon} repeats these panels for every mask of Fig.~\ref{fig:injmasks}. Besides the orange box of Fig.~6 of the main paper, it marks a red box at the left edge of the frame ($200\times113$ pixels, $x=0$--$199$, $y=250$--$362$ of the $848\times480$ frame); its mask and render panels show rows 70--409 of the frame. For the seven baselines, the Gaussian centres come from a refit with the same mask, seed and GPU type that kept the model; its regional PSNR is within 0.06\,dB of the scored fit shown.
\paragraph{Same-position control (Table~\ref{tab:g1}).} The same pixel and frame with vs.\ without injection, which removes location and content effects. Table~\ref{tab:g1} uses the test split (7 scenes), which NRG-PD (+injected) never saw in training. On the two full injected sets, which contain the training scenes of NRG-PD (+injected) and are therefore not used to compare it, the median pixel AUROC over the warped clips with defined scores (135 of 135 and 161 of 190) is GeCo 0.66 / 0.63, Easi3R 0.61 / 0.58, NRG-PD 0.58 / 0.57, MonST3R 0.57 / 0.56, MEt3R 0.55 / 0.54, VGGT4D -- / 0.51, MegaSaM 0.12 / 0.11.
\begin{table}[t]
\centering\footnotesize\suptab{G1}
\setlength{\tabcolsep}{4pt}
\begin{tabular*}{\columnwidth}{@{\extracolsep{\fill}}lcc@{}}\toprule
Method & Clips & Test split, AUROC \\
\midrule
NRG-PD (+injected) & 40 & \textbf{0.69} [0.54, 0.84] \\
Easi3R & 40 & 0.61 [0.53, 0.63] \\
GeCo & 40 & 0.57 [0.50, 0.76] \\
NRG-PD & 40 & 0.54 [0.50, 0.60] \\
VGGT4D & 25 & 0.53 [0.50, 0.61] \\
MEt3R & 39 & 0.53 [0.49, 0.57] \\
MonST3R & 40 & 0.51 [0.47, 0.56] \\
MegaSaM & 40 & 0.25 [0.08, 0.43] \\
\bottomrule
\end{tabular*}
\caption{Same-position control on the injected test split (7 scenes, 50 warped clips; NRG-PD (+injected) never trained on these scenes): for each warped clip, the AUROC of the method's scores at the warped pixels against its scores at the same pixels and frames of the clean clip of the same scene; median over the warped clips with a defined AUROC (Clips), with 95\% scene-bootstrap intervals; 0.5 = chance, bold = best.}\label{tab:g1}\suptabend
\end{table}
\paragraph{Masked 3DGS (Sec.~5.3 of the main paper).} For each deformation condition, we include a clip when the ground-truth mask improves regional PSNR by more than 1\,dB over its per-frame count-matched random mask. Of 30 candidate clips, 5 share frames with the injected data used to train or test the detector and were excluded before any fit; of the other 25, the rule keeps 17 per input. The partial input carries the warp only in frames 28--51, the full input in all frames.
\begin{itemize}
\item Fit and score: RobustSplat uses 71 fitting frames and 10 held-out frames (every eighth frame starting at frame 4). We pool squared RGB errors against the clean original over pixels displaced by at least 0.5\,px in the held-out views, using only frames 28--51 for the partial warp and all held-out views for the full warp. We report the median paired clip difference, with a 95\% interval from the 2.5th and 97.5th percentiles of 2{,}000 clip bootstrap medians. The positive share is the fraction of clips on which the detector mask strictly improves regional PSNR over its matched random mask.
\item Detector: NRG-PD (+injected); checkpoint chosen on 74 injected validation clips. Masks of NRG-PD overlap the injected region little: pooled IoU 0.075 and 0.098 for the two inputs, against 0.060 and 0.054 for random masks and 0.252 and 0.275 for NRG-PD (+injected).
\item Margin: on the prediction grid, we subtract each pixel's median angle across all 81 frames and threshold the resulting difference. We resize the binary mask by nearest-neighbor interpolation before fitting. The smallest tested margin with median precision at least 0.70 over the 74 validation clips is $\delta=0.06^\circ$ (median precision 0.715). For a frame-constant additive offset at a pixel, the temporal median and the predictions shift equally, leaving the mask unchanged.
\item Random control: the same number of masked pixels per frame as our mask, drawn at random. Each baseline masks per frame as many pixels as NRG-PD (+injected), taking its highest-scored pixels (Fig.~\ref{fig:injmasks}). The ground-truth mask masks the warped pixels (displacement $\geq0.5$\,px).
\end{itemize}

\paragraph{Both detectors on one GPU type.} Table~\ref{tab:inj3090} refits, for the same 17 included clips per input, the masks of NRG-PD and of NRG-PD (+injected), each with its own count-matched random mask, and no mask: 170 fits, all on an RTX 3090. Metric, inclusion rule and bootstrap as above.
\begin{table}[t]
\centering\footnotesize
\begin{tabular*}{\columnwidth}{@{\extracolsep{\fill}}lc@{}}\toprule
Paired difference & dB [95\% interval] \\
\midrule
\multicolumn{2}{@{}l}{\emph{Frames 28--51}} \\
NRG-PD (+injected) $-$ random & $+$3.38 [$+$0.28, $+$5.64] \\
NRG-PD (+injected) $-$ no mask & $+$3.46 [$+$0.49, $+$5.10] \\
NRG-PD $-$ random & $+$0.08 [$-$0.22, $+$0.91] \\
NRG-PD $-$ no mask & $+$0.05 [$-$0.03, $+$1.21] \\
NRG-PD (+injected) $-$ NRG-PD & $+$1.54 [$+$0.42, $+$4.94] \\
\midrule
\multicolumn{2}{@{}l}{\emph{All frames}} \\
NRG-PD (+injected) $-$ random & 0.00 [$-$0.06, $+$0.28] \\
NRG-PD (+injected) $-$ no mask & $+$0.09 [$+$0.01, $+$0.25] \\
NRG-PD $-$ random & $-$0.09 [$-$0.36, $+$0.00] \\
NRG-PD $-$ no mask & $+$0.00 [$-$0.11, $+$0.07] \\
NRG-PD (+injected) $-$ NRG-PD & $+$0.11 [$+$0.02, $+$1.45] \\
\bottomrule
\end{tabular*}
\caption{Injected 2D warp, masked 3DGS with both detectors, all 170 fits on an RTX 3090: median paired clip difference of regional PSNR with 95\% clip-bootstrap interval (17 clips per input). NRG-PD: trained on twins only.}
\label{tab:inj3090}
\end{table}

\section{Human study}\label{app:human}
The human study (Fig.~2 of the main paper) shows the same real clip twice, one side with a known 2D deformation (0.15, 0.3, 0.6, 1.2 or $2.4^\circ$; 6 clips per angle; both sides encoded identically), and asks whether the left clip, right clip or both clips appear deformed, with a can't-tell option and four obvious attention checks. Each stacked bar in Fig.~2 of the main paper shows the fraction of judgments in each response category at that angle. Eight people took part, and each agreed to take part before starting. One participant who failed three of the four attention checks is excluded, leaving 7 raters and 42 judgements per angle. Of these judgements, 2, 7, 13, 15 and 16 are correct at the five angles, and most others are ``can't tell'': even at $2.4^\circ$ fewer than half of the judgements identify the deformed side, and at $0.6^\circ$ and below at most 13 of 42 do. The stimulus applies a round-trip image-space warp to a single segmented region, with peak displacement $f\tan\theta$ for focal length $f$ and prescribed angle $\theta$. The deformed clip appears equally often on each side at each angle. The trial order is randomized with non-adjacent attention checks.

\section{Track test on prompt-generated video}\label{app:track}
\paragraph{What the test measures.} TN90 is computed from the tracks and the fitted cameras alone: no method score enters it, and the tracks it scores are not used to fit the cameras. A method enters the test in two places: its clean group supplies the tracks that fix the cameras, and its two groups are compared. The test therefore asks whether the tracks a method calls consistent support a rigid reconstruction that the tracks it flags do not fit.

\paragraph{Clips.} The 48 prompts combine 8 static scenes with three camera motions at two scales. Each prompt is a motion sentence, the scene description and the sentence ``Smooth, continuous camera movement from the first frame to the last.'' The motion sentences, with the scene's subject $\langle s\rangle$, feature $\langle f\rangle$ and background $\langle b\rangle$ inserted, are:
\begin{itemize}\setlength\itemsep{0pt}
\item orbit, small: ``The camera slowly arcs about 30 degrees to the right around $\langle s\rangle$, keeping it centered at the same distance and height, as $\langle f\rangle$ begins to turn toward the camera.''
\item orbit, large: ``The camera smoothly orbits a quarter circle, 90 degrees, to the right around $\langle s\rangle$, keeping it centered at the same distance and height, until $\langle f\rangle$ faces the camera.''
\item truck, small / large: ``Tracking shot: the camera glides sideways to the right by about half (the full) width of $\langle s\rangle$, facing straight ahead, so $\langle s\rangle$ drifts left in the frame with $\langle b\rangle$ sliding past behind it.''
\item dolly, small / large: ``The camera dollies forward toward $\langle s\rangle$, travelling a quarter of the way (halfway) to it, so $\langle s\rangle$ grows noticeably (much) larger in the frame.''
\end{itemize}
The scenes are: a stack of colorful wooden geometric blocks with a red arch block in front of a studio shelf; a matte clay pitcher with two cups in front of a wall shelf; a kitchen island with a butcher-block top and a wooden stool in front of cabinets; a wooden plinth with an architectural model and a small clock tower in a gallery; a stone sundial with a bronze plaque in a paved plaza; a wooden planter box of herbs with a watering can in a brick courtyard; a polished chrome vessel with a curved spout on a plinth in a studio; and a clear glass vessel with a gridded metal stand in a gallery. Each generator produced one clip per prompt with seed 0: LTX-2.5 (distilled pipeline; $832\times480$, 24\,fps, 81 frames, 8 sampling steps, guidance 1.0), CogVideoX1.5-5B ($1360\times768$, 16\,fps, 81 frames), Wan2.2-T2V-A14B ($832\times480$, 16\,fps, 81 frames) and HunyuanVideo-1.5 ($848\times480$, 24\,fps, 121 frames, of which the first 81 are used). All clips are resized to $848\times480$ without cropping and re-encoded before tracking.

\paragraph{Tracks and groups.} Tracks are the average of the point trackers CoTracker3~\cite{karaev2024cotracker3} and TAPNext++~\cite{jung2026tapnextpp}. A track with at least 10 valid samples takes the median score of the method along it. The methods' scores have no common unit, so on prompt videos the groups are relative: per clip, the top 30\% of tracks are called inconsistent and the bottom 30\% clean. On the 40 labelled twins, the reference score of a track is the median of its valid ground-truth angles $\theta_c$, which is in degrees, so tracks are grouped by the label definition: inconsistent if $\theta_c\ge0.6^\circ$ and clean if $\theta_c<0.075^\circ$ (the lowest 30\% if fewer than 50 tracks are below $0.075^\circ$); 23 of the 40 twins contain inconsistent tracks. The twins are the first 20 kept training twins and the first 20 kept validation twins in dataset order.

\paragraph{Cameras and TN90.} For each of three light bundle-adjustment fits, we draw up to 500 clean-labelled tracks without replacement, with a fixed seed per clip and draw. The fit starts from the VGGT cameras and refines the camera poses and one 3D point per track with Levenberg--Marquardt and a Schur complement over the points, a Cauchy loss of scale 1\,px and fixed VGGT intrinsics; the first camera is fixed and the second camera's translation has fixed length. Three damping values are tried and the lowest objective is kept. With the cameras fixed, we triangulate and reproject the tracks excluded from the fit and compute AUROC between the two groups using their finite TN90 values. For each scored track $j$, the reprojection error $e_j(f)$ in frame $f$ is divided by the clip's median error $m(g)$ at the same frame gap $g$, and
\begin{equation} \mathrm{TN90}_j=\operatorname{P}_{90}\Big\{\,e_j(f)/m(g_j(f))\;:\;f\,\Big\}, \end{equation}
is the 90th percentile over frames. We measure the gap from the track's seed frame and use positive gaps only. We estimate $m(g)$ from gaps with at least 50 finite errors from successfully triangulated tracks and require at least five such gaps. For a gap with fewer observations, we use the nearest supported gap's median. We floor $m(g)$ at 0.001\,px. The baselines read re-encoded copies of the clips; read on these same re-encoded copies, NRG-PD reaches a track AUROC of 0.845 (0.848 on its own inputs), so its lead does not come from input quality.

\paragraph{Why TN90.} The reprojection error of a track grows with the number of frames since its seed frame, so a score that summarises raw errors ranks long tracks higher regardless of inconsistency. In the track test the two groups of a method need not have the same length distribution, so the score must treat long and short tracks alike. TN90 divides each error by the clip's median error at the same frame gap, which removes this growth and keeps the error's evolution along the track. We chose it on two calibration sets with known injected deformation, 90 clips with an image-space warp injected in one region and 71 clips rendered from 3D Gaussian splats with one region deformed (clips with more than 1\% of fits on the depth floor excluded). The candidates are computed from the per-frame reprojection error $e(f)$ of a track with the cameras fixed and fall into four groups: summaries of $e$ over the track's frames (5th-largest value, mean, median, RMS, 90th percentile, the share of frames with $e>\tau$ averaged over $\tau=1,2,4,8,16$\,px, and the 5th-largest value over the first 20 frames after the seed); the 5th-largest value divided by the number of frames $n$ or by $\sqrt{n}$; descriptors of how the residual moves (Theil--Sen slope of $e$; directional cumulative error, the path length of the residual with steps that reverse direction counted negative; departure, the 90th percentile of the signed path toward or away from the fitted point; direction consistency, the length of the mean unit step of the residual); and time-normalised scores, which divide $e$ by the clip's median error at the same frame gap before taking the 5th-largest value or the 90th percentile (TN90). For each candidate score, Table~\ref{tab:tn90choice} gives the median over clips of the ratio between the median score of clean tracks of 41--80 frames and of 8--20 frames (1 = no length bias), and the AUROC of deformed against clean tracks. Two scores stay within 0.25 of 1 on both sets, TN90 (0.94 and 1.23) and the departure (1.04 and 1.04); TN90 has the higher AUROC (0.794 vs.\ 0.756 on 2D, 0.994 vs.\ 0.984 on 3DGS), within 0.03 of the best score on each set.
\begin{table}[h]\centering\footnotesize
\begin{tabular*}{\columnwidth}{@{\extracolsep{\fill}}lcccc@{}}\toprule
& \multicolumn{2}{c}{Long / short, clean} & \multicolumn{2}{c}{AUROC} \\
Track score & 2D & 3DGS & 2D & 3DGS \\\midrule
5th-largest error & 3.51 & 2.36 & 0.792 & 0.992 \\
5th-largest error $/\,n$ & 0.89 & 0.53 & 0.802 & 0.983 \\
5th-largest error $/\sqrt{n}$ & 1.79 & 1.12 & 0.806 & 0.992 \\
Mean error & 1.70 & 1.39 & 0.810 & 0.995 \\
Median error & 1.47 & 1.38 & 0.819 & 0.994 \\
RMS error & 1.82 & 1.43 & 0.798 & 0.995 \\
90th-percentile error & 1.93 & 1.53 & 0.793 & 0.993 \\
Share of frames above $\tau$ & 1.51 & 2.35 & 0.822 & 0.995 \\
5th largest, first 20 frames & 1.24 & 1.39 & 0.809 & 0.975 \\
Error slope & 0.56 & 0.18 & 0.697 & 0.880 \\
Directional cumulative error & 2.70 & 3.31 & 0.706 & 0.929 \\
Departure & 1.04 & 1.04 & 0.756 & 0.984 \\
Direction consistency & 0.47 & 0.33 & 0.598 & 0.741 \\
Time-normalised 5th largest & 1.49 & 1.75 & 0.805 & 0.993 \\
\textbf{TN90} & \textbf{0.94} & \textbf{1.23} & 0.794 & 0.994 \\\bottomrule
\end{tabular*}
\caption{Choice of the track score on the 2D (90 clips) and 3DGS (71 clips) calibration sets: every candidate we compared. Long / short: median score of clean tracks with 41--80 frames divided by that with 8--20 frames, median over clips (1 = no length bias). AUROC: deformed against clean tracks, median over clips. Scores as defined in the text.}
\label{tab:tn90choice}
\end{table}

\paragraph{Clip roster.} A clip is scored when every method has at least one light-BA draw that passes the publication checks of the frozen scorer on depth, error range and fit stability. This holds for 144 of the 192 clips for all six original methods, i.e.\ NRG-PD and five baselines (CogVideoX 31, HunyuanVideo 36, LTX 45, Wan 32). RoMo fails these checks in all three draws on one of the 144 clips, and MegaSaM is scored on the 168 clips where its own draws pass.

\paragraph{Reliable tracks and controls.} The forward--backward check measures the Euclidean distance between the two predictions at jointly visible frames, indexed in the original clip; a track is reliable if this distance is at most 2\,px at its seed frame and at every jointly visible frame, with at least two such frames. On the twins, the shuffle control permutes the reference scores among eligible tracks within each clip, regroups them and refits the cameras from the resulting clean group; on prompt videos, it permutes each method's labels within each clip with the cameras fixed. We resample whole clips with replacement, take the 2.5th and 97.5th percentiles of the bootstrap distribution, and report half the interval width.

\section{RobustSplat's learned mask}\label{app:rsmask}
RobustSplat learns a keep weight $W\in[0,1]$ for every pixel of every training frame and leaves a pixel out of the loss where $W\le0.25$ (after a $7\times7$ erosion). We fit RobustSplat as published (30{,}000 iterations, learned mask on) to 29 training twins with their own cameras and 71 training frames, and export $W$ at the end of training. RobustSplat stores the $W$ maps in its shuffled training order and records that order; we match each map to its frame through this record and compare the rejection score $1-W$ with the labels of the same frame (keyframes excluded), with pixels of $\theta_c\ge0.6^\circ$ as positives and $\theta_c<0.075^\circ$ as negatives, as in the main evaluation. An earlier run of this analysis paired the maps with the frames in sorted order, so each map was compared with the labels of another frame, used the raw angle with positives $\theta\ge0.3^\circ$ and negatives $\theta<0.15^\circ$, and gave AUROC 0.499. Matching the frames alone gives 0.446; with the matched frames and the labels of Sec.~5.1 of the main paper ($\theta_c\ge0.6^\circ$ and $\theta_c<0.075^\circ$), the AUROC is 0.640, the value in the main paper. Each measure is computed per twin; Table~\ref{tab:rsmask} gives the median over twins with a 95\% bootstrap interval.
\begin{table}[h]\centering\footnotesize\setlength{\tabcolsep}{3pt}
\begin{tabular*}{\columnwidth}{@{\extracolsep{\fill}}lc@{}}\toprule
Measure (29 twins) & Median [95\% interval] \\\midrule
AUROC of $1-W$, positives vs.\ negatives & 0.640 [0.595, 0.805] \\
Spearman corr.\ of $1-W$ with $\theta_c$ & $+0.007$ [$-0.042$, $+0.091$] \\
Labelled pixels left out of the loss & 0.000 [0.000, 0.000] \\
\bottomrule\end{tabular*}
\caption{RobustSplat's learned mask against the labels; AUROC 0.5 = chance. Pixels are left out in only 12 of the 29 twins, at most 10.9\% of a twin's labelled pixels.}
\label{tab:rsmask}
\end{table}

\section{Evaluation metrics}\label{app:metrics}

The baselines' scores are on different scales: none is an angle, and none is a probability that a pixel is an NRG pixel (MegaSaM's is a motion probability). We therefore evaluate them only with rank metrics (AUROC, Kendall $\tau$, clip correlation, average precision, and precision and recall of the top 5\%). These depend only on the order of the scores and are unchanged by any strictly increasing transform applied to all scores of a method, so no conversion to degrees is needed. The fixed cutoffs $\theta_c\ge0.6^\circ$ and $\theta_c<0.075^\circ$ define the NRG and rigid ground-truth pixels for every method; the angle error is defined only for NRG-PD, whose output is in degrees, and the empty MAE cells count neither for nor against a baseline.

For the paired AUROC intervals, we estimate pooled AUROC from score quantile histograms on the common valid pixel support. Positive--negative pairs in the same score bin receive half credit. We resample whole clips with replacement $2{,}000$ times and use the same clip draw for both methods in each comparison. For each detector seed and baseline, the $2.5$th and $97.5$th percentiles of the resampled AUROC differences give the interval. Table~\ref{tab:pairdiff} lists these differences for NRG-PD; all 21 intervals lie above 0. On its own pixels, the mean absolute error of NRG-PD rises with the NRG angle: $0.074$, $0.115$, $0.267$, $0.598$ and $1.600^\circ$ for $\theta_c<0.075^\circ$, $0.075$--$0.3^\circ$, $0.3$--$0.6^\circ$, $0.6$--$1.5^\circ$ and $>1.5^\circ$ (mean over the three seeds).
\begin{table}[h]\centering\footnotesize\setlength{\tabcolsep}{3pt}
\begin{tabular*}{\columnwidth}{@{\extracolsep{\fill}}lccc@{}}\toprule
 & \multicolumn{3}{c@{}}{NRG-PD training seed} \\\cmidrule(l){2-4}
Baseline & 20261108 & 20261109 & 20261110 \\\midrule
\multicolumn{4}{@{}l}{\emph{Difference}} \\
Easi3R & $+0.130$ & $+0.075$ & $+0.141$ \\
RoMo & $+0.213$ & $+0.158$ & $+0.225$ \\
MonST3R & $+0.290$ & $+0.235$ & $+0.302$ \\
GeCo & $+0.279$ & $+0.224$ & $+0.291$ \\
MEt3R & $+0.275$ & $+0.220$ & $+0.286$ \\
VGGT4D & $+0.301$ & $+0.246$ & $+0.313$ \\
MegaSaM & $+0.463$ & $+0.408$ & $+0.475$ \\\midrule
\multicolumn{4}{@{}l}{\emph{95\% interval}} \\
Easi3R & [0.080, 0.189] & [0.036, 0.121] & [0.097, 0.188] \\
RoMo & [0.143, 0.265] & [0.061, 0.236] & [0.130, 0.293] \\
MonST3R & [0.140, 0.427] & [0.071, 0.393] & [0.148, 0.452] \\
GeCo & [0.208, 0.341] & [0.134, 0.306] & [0.210, 0.362] \\
MEt3R & [0.187, 0.369] & [0.130, 0.309] & [0.191, 0.386] \\
VGGT4D & [0.210, 0.416] & [0.167, 0.357] & [0.237, 0.412] \\
MegaSaM & [0.398, 0.551] & [0.340, 0.500] & [0.416, 0.551] \\
\bottomrule
\end{tabular*}
\caption{Paired test AUROC difference, NRG-PD minus baseline, per NRG-PD training seed, on the 441 million pixels scored by all methods, with 95\% intervals from 2{,}000 resamples of the 83 test pairs.}
\label{tab:pairdiff}
\end{table}

For the ranking columns, each method is evaluated on the valid non-keyframe pixels covered by its score map. AUROC pools NRG pixels ($\theta_c \ge 0.6^\circ$) and rigid pixels ($\theta_c < 0.075^\circ$) across clips. NRG-PD uses the signed regression output for AUROC, Kendall $\tau$, and clip correlation. Baselines use their continuous scores oriented so that higher values indicate greater inconsistency. For angular MAE, we average $|\hat{\theta}-\theta_c|$ in degrees over all valid pixels for each seed. We report the arithmetic mean of the seed-level metrics over three seeds and use the sample standard deviation for the seed variability shown in the ranking columns.

With $s$ the score of a method, $\mathcal{N}$ the NRG pixels and $\mathcal{R}$ the rigid pixels (Sec.~5.1 of the main paper): Kendall $\tau$ compares pixel pairs whose labels differ by more than twice the noise: in each clip $v$ we draw 20{,}000 random pixel pairs and keep the set $S_v$ of pairs with $|\theta_c(p)-\theta_c(q)|>2\max\big(\hat\sigma(t_p),\hat\sigma(t_q)\big)$, where $t_p$ is the frame of pixel $p$ (Sec.~3.6 of the main paper); over the $V$ clips,
\begin{equation}
\begin{aligned}
\tau=\frac{1}{V}\sum_{v}\frac{1}{|S_v|}&\sum_{(p,q)\in S_v}\operatorname{sign}\big[(s(p)-s(q))\\
&\qquad\cdot(\theta_c(p)-\theta_c(q))\big],
\end{aligned}
\end{equation}
with 0 for chance and 1 for perfect order. Average precision is the mean, over the NRG pixels, of the share of NRG pixels among the pixels of $\mathcal{N}\cup\mathcal{R}$ scored at least as high:
\begin{equation} \mathrm{AP}=\frac{1}{|\mathcal{N}|}\sum_{p\in\mathcal{N}}\frac{|\{q\in\mathcal{N}:s(q)\ge s(p)\}|}{|\{q\in\mathcal{N}\cup\mathcal{R}:s(q)\ge s(p)\}|}, \end{equation}
which is $|\mathcal{N}|/|\mathcal{N}\cup\mathcal{R}|\approx0.023$ for a random score. Precision $P$ and recall $R$ use the top 5\% scored pixels $M_v$ of each clip $v$, with $\mathcal{N}_v$ its NRG pixels:
\begin{equation}
P=\frac{\sum_v|M_v\cap\mathcal{N}_v|}{\sum_v|M_v|},\qquad R=\frac{\sum_v|M_v\cap\mathcal{N}_v|}{\sum_v|\mathcal{N}_v|}.
\end{equation}
On the common labelled support, NRG pixels account for approximately $1.6\%$ of pixels. Dividing this prevalence by a nominal budget of $5\%$ of the shared labelled pixels gives a precision ceiling of approximately $0.31$. The clip correlation tests whether a method ranks whole clips: the Spearman correlation over clips between the 90th percentile of the score and that of $\theta_c$ within each clip.

\setlength{\bibsep}{0pt}

\end{document}